\documentclass[12pt]{article}

\usepackage{hyperref}
\usepackage{PRIMEarxiv}
\usepackage[utf8]{inputenc} 
\usepackage[T1]{fontenc}    
\usepackage{hyperref}       
\usepackage{url}            
\usepackage{booktabs}       
\usepackage{amsfonts}       
\usepackage{nicefrac}       
\usepackage{microtype}      
\usepackage{lipsum}
\usepackage{fancyhdr}       
\usepackage{graphicx}       
\graphicspath{{media/}}     
\usepackage{natbib}
\usepackage{todonotes}

\begin{document}
\title{Multilevel interpretability of artificial neural networks: leveraging framework and methods from neuroscience}
\maketitle

\vspace{-8em}


\textbf{Zhonghao He} \textsuperscript{\S,1},
\textbf{Jascha Achterberg} \textsuperscript{+,1,2},
\textbf{Katie Collins} \textsuperscript{*,1},
\textbf{Kevin Nejad} \textsuperscript{*,2},
\textbf{Danyal Akarca} \textsuperscript{*,2,3},
\textbf{Yinzhu Yang} \textsuperscript{*,1},
\textbf{Wes Gurnee} \textsuperscript{*,4} \\[0.5em]
\textbf{Ilia Sucholutsky} \textsuperscript{6,7},
\textbf{Yuhan Tang} \textsuperscript{1},
\textbf{Rebeca Ianov} \textsuperscript{1},
\textbf{George Ogden} \textsuperscript{1},
\textbf{Chole Li} \textsuperscript{1},
\textbf{Kai Sandbrink} \textsuperscript{2},
\textbf{Stephen Casper} \textsuperscript{4} \\[0.5em]
\textbf{Anna Ivanova} \textsuperscript{\P,4,5},
\textbf{Grace W. Lindsay} \textsuperscript{\P,6}

\small
\textsuperscript{1} University of Cambridge,
\textsuperscript{2} University of Oxford,
\textsuperscript{3} Imperial College London,
\textsuperscript{4} Massachusetts Institute of Technology,
\textsuperscript{5} Georgia Institute of Technology,
\textsuperscript{6} New York University,
\textsuperscript{7} Princeton University
\begin{center}
\footnotesize
\textsuperscript{\S} First author, correspondence to zh378@cam.ac.uk \\
\textsuperscript{+} Second author, significant contribution to project direction and framing. \\
\textsuperscript{*} Third co-authors, significant contributions to writing. \\
\textsuperscript{\P} Senior authors.
\end{center}

\vspace{2em}

\begin{abstract}

As deep learning systems are scaled up to many billions of parameters, relating their internal structure to external behaviors becomes very challenging. Although daunting, this problem is not new: Neuroscientists and cognitive scientists have accumulated decades of experience analyzing a particularly complex system - the brain. In this work, we argue that interpreting both biological and artificial neural systems requires analyzing those systems at \textbf{multiple levels of analysis}, with different analytic tools for each level. We first lay out a joint grand challenge among scientists who study the brain and who study artificial neural networks: understanding how distributed neural mechanisms give rise to complex cognition and behavior. We then present a series of analytical tools that can be used to analyze biological and artificial neural systems, organizing those tools according to Marr's three levels of analysis: computation/behavior, algorithm/representation, and implementation. Overall, the multilevel interpretability framework provides a principled way to tackle neural system complexity; links structure, computation, and behavior; clarifies assumptions and research priorities at each level; and paves the way toward a unified effort for understanding intelligent systems, may they be biological or artificial.

\end{abstract}
\section{Introduction}
\subsection{Shared Goals and Joint Challenges}

Interpretability research aims to provide a human-understandable explanation for model outputs and behaviors based on the input and model's internal structure \citep{doshivelez2017rigorous}. The field's goal is to generate mechanistic explanations of how neural networks perform computations and produce behaviors \citep{nanda2023progress, olsson2022context}, which could help predict the behavior of such networks across a wide range of scenarios and possibly solve notable problems of AI systems, such as hallucination and toxic output \citep{Ji_2023}. Being able to interpret AI systems is therefore a key capability to be able to understand whether models are appropriately fair, reliable, robust, and worthy of user trust \citep{doshivelez2017rigorous}.

However, understanding the computations of frontier AI systems with hundreds of billions of parameters presents many technical challenges, from the curse of dimensionality \citep{zhao2024uncoveringlargelanguagemodel, altman2018curse}  to finding a suitable unit of analysis \citep{olah2020zoom, zou2023representation}. 
 
These challenges are par for the course when studying complex systems. In particular, many challenges around artificial neural networks (ANN) interpretability are intimately familiar to another group of researchers: neuroscientists. Neuroscience (often in partnership with cognitive science and psychology) investigates how  neurons, their connections, and their activity patterns give rise to cognition and behavior. 

Similar to how deep learning researchers have recognized, neuroscientists have realized that simply examining activity profiles of individual neurons in response to a particular input is often insufficient for understanding how the system performs computation. Instead, complex neural systems are best understood across multiple levels of analysis -- considering behavior alongside the brain's connectome, population codes, and codes of single neurons to gain a holistic understanding of the inner workings of the brain \citep{Krakauer2017, barack2021two}. 
\footnote{As an example, mapping the full set of neurons and their connections in a tiny worm \emph{C. elegans} by itself was insufficient for describing how the nervous system gives rise to the worm's behavior \citep{Krakauer2017}.} 
To deal with the immense complexity of biological neural networks, as well as to establish systematic links between their structure and function, neuroscientists have honed their own analysis toolkit over several generations of scientific toil. This toolkit includes both conceptual frameworks \citep{marr2010vision, barack2021two, Krakauer2017, cisek2019resynthesizing, gyorgy2019brain, tinbergen1963aims} and analytical tools to understanding complex neural systems  across different levels of analysis.

Under this mindset, we think that the disciplines studying biological brains and their supported behavior can provide valuable insights that may advance interpretability research. Here, we will focus on the framework of Marr's three levels of analysis \footnote{As this manuscript was being prepared for submission, a position paper \citep{vilas2024position} was released also arguing for the use of Marr's levels in interpretability research. We believe this concurrent work further supports the importance of the approach we highlight. Furthermore, the two papers offer complimentary perspectives; specifically, ours is more focused on methods and findings at each level while theirs is more on the overall scientific approach of understanding neural networks.}, which is widely used in the brain sciences \citep{peebles2015thirty,bechtel2015non, hardcastle2015marr, johnson2017marr, baggio2015logic, eliasmith2015marr, mitchell2006mentalizing}. We see promise in the following aspects: (i) Marr's framework offers interpretability researchers guidance and clarity on how to use levels of analysis and abstraction, and different techniques to build up understanding, (ii) inter-level mutual constraints help to guide research across levels: both top-down and bottom-up approaches introduce constraints that guide research at a difference level and inconsistency between levels necessitates further investigation \citep{bechtel2015non, eliasmith2015marr}, (iii) their decades of research help to point out promising research directions for studying complex and dynamic neural networks and and (iv) in a similar veinm identify pitfalls that neuroscientists would advise interpretability researchers to avoid.

\subsection{Interpretability Today}
There are several notable research agendas and approaches to interpretability today (see \citep{zhao2024opening} for a review).
\textbf{Mechanistic Interpretability (MI)} is a bottom-up approach focused on understanding how the circuits within a network give rise to its behavior \citep{nanda2023progress}. The main assumptions underlying MI are (1) that features \footnote{Features, as far as this paper is concerned, are meaningful and interpretable neural representations, such as edge detector and ear detector in visual models, and concepts such as "Golden Gate Bridge" or more abstract concept such as gender bias. Note that features don't perfectly correspond to neural nodes as long as polysemanticity occurs.} are the fundamental unit of analysis, (2) that features are connected into computational circuits via network weights, and (3) that features and circuits are universal across models \citep{olah2020zoom}. 
While mechanistic interpretability mainly focuses on analyzing \textit{trained} neural networks, \textbf{Developmental Interpretability} is an emerging research direction that attempts to draw inferences on the computations of neural networks by analyzing their training process and learning dynamics \citep{nanda2023progress, hoogland2024developmental, davies2023unifying}, often by applying tools from singular learning theory \citep{watanabe2009algebraic}. Finally, \textbf{Representation Engineering (RepE)}, inspired by Hopfieldian view in neuroscience \citep{barack2021two}, adopts a top-down approach to model transparency by focusing on localizing and editing latent representations of concepts \citep{zou2023representation}.

How do these approaches relate to one another? Should we focus on circuits, features, and/or representations? How do we draw principled links between the internal organization of a model and its behavior? 

Below, we show that different interpretability tools can be put to work to address questions at different levels of analysis. Identifying those levels can help interpretability researchers establish the questions that are most appropriate to answer with each tool. We also believe it can help organize and unify disparate approaches and spur new research questions. 

\subsection{Our Contributions}

Building on key methods and findings from neuroscience research, we propose four ways to enhance the interpretability of machine learning models: Marr's conceptual framework, methods, lessons learned, and promising research directions. 

\begin{itemize}
  \item \textbf{Frameworks} We introduce Marr's multilevels of analysis that breaks down the understanding of complex neural systems into three levels: the computational, algorithmic/representational, and implementation levels of analysis. We also briefly introduce Tinbergin's four questions as a comprehensive approach to study neural system behaviors which we believe may be relevant to AI practioners, too. 
  \item \textbf{Research Questions} Marr's levels of analysis identify research questions neuroscientists and cognitive scientists would ask at each level. Those questions can be concurrently considered alongside the set of questions the AI interpretability community may be interested in.
  \item \textbf{Methods} At each level, we highlight methods from neuroscience and cognitive science that we think can add value to an AI interpretability researcher's repertoire (see \ref{figure2}). For each method, we cover a sampling of findings from brain sciences and detail about its relevance to interpretability research. 
  \item \textbf{Shared issues} We cover issues that may be salient to both communities of researchers, including common research topics (Grandmother neurons, Mixed selectivity / Polysemanticity) and meta-topics such as "what is necessary for understanding" and historical lessons from brain sciences.

\end{itemize}

\begin{figure}[t]
\caption{Marr's three levels of analysis and their application in understanding a calculator}
\includegraphics[width=\linewidth]{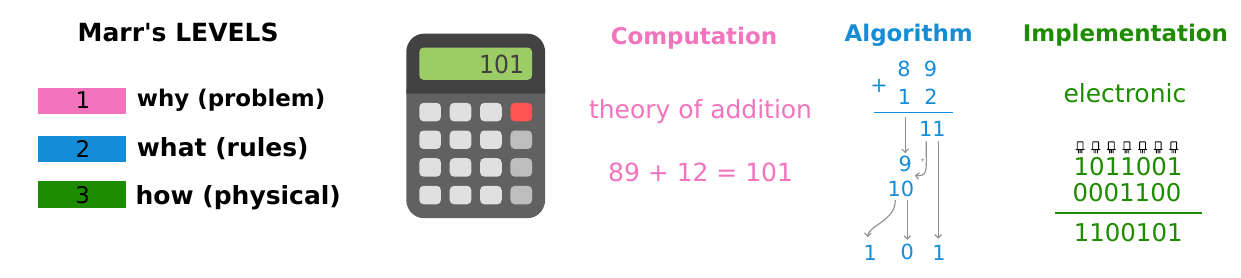}
\centering
\end{figure}

\subsection{How this paper is organized}
We first introduce \textbf{Marr's levels of analysis}, a conceptual multilevel framework for analyzing information processing systems, and explain how this framework can improve our understanding of system behaviors. 

We then survey methods from neuroscience/cognitive science that can yield insights at each level of analysis and compare these techniques with techniques used in interpretability research. When describing each technique, we give explicit examples of some of what has been learned through such techniques at this level and discuss any limitations.

In the discussion section, we explore what we think may be necessary for understanding neuoral systems and their behaviors, notable lessons from neuroscience that may help interpretability, and acknowledge differences between biological and artificial neural systems that may limit the connections between their studies. 

\section{Multilevel Analysis of Complex Neural Systems}

\subsection{Marr's Levels of Analysis: a framework for understanding neural systems}

David Marr proposed Marr's three levels of analysis as a conceptual framework for understanding complex neural networks\citep{marr2010vision}:

 \begin{enumerate}
\item \textbf{Computational.} What is the function of a neural system? What is the desirable behavior (outputs) as a function of inputs, current system state, and time?
\item \textbf{Representational/Algorithmic.} What is the series of computations that can achieve the computational function of the system? How should the relevant information be represented to implement these computations?
\item \textbf{Implementation.} What is the neural substrate of the algorithm? Which physical components of the system realize the algorithm/representation in question?
\end{enumerate}
 
For example, we could describe what a cash register at the checkout counter in a supermarket does:
\begin{itemize}
    \item At the computational level we know that the cash register performs addition.
    \item At the representation/algorithmic level we could choose from decimal, binary or Roman numerals as representations; for decimal numerals the addition algorithm follows the steps of (i) adding least significant digits first (ii) carrying to more significant ones if the sum exceeds ten.
    \item At the implementational level, there are different devices from mechanical ones to silicon-based circuits in mobile phones which can implement the required algorithmic steps.
\end{itemize}

\begin{figure}[t]
\label{figure2}
\caption{Three levels of analysis of brain and transformer processes}
\includegraphics[width=\linewidth]{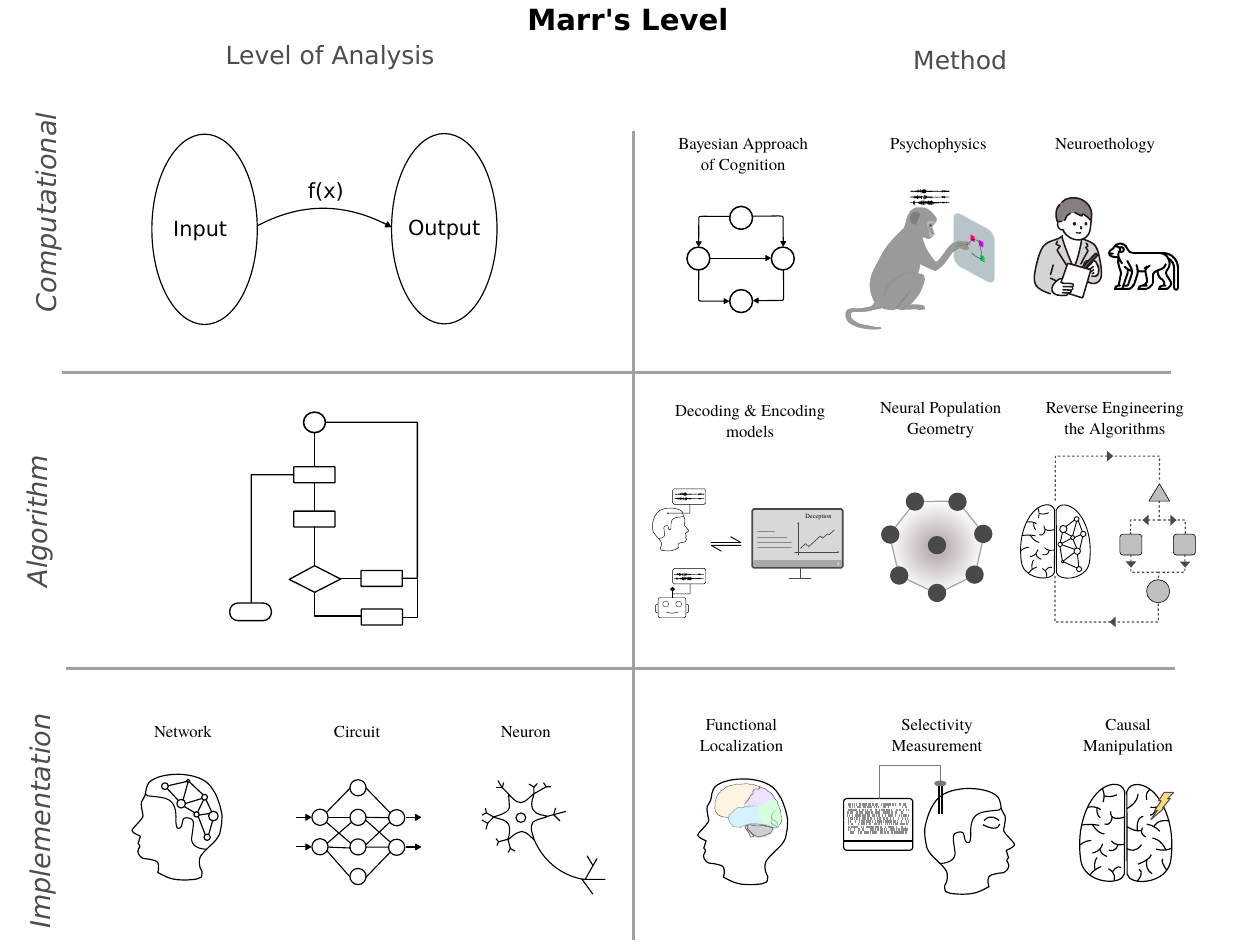}
\centering
\end{figure}

Ultimately, understanding a neural system requires addressing all three of Marr’s levels \citep{bechtel2015non}; however, considering them all simultaneously might be unfeasible within a single study. Thus, every individual study could be explicit about the level(s) of analysis they are trying to address: is the main claim at the computational level (“this model is capable of completing grammatically correct sentences”), algorithmic level (“it first locates the sentence subject, then identifies the number of the subject, and then corresponding number of the verb”), or the implementation level (“here is a set of neurons encoding the number of the subject”)? Circuit studies at the implementation level can result in causal understanding between structure and behavior, but it is the careful theoretical and experimental decomposition of behavior alongside corresponding algorithmic and neural representations that produce a more comprehensive understanding of complex systems \citep{Krakauer2017}.

Framing research in terms of the level of analysis is a standard practice in communities that study biological information processing systems \citep{Griffiths2015RationalUO, niv2016reinforcement, griffiths2012bridging}, and clarifying the level that research is being conducted at can help to explain its impact and implications (for an example, see\citep{rumelhart1985levels} ). Marr’s levels have started to permeate into the discourse amongst machine learning researchers (e.g., \citep{hamrick2020levels,zou2023representation,vilas2024position}); however, the ideas are far from commonplace. 

In this work, we call for interpretability researchers to engage with Marr’s levels when designing and discussing studies. Similar to the study of biological systems, researchers can investigate a phenomenon from different levels, each contributing to our broader understanding of the system and its capabilities. No level alone paints a complete picture; and each level makes a non-redundant contribution to our understanding of information processing\citep{bechtel2015non, cooper2015beyond, Krakauer2017}. 


\subsection{The benefits of multi-level analysis of neural systems}
Marr's levels of analysis is a conceptual framework for studying computational systems. The benefits of multi-level analysis frameworks include:

\begin{itemize}
    \item \textbf{Holistic understanding}. A holistic understanding of cognitive behaviors would require analysis across all three levels \citep{griffiths2023bayes}. In contrast, research at only one or two levels generates an incomplete understanding of the brain and interested behaviors \citep{Krakauer2017}. Each of three levels offers distinctive and non-redundant understanding of neural systems. Missing computational analysis would risk being mislead by wrong understanding of task \citep{bechtel2015non}; analysis at the representation/algorithm level allows us to understand how the task is done by the neural systems without committing to specific realized implementation \citep{sober1999multiple, Krakauer2017}; and the implementation level decpompose and localizes the computations in neural network by mapping them onto physical components \citep{bechtel2010discovering}, which paves the way for establishing causality between internal components to the final output.

    \item \textbf{Understanding Emergence}. Because of emergence, lower-level descriptions say little about neural network behaviors\citep{anderson1972more}. Behaviors and cognition are emergent from the organization of neurons and circuits, having effects that are not apparent in any single neurons \citep{miller2024cognition}, and therefore should be firstly studied at the computation level \citep{Krakauer2017, marr2010vision}. As Marr put it, you should start with aerodynamics if you are interested in bird flight, rather than study the materials of feathers; flocking in birds should be studied at the `group of birds' level, as some behavioral rules such as "steer to the average heading of your neighbors" can never be found by studying an individual bird \citep{Reynolds1987FlocksHA}. 
    
    \item \textbf{Consistency and Constraints}. Once we delineate three levels of analysis, it should be straightforward to see that different levels of analysis should be consistent with each other. Computational level identify problems to be solved in an environment and determines a computational function for them; through algorithmic and representational analysis we understand how information is encoded and transformed when solving the problems, and implementation identifies the physical form to support such function, representations, and algorithms \citep{bechtel2015non}. 
    
    Three levels of analysis also enables us to see the constraints applied across levels. \citet{vilas2024position} termed as "mutual constraints". Taking a top-down approach, a computational explanation identifies how environmental information constrains the functions performed by the systems; it therefore narrows down the search for possible algorithms and implementations \citep{griffiths2023bayes}. With a bottom-up approach, knowledge of the physical system helps to specify what algorithms are implementable, and therefore limit the range of functions possibly performed by the system \citep{bechtel2015non}. 
    \item \textbf{Goals and questions at each level}. One benefit of specifying the level of analysis is to understand what is the goal of the analysis, what questions can be asked, what form the answer should take, and what other analysis can be complementary and therefore necessary for a more complete understanding. 
\end{itemize}

\section{The case of deception}
Deception is a behavior that stimulates research interests from both cognitive and brain sciences \citep{ganis2009cognitive} and AI researchers \citep{hagendorff2024deception, shevlane2023model, williamson2024era}. Because deception is a cognitive behavior of critical legal, moral and social implications, there is strong interest from the brain sciences community in studying deception \citep{karim2010truth}. There is also growing evidence of AI systems exhibiting deceptive behaviors \citep{park2024ai}, such as providing false information to players to win a game \citep{meta2022human, xu2023exploring} or to a human worker for the completion of its task \citep{achiam2023gpt}. More recently, the mechanistic interpretability community has developed interest in reverse engineering the mechanism for deception in language models in order to control this behavior \citep{OlahJermyn2023}. 

What might we gain from a Marr's-levels analysis of AI deception? When investigating the basis of this behavior, we would ask different research questions depending on the level of analysis. 

Computational Level: what is the function of deceptive behavior in the context of a model’s overall behavior? What is considered success vs. failure? What aspects of a model’s input and internal states determine whether it is going to lie or respond truthfully? What is a formal definition of “lying” and “deception” when applied to an AI system and when is it theoretically possible for a model to possess such capabilities?

Algorithmic/Representational Level: what are possible algorithms underlying deceptive behavior and which of these algorithms is realized in a particular AI model that has been shown to exhibit deceptive behavior? How are inputs combined in order to determine whether deception or truth will be used? How does the model represent the true state of the world and the intention to deceive vs respond truthfully? How do these representations combine to determine the output? 

Implementation Level: what are the physical circuits that implement the deception algorithm? Are representations of the true state of the world and representations of the intention to deceive/respond truthfully localized to particular parts of the AI system or distributed throughout? What is the role of individual layers/units/attention heads in supporting deceptive behavior?

A reader familiar with the interpretability literature might recognize that much of the previous research has concentrated on implementation questions, with some work also starting to tackle the algorithmic level. Yet the computational level, which is perhaps primary, remains relatively underexplored. 

We come back to the deception example is Section \ref{deception-part2}, after reviewing techniques that might be most helpful at each level of analysis.

\section{Computational/Functional Level of Analysis}
The computational level of analysis aims to study the information-processing system as a whole and to understand the goal of the system. The key at this level is to emphasize a high level description of what functions a system implements without committing to a particular algorithmic strategy \citep{marr2010vision,griffiths2023bayes,willems2011re}. 

Examples of questions to ask at the computational level:
\begin{itemize}
    \item What is the system capable of? What patterns exist in its behaviors? \citep{ewert1980neuroethology, watson1913psychology}
    \item What are some objectives or computational constraints of the system that shape its functions? \citep{laughlin2003communication, gershman2015computational, mccoy2023embers}
    \item What are the rules governing different functions such as learning and adaptation? \citep{dayan2005theoretical, botvinick2019reinforcement}
    \item Do any general principles in function emerge across many different behaviors (or different species)? \citep{hulse2018cognitive, dewsbury1978comparative}
\end{itemize}

In brain studies, the computational level is frequently tackled with behavioral studies \citet{Tinbergen1963, Krakauer2017}, sometimes accompanied by abstract mathematical models used to summarize a wide range of behavioral findings. Well-characterized behavior gives insights to the computational goals of a neural system and can provide hints as to the algorithms used to achieve those goals\citep{blokpoel2018sculpting}. Computational study also guide the search for algorithms and representations \citep{griffiths2010probabilistic}

Existing interpretability studies often do not intimately engage with the computational level (problems to be solved by the systems \citep{mccoy2023embers}, why they are solving those problems \citep{laughlin2003communication}, environmental constraints \citep{bechtel2009looking}, and, ideal solutions expressed by Bayesian models \citep{Griffiths2015RationalUO}) but rather focus on mapping functions to structures (we call implementation level of analysis). Where it exists, research at the computational level of ANNs usually focuses on the general principles of deep learning systems (e.g., scaling law \citet{kaplan2020scaling, hestness2017deep, seung1992statistical}, memorization \citet{carlini2022quantifying}, hallucination \citet{zhang2023language}, inductive biases \citet{white2021examining}, rather than aligning with the interpretability research agenda that focuses on specific behaviors of interest. We believe a synergy between the two can potentially advance our understanding of neural network behaviors. 

In this section, following an approach from the cognitive science/neuroscience communities that takes computational analysis as the first step into investigating neural systems \citep{marr2010vision}, we highlight the importance of studies at the computation level and the links between different levels. Under this mindset, we advocate an approach that first curates a list of behaviors of interest in the computational system under study (e.g, the mathematics ability of an LLM; see sections 2.2 and 2.4 in the survey paper \citep{anwar2024foundational} for a detailed discussion) and understands basic computational principles behind those behaviors, studying problems, tasks, and goals before they try to identify algorithms for it and locating circuits \citep{Krakauer2017, griffiths2010probabilistic, mccoy2023embers}. 

A key takeaway of this section for the interpretability community is to reflect on the question of whether you are starting at the top or at the bottom. While the classic bottom-up approach of mechanistic interpretability can help generate explanations with high granularity, a top-down approach may lead to more robust and general explanations of model behavior in some instances. As the neuroscientist Horace Barlow wrote in 1961, ``A wing would be a most mystifying structure if one did not know that birds flew'' \citep{barlow1961possible}.



\subsection{Neuroethology}
Ethology is the study of natural animal behavior. Neuroethology is the specific study of behaviors for the purposes of understanding the brain-behavior relationship and how it arises through natural selection \citep{ewert1980neuroethology}. Advances in experimental techniques and methods for monitoring behavior are making it possible to collect large amounts of precise data while animals engage in self-driven and unconstrained behavior \citep{datta2019computational}. Neuroethology tends to focus on innate behaviors (such as foraging) in order to characterize the computational principles of behavior that have emerged through evolution \citep{cisek2019resynthesizing}. This is particularly important to the Marr's levels approach as it has been argued that an evolutionary perspective is necessary when defining the computational level \citep{anderson2015can}.

Tinbergin's four questions, proposed by Niko Tinbergin in 1963 \citep{tinbergen1963aims}, provide a framework for studying ethology that captures the multi-faceted style of explanation one may want in order to `understand' behavior. The questions include: Why is the animal performing the behaviour, i.e. what is its function? Through what historical stages did the behaviour evolve? What mechanism in the animal cause the behaviour to be performed? How does the behaviour develop during the lifetime of an individual? Neuroethology is primarily interested in the neural mechanisms that produce the behavior, but importantly is informed by answers to the other questions as well. While there is not a direct link between Marr's level and the different Tinbergin's questions, work has been done to try to integrate these two frameworks in psychology \citep{al2024levels}. 

\paragraph{Findings}
Neuroethology has been applied to many topic areas. It has been used to identify common principles across vocalization behaviors and their deep evolutionary origins \citep{takahashi2012computational, bass2012shared,leininger2015evolution, benichov2016forebrain}. Characterization of unconstrained spatial exploration in rodents has shown certain reliable behavioral patterns, such as bouts of exploration that allow the animal to explore increasingly farther from its home base before returning each time; such observations have informed the development of path integration models (at the algorithmic level) that could support these behaviors \citep{dudchenko2018neuroethology}. Furthermore, neuroethological studies of decision making across animals have revealed common principles and re-occuring problems. For example, when should an animal leave an area with a depleting food resource in order to search for a new source? It has been found that the computationally optimal solution (leaving an area when the rate of caloric return becomes below average) is the common one across many species \citep{adams2012neuroethology}. Neuroethological studies of social behavior in primates have shown how computations and brain mechanism initially used for non-social purposes have been co-opted and repurposed for social cognition \citep{chang2013neuroethology}, a key feature of evolved systems \citep{cisek2019resynthesizing}. 

\paragraph{Relevance}
Defining `natural' behavior for an ANN is difficult. These models display limited `innate' behavior: while randomly initialized networks can do some useful computations \citep{gallicchio2020deep}, ANNs must learn most of what they are capable of. However, interpretability researchers may still be able to take some inspiration from the idea of exploring unconstrained or typical behavior. For example, for models that are deployed in the real world, large-scale analysis of model behavior on the data distribution it encounters in practice may help identify the most relevant and common behavioral trends, be they positive or problematic, for interpretability studies to focus on. This could include identifying the types of hallucinations most commonly encountered by users or linguistic patterns that commonly reoccur in a model's output.  This would therefore be a more data-driven approach to problem selection compared to researchers choosing on their own what set of behaviors they are interested in explaining. Furthermore, research on AI behaviors frequently goes under the title of "evaluation" or "benchmarking" and focuses on accuracy on hand-picked test datasets. It is possible that taking a more expansive view of model behavior, especially for models like LLMs that can produce complex output, will lead the way for more well-formed interpretablity questions, which is aligned with neuroethology's more holistic approach to animal behavior.

In this direction, Tinbergin's four questions have also been adapted as a framework for understanding machines in \citep{rahwan2019machine}. For AI interpretability the questions can be framed as:
\begin{itemize}
    \item \textbf{Development} By what engineering methods does the machine acquire its behaviors (training datasets, benchmarks, supervised learning, hyperparameters, architecture)? How do its behaviors develop over its "lifetime" ? Examples may include in-context learning or meta-optimization \citep{anwar2024foundational}
    \item \textbf{Mechanism} What internal mechanism causes the machine to behave in certain ways? This question is already the standard focus in mechanistic interpretability.
    \item \textbf{Evolution} What incentives may have shaped the dominant models used in AI? This can include market forces and other social or technical trends and limitations.
    \item \textbf{Function} What purpose does the machine serve or what consequence does it bring? In analogy to an animal's evolutionary fitness, we may ask why certain features of models are selected over others. 
\end{itemize}

 The field of interpretability is witnessing an increasing amount of research papers studying behavioral aspects of language models (and other deep learning systems).\citep{anwar2024foundational} summarizes the observed behavioral phenomena of LLMs by saying the "shape" of language model capabilities appear to be different from human capabilities. The message here is, before understanding internals, we ought to think more about what behaviors and capabilities we are aiming to understand. \citep{mccoy2023embers} takes a more ethological/`evolutionary' approach (which they call teleological) by trying to understand LLM behavior specifically in the context of the main task LLMs are trained on, that of next word prediction. For instance, when considering how an LLM solves a simple cipher, it is best to frame it in the context of looking for tokens with high-probability based on training data, rather than, for example, searching for abstract circuits for "cipher decoding" \citep{mccoy2023embers}. This is and should be different than how we would explore the same questions in humans, which have a much different evolutionary past.

\subsection{Psychophysics}
Unlike neuroethology, psychophysics focuses on the study of very precise--and largely `unnatural' behaviors--in a laboratory setting. The use of tightly controlled, pared-down stimuli and responses is meant to help isolate specific capabilities, limitations, and mechanisms \citep{Krakauer2017, gescheider2013psychophysics, prins2016psychophysics}. In a psychophysics experiment, a subject may, for example, sit in a dark room viewing grayscale images of simple lines or dots on a screen and respond when they perceive a particular stimulus. While psychophysics emerged as a means of measuring how physical stimuli give rise to perception, its methods are also readily applicable to motor and even `cognitive' questions \citep{sanes1984motor, waskom2019designing}. 

\paragraph{Findings}
Many visual behaviors have been thoroughly characterized through psychophysics. For example, in visual search, studies have shown what features of a stimulus will make it `pop out' or not, how search behaviors depend on visual statistics, and how it interacts with both short and long-term memory \citep{wolfe2020visual}. These precise findings put strong constraints on the potential algorithms that could be used by the brain to achieve this behavior. The impacts of visual attention have been similarly characterized and used to explain and predict findings from neurophysiology \citep{carrasco2009visual}. 

In the case of sound localization, behavioral work from nearly 100 years ago established the importance of inter-aural time differences (that is the difference in time it takes for a sound to reach one ear versus the other, ITD) in the sound localization computation \citep{stevens1936localization}. Following this, 
Lloyd Jeffress proposed a place theory of sound localization which hypothesizes that ITD is calculated through slow-conducting nerves and represented by the "place" of a cell in a nerve cell array \citep{Jeffress1948, AshidaCarr2011}. Significant progress was made in anatomical, physiological, and psychophysical research to understand the mechanism of sound localization after this\citep{Grothe2003}. This shows how a precise characterization of how perception depends on inputs can lead to productive theories of the algorithms and implementation used by the brain. 

On the motor side, visuomotor adaptation--whereby execution of a movement under challenging conditions is corrected through trial and error--- was once believed by neuroscientists to occur mainly by adapting an internal model of motor control through error-based learning performed by cerebellar circuitry \citep{therrien2015cerebellar}. Later, precise behavioral studies revealed that additional mechanisms are likely at play, including use-dependent plasticity and reinforcement learning \citep{huang2011rethinking, taylor2014explicit, izawa2011learning, shmuelof2012overcoming}. 

These examples support a top-down approach whereby the first step should be to characterize behaviors and potentially come up with computational/algorithmic hypotheses that may later inspire experiments to identify circuits \citep{Krakauer2017}. 



\paragraph{Relevance} For interpretability research on algorithms and neural mechanisms to be effective, researchers should consider whether the involved model behaviour is thoroughly enough understood for them to be able to uncover said mechanisms. If not, researchers should prioritise analysing and decomposing the behaviour through careful experiments and observation, in order to derive theories that can be used as a tool to understand the internal mechanism of the model. For example, interpretability researchers aim to understand concepts like planning, deception, and situational awareness \citep{OlahJermyn2023}. But are these behaviors well-characterized enough at scale in LLMs at this point in time? A lack of understanding of the actual model's behaviour may pose a risk of researchers anthropomorphising (i.e. assuming the model behaves in a human-like way) which may hinder their ability to properly discover the model's internal mechanisms \citep{salles2020anthropomorphism}. Furthermore, behavioral results from one model may not hold for another \citep{tuli2021convolutional}, and assuming that they do might result in misleading theories. 

One example of a psychophysics-like exploration of model behavior is the study of shape versus texture biases in convolutional neural networks. By producing unnatural images, such as objects with mismatched shapes and textures (i.e. a cat with the texture of an elephant), \citep{geirhos2018imagenet} was able to show how object recognition performance of CNNs relied more on texture than shape, and were able to use these insights to propose a training algorithm to reduce this bias. Later work further explored how this bias may arise from training data, and not from the architecture of the models \citep{hermann2020origins}. 

In general, the psychophyics approach encourages testing of a system at its extremes or under other non-standard conditions that go beyond simple accuracy measures and can be revealing of internal mechanisms. Such an approach has shown up in AI in the cases of, for example, texture biases as mentioned, adversarial images \citep{szegedy2013intriguing} and length generalization in transformers \citep{zhou2024transformers}.

\subsection{Bayesian Approaches to Cognition}

Unlike the above two examples, this line of work does not advocate for a particular type of behavioral experiment but rather puts forth a computational framework that can encapsulate many behavioral findings. Specific, probabilistic models of cognition champion the idea that one can model cognition, from vision and motor control, to language and beyond, with probabilistic models \citep{chater2006probabilistic, l2008bayesian}. This thread of research emphasizes that many aspects of human cognition can be captured by positing that humans build generative models of the world and conduct inference under uncertainty in and over models ~\citep{ullman2020bayesian, tenenbaum2011grow}. This kind of explanation places principal importance on \textit{inductive biases}, which may be helpful in understanding the kinds of problems artificial systems are equipped, or can be equipped, to solve ~\citep{tenenbaum2011grow, lake2017building,  mccoy2023embers}. Probabilistic frameworks allow us to address computation level questions such as: What constraints on learning are necessary \citep{Griffiths2015RationalUO, griffiths2006optimal}?  How do children learn to infer the meanings of new words based on just a few examples \citep{xu2007word}? How do people infer the mental states underlying actions of others \citep{baker2009action}? Computational questions, as a result, should be addressed with computational theories. 

\paragraph{Findings}
Neuroscientists have identified strong empirical support for the idea that humans employ Bayesian strategies that combine both knowledge of the statistical distribution of the tasks (prior knowledge) and sensory uncertainties when performing sensorimotor learning \citep{kording2004bayesian}; similarly, human behavior when combining visual and haptic information is well-explained by a maximum-likelihood model \citep{ernst2002humans}.
More broadly, researchers have successfully applied probabilistic models to account for many aspects of vision \citep{kersten2003bayesian, chater1996reconciling, weiss2002motion, tu2005image}, language \citep{xu2007word, chater2006probabilisticlanguage, goodman2016pragmatic, wong2023word}, navigation \citep{kaelbling1998planning, stankiewicz2006lost}, causal learning and inference \citep{pearl1988probabilistic, gopnik2004theory, lagnado2013causal}, concept learning \citep{tenenbaum1998bayesian, tenenbaum2001generalization}, emotion prediction \citep{houlihan2023emotion},reasoning about other minds ~\citep{baker2009action, baker2017rational, baker2011bayesian, ho2022planning}, how people couple different sensory data to integrate perception with planning \citep{ernst2002humans, clark2013data} and reason about physics ~\citep{ullman2017mind, battaglia2013simulation}. 

Probabilistic models of cognition primarily sit at computational level description however, as with other computational level strategies, they can motivate the search for algorithms which underpin behaviour and ultimately be used to explain the roles of neurons and neural populations \citep{coen2023mouse, vaghi2017compulsivity}. Indeed, researchers have been able to find evidence for the neural implementations and mechanisms that convey probabilistic information \citep{chater2006probabilistic,echeveste2020cortical, aitchison2016hamiltonian}.

\paragraph{Relevance}
Machine learning researchers have already applied probabilistic models to understand the general computational principles of neural networks. For example, a Bayesian framework was proposed to understand in-context learning as a process of using "cues" in the prompts to locate latent learned concept which language models have acquired from pre-training \citep{xie2022explanation}. Such a practice places an important emphasis on \textit{uncertainty} and the study of the inductive biases that arise from the computational task of interest, which has already proved valuable for understanding LLMs~\citep{mccoy2023embers}. This can complement lower-level mechanistic explanations of the same phenomona (e.g. \citep{olsson2022context} offers a mechanistic explanation of in-context learning).
While there are efforts of combining probabilistic models and deep learning models to address the limitations of each \citep{chua2018deep}, even if models are not explicitly probabilistic, their behavior can still be approached with Bayesian methods assuming it is complex enough \citep{jordan2019deterministic}.

\subsection{Limitations and Considerations for the Computational Level}
While it is common for researchers studying the brain to start at the computational level and work down through algorithm and implementation, the question of whether this is the best approach is actively debated \citep{gyorgy2019brain,poeppel2020against}. Specifically, in \cite{gyorgy2019brain}, the author argues that instead of starting with psychological terms like `memory', neuroscientists should focus on the `hardware' first, by documenting features of neurons and their activity and then asking what function they could implement. This bottom-up approach is meant in part to help move hypotheses about brain function away from simple and intuitive ideas derived from folk psychology and towards more complex and surprising mechanisms \citep{gyorgy2019brain}. The debate around which approach is best highlights the ways in which characterizing behavior may inject biases or assumptions into the analysis of a neural system.

An important difference between ANNs and the brain with respect to the computational level is the fact that technically the computational aims of an ANN are known and even specified by the builder in the form of the objective function. However, in biological neural systems, a goal of our study is to reverse engineer something akin to the objective function for a brain region \citep{richards2019deep}. It is clear that the capabilities of large ANNs can be described in ways that go well beyond the explicit task they were trained for (e.g., next word prediction for LLMs), and so it may make sense to still try to break down the behavior of these large models into different computational capacities. This is akin to acknowledging from an evolutionary perspective the only `objective function' of the brain is to support the production of more offspring; however we can still speak of different sub-types of behaviors and abilities.

\section{Algorithmic \& Representational Level of Analysis}
\label{AlgoRep}

At the algorithmic or representational level, researchers aim to figure out the question of "how" neural systems solve computational problems, in terms of algorithmic steps taken and latent representations used. At this level, the exact nature of the physical implementation becomes less relevant as it is abstracted away through an algorithmic description. Researchers operating on this level are concerned with how information is encoded (therefore "representation") and transformed (therefore "algorithm") within the system. 

Examples of questions to ask at this level:
\begin{itemize}
    \item How is the relevant information encoded in neural systems? \citep{quiroga2005invariant, nogueira2023geometry, stringer2019high, sucholutsky2023getting} 
    
    \item How are abstract concepts represented and manipulated in neural systems? \citep{dayan2005theoretical, bernardi2020geometry, johnston2023abstract}

    \item How do neural representations change with learning to achieve computational goals? \citep{yasuda2006supersensitive, wojcik2023learning}

    \item How is information transformed and combined in the neural system to produce intelligent behaviors? \citep{harris2013cortical}

    \item What are some strategies neural systems can utilize to solve their computational problems? \citep{banich2018cognitive, van2023expertise}

\end{itemize}

Many exciting recent advances in mechanistic interpretability aim to address Level 2 questions. These include the use of sparse autoencoders to separate out responses to different input features \citep{bricken2023towards, templeton2024scaling} and representation engineering approaches that track latent neural trajectories when processing different input types \citep{zou2023representation}. Here, we survey a variety of approaches, old and new, that can help make progress in understanding the algorithms and representations employed by a neural system to achieve its computational goals.


\subsection{Reverse-engineering the Algorithm}
Cognitive science has long tried to understand the principles of intelligence common across both minds and machines \citep{simon1980cognitive}. In doing so, much of the field inherently works at the algorithmic level, as it builds frameworks that are not meant to be specific to any one physical implementation. One approach to developing such algorithmic descriptions and frameworks is through a form of normative reverse engineering \citep{zednik2016bayesian}. In this approach, a detailed characterization of behavior is undertaken and combined with existing knowledge from other fields about what algorithms can achieve this behavior. When this existing knowledge is pulled from other domains such as statistical analysis, it has been referred to as the `tools to theories' heuristic \citep{gigerenzer1991tools}. A variety of analyses are then used to look for evidence of the hypothesized algorithm at play.  

\paragraph{Findings}
One prominent example of a reverse-engineered algorithmic explanation is in the domain of reinforcement learning \citep{niv2016reinforcement}. More than a century of behavioral work has clearly demonstrated the ability of animals to learn to modify their behavior based on reward. Further and more quantitative behavioral work laid the groundwork for hypotheses about the underlying algorithms supporting this learning \citep{bush1955stochastic, rescorla1972theory}. Ultimately, the prominent frameworks for understanding the algorithms of reinforcement learning have come from engineering domains. For example, the temporal difference learning algorithm was developed by computer scientists primarily for the purposes of solving complex planning problems \citep{sutton1988learning} but also serves as a successful algorithmic level description of learning in the brain \citep{o2003temporal}. This was further verified by findings at the implementation level demonstrating neural populations that encode reward prediction error \citep{hollerman1998dopamine}. 

\paragraph{Relevance}
One of the challenges of the `reverse engineering' approach is the need to come up with an algorithm that can achieve the complex computational goal of the system. Machine learning models are usually built exactly because an existing algorithmic solution to a problem does not exist, and so researchers can not know what algorithm to look for in the trained network. One area of research, however, wherein machine learning researchers do study simpler tasks with known representational/algorithmic solutions is `grokking'. In grokking studies, overparameterized neural networks are trained to a point where they have perfect generalization accuracy on simple mathematical tasks like modular division \citep{power2022grokking}. Here, looking for the expected mathematical structure in activity patterns has helped characterize the progression of learning \citep{liu2022towards,nanda2023progress}.

While most models trained on complex tasks will not have an obvious algorithmic hypothesis to test, interpretability researchers can continue to try to bootstrap by using what has been learned in simpler settings to inform the search in more complex settings \citep{liu2022omnigrok}. Methods have been developed for identifying where in a complex network parts of a known algorithm are represented \citep{geiger2021causal}, which can also be used to explore the ways in which networks deviate from known algorithms.

\subsection{Decoding models (probes) and encoding models} \label{decoding-encoding}

Often researchers might have specific hypotheses about the features that a system needs to represent to achieve its goal. To test whether a specific neural population represents these features, they can design a mapping between neural responses and the features of interest. If the mapping predicts features of interest (e.g., deceptive vs.~truthful behavior) from neural data, it is called a \emph{decoding model}. If the mapping predicts neural responses from the features of interest, it is called an \emph{encoding model} (\citealp{naselaris2011encoding}); note the difference between this terminology and the traditional use of the terms ''encoder'' and ''decoder'' in machine learning). 

Decoding models have a long history of use in neuroscience and have guided understanding of many different brain regions and systems \citep{abbott1994decoding, pouget2000information, ritchie2019decoding, kriegeskorte2019interpreting}. The main criticisms of this approach, however, concern the potential lack of a causal connection between the decoded features and the system's function \citep{weichwald2015causal,holdgraf2017encoding, kriegeskorte2019interpreting}. This can happen for many reasons: first, the decoder might pick up on features that are simply correlated with the target features; second, constraining the decoder to be linear (as is commonly done in neuroscience) might be overly limiting (but providing no constraints at all is overly permissive as the decoder may come to implement computations not done by downstream brain regions) \citep{ivanova2022beyond}; third, there is no guarantee that the decoded information is actually necessary for performing the task, unless causal manipulations are done in addition to decoding. Another reason for caution, particularly when trying to create an algorithmic level description, is that simply knowing what information is where doesn't tell you what is being done with that information. As said in \cite{kriegeskorte2019interpreting}, ``Decoding reveals the products, not the process of brain computation''. The ultimate goal at the algorithmic level is to understand not just how information is represented but also how it is processed and transformed. 

A popular alternative to standard linear decoders are information-theoretic approaches \citep{quiroga2009extracting, timme2018tutorial}. These tools can be used to trace information flow from one brain region to the next, quantify the amount of redundant information between regions, and measure the rate of information loss. 

Encoding models are also ubiquitously used by neuroscientists \citep{kriegeskorte2019interpreting}. They quantify the amount of variance in neural responses that can be explained by a particular set of features \citep{dupre2024voxelwise}. They are also useful for mapping between two different systems for comparison, e.g. a brain region and a layer of an artificial NN (which represents a set of pre-processed features) \citep{yamins2016using, schrimpf2018brain, antonello2023scaling}. 

\paragraph{Findings} 
In neuroimaging, multi-voxel pattern analysis  (MVPA) has been used to decode task information from patterns of BOLD values \citep{norman2006beyond, mahmoudi2012multivoxel}; this was designed to move fMRI analysis beyond simple questions of where activity goes up or down to questions of what information is present in this activity. MVPA studies have been used, for example, to demonstrate how spatial information is encoded in human hippocampus \citep{kim2017multivoxel}, to predict symptom severity for certain disorders \citep{coutanche2011multi}, and to verify the frontoparietal network's role in domain-general computations \citep{woolgar2016coding}.  

Decoders have also been used to identify the time course of information processing. Specifically, by measuring how well a decoder trained on activity from one time point in an experiment's trial generalizes to other time points, researchers can learn about the stability of the neural code within a brain a region and also how information gets passed between regions \citep{grootswagers2017decoding,contini2017decoding,dirani2023time}. This has revealed, for example, how the computations supporting object recognition play out dynamically in the visual system \citep{dirani2023time}.

Recent decoding models in neuroscience (which increasingly rely on features from trained ANNs) have also led to impressive "mind-reading" studies in the domains of vision \citep{kay2008identifying, Scotti2023Reconstructing} and language \citep{mitchell2008predicting, tang2023semantic}. In addition to the wow-factor of the ability to decode the contents of a person's mind, they bring important scientific insights, such as the fact that mental imagery re-creates detailed perceptual images \citep{naselaris2015voxel} or that semantic representations evoked by language are broadly distributed in the brain \citep{tang2023semantic}.

Encoding models focus on predicting activity in a given brain region (in a way that generalizes across stimuli) and have a long history in the study of sensory systems. For example, `linear-nonlinear' models have been used to capture the computations performed by the retina \citep{shi2019functional}; specifically, these models take in an image and apply filters followed by a nonlinear activation function as means of predicting the response of specific retinal cells. When done successfully, encoding models provide a compact description of the computational abilities of a portion of the sensory system, which makes them amenable to further theoretical analysis and computational modeling. Larger encoding models of the entire ventral visual stream have also been built using task-trained convolutional neural networks \citep{yamins2016using}. Example findings outside of sensory systems include the fact that semantic and syntactic features jointly predict responses in the same brain regions \citep{caucheteux2021disentangling, kauf2024lexical} and that semantic information evoked by a narrative activates many distributed brain regions, often with a highly selective semantic profile (e.g., family-related words or color-related words; \citep{huth2016natural}.

\paragraph{Relevance}
Decoder models are known as \emph{probes} in the interpretability literature \citep{belinkov-2022-probing}. Probes are commonly used for identifying and even manipulating activity in ANNs. They have uncovered the presence of hierarchical syntactic structure in language models \citep{hewitt2019structural}, showed that high-level semantic features are easiest to decode linearly in the middle layers of a transformer, as opposed to early or late layers \citep{tenney-etal-2019-bert}, and uncovered latent representations of syntactic parses for ambiguous sentences (``I saw the boy and the girl was/were tall''), which could then be causally manipulated to alter a model's prediction \citep{tucker-etal-2022-syntax}. The ability to perform precise causal manipulations on ANNs lets interpretability researchers avoid some of the pitfalls that neuroscientists face when trying to interpret decoder results. Some of the previously identified issues with decoder model interpretation have however been encountered in ANNs as well, including the probe picking up on features that are correlated with the target feature \citep{hewitt-liang-2019-designing} and the problem of choosing the right level of complexity for the probe \citep{pimentel2020information}.  

Information-theoretic approaches are also used in interpretability research \citep{voita-titov-2020-information} and can help overcome the shortcomings of traditional linear probes by reducing reliance on linear decodability. An alternative approach would be to devise a probing classifier that would resemble the real circuit responsible for transmitting information from one neural network component to the next, which is possible in ANNs because the model architecture and weights are explicitly known. 

Encoding models are comparatively absent in the interpretability literature. Yet they hold substantial promise for identifying gaps in our understanding of ANN activity. Specifically, by building encoding models based on features that are known to be present in a given layer's activity, researchers can identify the amount of variance that has \textit{not yet} been explained by existing feature sets \citep{ivanova2021probing}.

\subsection{Neural Population Geometry}
Neural population geometry studies the arrangement of high-dimensional neural representations during complex cognitive tasks using mathematical and computational tools \citep{chung2021neural}. The goal is to understand how information is embedded and processed in neural population activity and how these properties reflect the algorithm implemented by the population. The main benefit of analyzing the geometry of neural representations is the ability to abstract away from the specific architecture of a given neural system with the goal of comparing their representational solutions to a given computational problem. Often these approaches data-driven rather than hypothesis-driven (in contrast to the approaches above).

Geometry is typically defined by how the neural responses to different inputs or conditions relate to each other in neural activity space \citep{kriegeskorte2013representational}. This geometry can be thought of as an emergent property of individual cell responses, but it is inherently defined at the population level \citep{kriegeskorte2021neural}. The way in which neural activity is organized puts constraints on how the information it encodes can be read out and transformed by downstream networks, which makes it important for understanding task performance \citep{chung2021neural}. 

A variety of methods can be used to study different aspects of neural geometry \citep{sucholutsky2023getting}. \emph{Representational similarity analysis} measures the correlation between neural responses to a set of inputs or states; this indicates which features most affect this neural population's response patterns. Variants of this approach aim to relax the strict correlation measurement by fitting different weights to different neurons and/or creating linear combinations of original features, thus effectively performing affine transformations on  the neural activation space \citep{khaligh2017fixed, khaligh2014deep}. 

\emph{Dimensionality reduction} is a common strategy to project neural responses into a low-dimensional space and then track their clustering patterns for different inputs, or the evolution of response trajectories over time \citep{cunningham2014dimensionality}. To elicit meaningful latent dimensions during dimensionality reduction, neuroscientists have used a variety of techniques, from the standard principal component analysis, PCA \citep{sohn2019bayesian} to independent component analysis, ICA \citep{norman2015distinct} to bespoke adaptations like demixed PCA \citep{brendel2011demixed, kobak2016demixed}, targeted dimensionality reduction \citep{aoi2020prefrontal, aoi2018model}, and sparse component analysis \citet{zimnik2024identifying}. The choice of a particular dimensionality reduction method relies on a set of assumptions about the latent factors and the relationship between them, which need to be carefully acknowledged in each study. 

Finally, \citep{cohen2020separability} is more explicitly focused on the separability of neural representations and the ability of a downstream system to linearly read out relevant information \citep{chung2018classification, cohen2020separability}.

As with decoder analyses, causal claims cannot be made directly from representational geometry metrics. Rather, these analyses are useful for generating and testing hypotheses about algorithms that could be at play in neural populations.

\paragraph{Findings}
Neural population geometry research can reveal how population activity dynamics support complex functions, such as perception \citep{sohn2019bayesian, chung2021neural}, motor control (e.g. reaching \citep{churchland2012neural, michaels2016neural}, motor sequencing \citep{russo2020neural}), and cognition (e.g. memory \citep{rajan2016recurrent} and context-dependent tasks \citep{rigotti2010internal, mante2013context, bernardi2020geometry}).

Representational similarity analysis (RSA) has also been used to show how neural populations encode task relevant information and how this differs across brain regions. For example, in \citep{samborska2022complementary} the authors use RSA to determine the extent to which prefrontal cortex and hippocampus encode either abstract task information or the physical nature of the necessary response. They find that prefrontal cortex captures task-general information, and the hippocampus maps this to the current specific task scenario.

\citet{aoi2020prefrontal} used targeted dimensionality reduction to get insights into the dynamic nature of neural representations within the prefrontal cortex (PFC). TDR-based analysis reveals that sensory stimuli, decision outcomes, and task context are often encoded within multiple, distinct subspaces. Interestingly, these representations commonly display an initial linear accumulation phase followed by a rotational phase with more complex dynamics. TDR has also highlighted differences in encoding properties between individuals, shedding light on how varying neural representations might relate to differences in perceptual abilities.

In some cases, the decoding classifier approach from section \ref{decoding-encoding} can also be used to shed light on the underlying representational geometry of a neural space. For example, in \cite{bernardi2020geometry}, a classifier probe was trained to distinguish between different task conditions, with the hypothesis that a particular geometric arrangement of the underlying neural space would make it possible for the probe to generalize to a different set of task conditions. The authors find that different task parameters are encoded in separate subspaces, allowing for multiple different flexible readouts of different task-relevant information.

\paragraph{Relevance} As in neuroscience, analyzing the geometry of neural representations is a powerful approach in interpretability. \citet{balestriero2023characterizing} proposed a geometric characterization of large language models that can extract interpretable features sufficient for solving tasks like toxicity detection, domain inference, and characterizing types of toxicity. However, in some ways this form of geometric analysis has been present in natural language processing for much longer; it can be seen, for example, in analysis of word2vec embeddings showing how directions in embedding space represent conceptual relationships (such as countries and their capital cities) \citep{mikolov2013distributed}. RSA has also been commonly used to compare networks and network layers in the ML literature \citep{kornblith2019similarity}. 

Interpretability researchers have used dimensionality reduction techniques, such as t-SNE, in a variety of ways to help summarize the activity of large populations of units \citep{narayanan2022challenges,geiger2021causal}. PCA has also been used to show how vision models producing the same behaviors can do so with different underlying mechanisms \citep{lindsay2022bio}. Some of the techniques developed in neuroscience, however, are more explicitly designed to look for dimensions that represent pre-determined types of information. The interpretability community may benefit from this more hypothesis-driven exploration into population activity. For example, these supervised dimensionality reduction techniques could be used to separate syntactic information from semantics or identify traits like deceit across many examples in LLMs. The emphasis on disentangling representations aligns with the goal of interpretability, potentially making model decision-making processes more transparent and controllable.

Sparse autoencoders \citep{bricken2023towards}, which have recently become popular for dealing with polysemanticity, are in some ways the opposite of a dimensionality reduction technique insofar as they aim to find a set of represented concepts that is \emph{larger} than the dimensionality of the population. However, they share with some dimensionality reduction techniques in neuroscience the goal of identifying the latent factors that are driving activity patterns. In neuroscience these latent factors are assumed to be fewer than the number of neurons used to represent them \citep{zimnik2024identifying}, but that assumption can be relaxed.

Given that representational geometry approaches are agnostic to many details of how neural units in a population work, they are well-suited for studying and comparing a wide range of biological and artificial neural networks \citep{chung2021neural}. Identifying the relevant geometries for implementing different types of algorithms is thus a ripe topic area for advancing both neuroscience and AI.

\subsection{Limitations and Considerations for the Algorithmic/Representational Level}
The algorithmic level can be heavily constrained by the implementation and computational levels. By many respects, the algorithmic/representational level is the least constrained and therefore most difficult level of description to pin down. Insofar as an algorithm is meant to describe a process that could be implemented by many different physical implementations, it is not itself physical. While the algorithm is meant to be independent of the implementation, for any given information processing system, the physical implementation will put constraints on what algorithms it can realize \citep{bechtel2015non}. This, combined with multiple realizability, make it difficult, though not necessarily impossible, to infer algorithms from the behavior of a system alone (as discussed below careful consideration of behavior can be crucial in narrowing the space of possible algorithms). Similarly, looking directly at neural activity and representations alone is rarely sufficient to reveal the higher level organization that implements algorithms either \citep{jonas2017could}. For these reasons, some researchers have emphasized that algorithms are often better inferred through measurements of both behavior and neural activity \citep{love2015algorithmic}. 

\section{Implementation Level of Analysis}
At the implementation level, researchers aim to attribute mechanistic functions to individual and independent parts of the system. In brains, system parts range from sub-cellular components, to individual neurons and their connections, up through neural sub-types and full populations; if we cast our representation even more abstract, brain areas could also be viewed as a functional part. In ANNs, system parts might include neural units, attention heads, weights, and/or network layers. A key goal of implementation-level analysis is to identify a mapping between individual components and their function by describing how the components implement the specified algorithm. 

Examples of questions to ask at the implementation level:
\begin{itemize}
    \item Can a particular function be localized to certain components of the brain/ANN (where component refers to specific parts such as a neurons or a attention head) \citep{olsson2022context, rehman2019neuroanatomy, anand2012hippocampus, huff2018neuroanatomy}?
    \item Does a given component selectively respond to certain kinds of inputs and/or tasks? \citep{goh2021multimodal, quiroga2005invariant}
    \item Does damaging this component affect a specific function but leave others intact \citep{wang2023interpretability, dronkers2007paul}?
\end{itemize}   
This level of analysis is often an intuitive starting point to examine the structure-function relationships of an information processing system. As such, much of the neuroscience research throughout the 20th century and much of the interpretability research today focuses primarily on the implementation level.

\subsection{Function Localization}
\label{func_loc}
In this approach, researchers start with a function of interest and aim to identify a set of structural components that are engaged in that function. To do so, they create an experimental condition that targets a particular function and contrast it with conditions that do not involve that function (e.g., viewing pictures of faces vs. objects is a possible contrast for identifying components engaged in face processing; \citealp{kanwisher1997fusiform}). Then researchers compare responses to all the conditions across all structural units and identify those that are preferentially recruited during the task of interest. 

However, we should be aware that this type of functional localization is merely a correlation and does not prove a causal role of these brain regions in a given task. For example, in the case of deception, neuroscientists have identified cortical areas such as regions of the prefrontal cortex that are activated during deceptive behavior \citep{abe2007deceiving}. It is currently debated, however, if these areas are causally important for deception or if activating them reliably alters deceptive behavior \citep{robertson2003studies, luber2009non}.   

If a particular function consistently recruits only a sub-component of the system, that system might be said to exhibit some amount of \emph{functional specialization}. If the functional specialization is very pronounced, the system might be considered \emph{modular} \citep{grossberg2000complementary, sporns2016modular, achterberg2023spatially}.

\label{Modularity}
\textbf{Modularity} refers to the idea that the function of a system can be broken down into separable sub-components; these sub-components are usually mapped on to physically-distinct parts of the system. Modularity research has probably gained a larger audience in the neuroscience community relative to the interpretability community. There are both theoretical and empirical motivations for researchers to investigate modularity. Theoretically, Herbert Simon framed the idea of "near-decomposability": the claim that having inter-connected modules allows complex systems to adapt and evolve quickly in response to environmental changes without affecting other modules \citep{MeunierLambiotteBullmore2010, simon2018near}. It is also claimed that modular architectures are naturally selected via evolution because they are able to more easily optimise the usage of energetic resources \citep{achard2007efficiency,bullmore2012economy} while facilitating specialisation of effectual function within the organism’s evolutionary niche \citep{kashtan2005spontaneous,clune2013evolutionary}.

Functional specialization/modularity has been heavily-studied within neuroscience \citep{mahon2011specialization}. Historically, lesion studies, and more currently fMRI studies, have been used to make claims about the role of specific brain regions in specific tasks (see Section \ref{func_loc}). These approaches assume the brain is modular \citep{fodor1983modularity}, i.e. that different brain regions function as different modules that implement specific computations or functions. Studies have also tried to describe the functions played by different neuronal sub-types \footnote{Note, the criticism of this view is best put by \citep{karmiloff1994beyond}}.

\paragraph{Findings} Whole-brain function localization studies in humans have isolated multiple brain regions and/or networks that are associated with specific functions. Areas of brain cortex are subdivided into perceptual areas (visual, auditory, somatosensory, etc.), motor areas, and association areas \citep{buckner2013evolution}. In addition to this broad organizational pattern, there is a more fine-grained association between certain brain regions and certain functions: for instance, face processing selectively recruits the fusiform face area \citep{kanwisher1997fusiform}, language processing relies on a set of specialized regions called the language network \citep{fedorenko2011functional}, and theory of mind reasoning relies on an area in the temporo-parietal junction \citep{saxe2003people}. It should be noted that some brain regions do seem to be inherently multi-functional, which necessitates looking beyond simple functional localisation to characterise the role of these areas \citep{duncan2001adaptive, duncan2010multiple}. 

\paragraph{Relevance}
As we wrote previously in Section \ref{Modularity}, functional-localization research in ANNs, where researchers identify areas in neural networks specialized for certain tasks, is at its infancy in deep learning. Deep learning researchers have preliminarily identified regions of neural networks that are associated with particular computational roles. For example, earlier layers of convolutional neural networks (CNNs) have been shown to learn low-level visual features like edges and textures, while later layers capture more complex and abstract features \citep{zeiler2014visualizing,olah2020zoom}. Language models similarly seem to process information in stages, similar to more classical natural language processing pipelines \citep{tenney2019bert}. Probing classifiers trained on intermediate representations from different layers can reveal how task-relevant information is distributed across the network \citep{alain2016understanding,belinkov2022probing}. The mechanisms for in-context learning during chain of thought reasoning abruptly emerge in the later half of large language models (LLMs), providing evidence for functional specialization in neural networks \citep{dutta2024think}.  

Furthermore, there have been two broad approaches used to instantiate modularity within ANNs to aid interpretability \citep{rauker2022toward}. The first approach has been to \textit{specify} modularity directly into the neural network architecture. This is typically done by training independent neural networks, each serving as a module, operating on separate inputs that each contribute to some sub-component of an overall task. By imposing structural modularity on the architecture in this way, it is possible to better control how features, and sub-problems, are represented and utilised in a relatively top-down fashion. The second approach, somewhat more congruent with neurobiological principles, has been to \textit{encourage} modularity as a result of learning in neural networks. This can be done by providing additional constraints to the neural network to facilitate the self-organisation of an internal modular structure (see \citep{amer2019review} for a survey of techniques). For example, one exciting direction has been to provide networks with a physical geometry in three-dimensional space to understand how, like the brain, neural networks can efficiently organise their computations in space \citep{achterberg2023spatially,liu2023growing, margalit2024unifying}. Another is to observe how branches (which are topologically similar to modules) become specialised in the absence of any branch-specific design rules. That is, neural network layers, through training become separated from each other temporarily before later integrating information \citep{voss2021branch}. This is indicative of how principles of modularity can be used to probe network specialization.

\subsection{Measurement of Neural Selectivity} \label{method:neuronproperties}

A classic approach in neuroscience is to examine the response properties of a single structural component. The method goes as follows: take a component of interest (neuron, population, brain region) and carefully map out the space of inputs it responds to. For discrete features, the goal is to identify a set of features to which a component responds over and above the rest (selectively) or, in contrast, to which its response is suppressed. For continuous features, the goal is to construct a \emph{tuning curve} \citep{butts2006tuning}, i.e., a mapping between feature magnitude and the component's response magnitude \footnote{Note that this is not the same usage of the term `tuning curve' as used in the machine learning literature related to hyperparameter tuning \citep{lourie2023show})}. 

The measurement of neural selectivity commonly follows function localization: once a function is shown to recruit a particular system component, a natural question is: what specific features of the input might it respond to? However, it is possible to select a component of interest for other reasons, so the two methods do not have to be used together.

\paragraph{Findings} Hubel and Wiesel famously received a Nobel prize for discovering edge detector neurons in the visual cortex \citep{hubel1962receptive}. Another Nobel prize was awarded to researchers who characterized the navigation circuit in the hippocampus, including place cells \citep{okeefe1978hippocampus} and grid cells \citep{hafting2005microstructure}. In all of these cases, the goal is to take a unit of interest (in this case, a neuron) and carefully map out its selectivity, identifying which features modulate its activity.

\paragraph{Relevance} There is a large and growing interpretability literature cataloging the selectivity of components in neural networks. Researchers have found individual neurons selectively responding to specific objects \citep{bau2020units}, curves at particular orientations \citep{cammarata2020curve}, high frequency boundaries \citep{schubert2021high}, sentiment \citep{radford2017learning,donnelly2019interpretability}, syntax \citep{lakretz2019emergence}, knowledge \citep{dai2021knowledge}, tokenization \citep{elhage2022solu}, contexts \citep{gurnee2023finding,bills2023language}, position \citep{voita2023neurons}, spatiotemporal features \citep{gurnee2023language}, toxicity \citep{lee2024mechanistic}, and more. Similarly, attention heads have been found to specialize in positional and syntactic processing \citep{voita2019analyzing}, grammatical relations \citep{clark2019does}, induction \citep{olsson2022context}, and copy suppression \citep{mcdougall2023copy}.

Selectivity analysis has been so commonly used in both neuroscience and interpretability that many specific parallel concepts have emerged. For example: 

\begin{itemize}

\item \textbf{Grandmother neurons} A lighthearted example in neuroscience refers to the so-called \emph{grandmother neuron}, a hypothetical individual neuron that responds when a person is perceiving or thinking about one's grandmother \citep{bowers2009biological}. In this hypothetical example, there is a direct link between a physical structure (a single neuron) and a cognitive function (representing one's grandmother). Such grandmother neurons indeed have been discovered in some regions of the brain (e.g., the `Halle Berry' neuron that responds selectively to images or the written name of the actress Halle Berry; \citealp{quiroga2005invariant}) and in some ANNs (e.g., a sentiment neuron; \citealt{radford2017learning}; or an airplane detector neuron \citealt{bau2020understanding}). However, in both brains and ANNs, the function of individual components is often much harder to understand, likely due to the lack of a 1:1 mapping between a physical component of a system and a single nameable feature. 

\item \textbf{Mixed selectivity / Polysemanticity} One mechanistic interpretability direction that has recently gained prominence is the study of \emph{polysemanticity}, a phenomenon where a single unit in an ANN responds to multiple features \citep{elhage2022toy}---in contrast to a hypothetical grandmother neuron. A similar notion has long been studied in neuroscience under the name of \emph{mixed selectivity} \citep{fusi2016neurons}. Theoretical discussions in neuroscience have shown how nonlinear mixed selectivity leads to population representations that are high dimensional, flexible, and expressive \citep{tye2024mixed, averbeck2006neural, johnston2020nonlinear}. However, from the view of the interpretability community, having such polysemantic neurons makes interpretability and localization more challenging. The emerging view from interpretability research is to think of populations with polysemantic neurons as being in `superposition', that is, that they are representing more features than they have dimensions \citep{elhage2022toy}. A priority of mechanistic interpretability has thus become to develop tools to take representations out of superposition \citep{bricken2023towards}. Mixed selectivity and superposition present a very interesting case study for a topic of interest to both neuroscience and interpretability, where the unique goals and assumptions of each field can help to develop a multifaceted image of a complex phenomenon found across intelligent neural systems. Neuroscience might benefit from the use of sparse autoencoders and other interpretability techniques for revealing individual features\citep{cunningham2023sparse, bricken2023towards}; interpretability may benefit from neuroscience's approach of identifying the benefits of polysemanticity in creating a flexible system \footnote{In \ref{AlgoRep} we cover population-level coding methods to inform algorithmic descriptions}.
\end{itemize}

\subsection{Causal Manipulation of Individual System Components}

Once we establish a link between a system component and a particular function, the question is: is the component necessary for that function? For both biological and artificial NNs, this question is usually answered with causal intervention experiments.

The simplest causal approach is to test the effect of a component's malfunction. In neuroscience, one can test what happens when a component is permanently inactive due to naturally occurring brain damage (in humans) or experimenter-induced lesions (in animal models). A researcher can also inactivate a component temporarily, through techniques such as transcranial magnetic stimulation (TMS) in humans \citep{hallett2007transcranial,bergmann2021inferring} or optogenetic inactivation in animals \citep{acker2016fef}. For instance, neuroscientists applied TMS to study the context-agnostic  propensity to deception and found that brain simulation can influence spontaneous choice of deceptive behavior \citep{karim2010truth}. In general, care must be taken when interpreting results from manipulation studies in the brain, as unobserved confounding factors can influence the outcomes \citep{bergmann2021inferring}.

\emph{Dissociation logic} involves comparing the effects of damage in different system components on different functions. For example, if damaging region A impairs function X, while damaging region B impairs function Y, it suggests that region A is crucial for X and region B is crucial for Y. However, a single dissociation (showing one region affects one function) is not always conclusive. If both functions stem from the same underlying process, damage to either region might impact both. A double dissociation is stronger evidence. It shows that damaging region A specifically affects function X but not Y, while damaging region B specifically affects function Y but not X. This suggests a clear separation of function between the two regions, which enables breaking down a system into independent and easier-to-understand pieces \citep{riddoch1983effect, shallice1988neuropsychology}.

A more sophisticated version of this analysis is to replace internal representations from one input with that of another. In neuroscience, this approach is often technically infeasible (although some attempts have been made \citep{sur1988experimentally, josselyn2015finding}), but it is gaining prominence in ANN interpretability research \citep{meng2022locating, geiger2021causal}. Successful representation swapping can show that altering representations in a particular system component is sufficient for altering the system's behavior \citep{zou2023representation}, although it does not show that it is necessary (i.e., the same behavior can potentially come about by other means). It is common in neuroscience to search for neural populations that are both necessary and sufficient for a specific behavior (though the value of this approach is debated \citep{gomez2017causal}). 

\paragraph{Findings}. Evidence from people with brain damage shows a causal dissociation between language and many other aspects of cognition, such as navigation, reasoning, math, and theory of mind \citep{fedorenko2016language, woolgar2018fluid}. In rodents, optogenetic manipulations that can selectively activate or suppress individual neurons have identified the neural basis for specific memories (typically, contexts that elicit a fear response; \citet{josselyn2020memory}), suggesting a presence of a sparse neuronal code that mediates the mapping between specific context and specific behavioral responses. Lesioning or optogenetic silencing has also helped to identify circuits for goal-directed spatial navigation \citet{ito2015prefrontal}.

\paragraph{Relevance} Such causal interventions are also popular experiments in interpretability. Depending on the empirical context, an activation is replaced with zero, the mean over a large distribution, the identity with additive Gaussian noise \citep{meng2022locating}, or a completely different sample \citep{geiger2023finding,zhang2023towards} while measuring the difference of some metric of interest compared to an unmodified forward pass. These causal intervention experiments have been used to find neural circuits in language models that implement indirect object identification \citep{wang2023interpretability}, multiple choice question answering \citep{lieberum2023does}, entity binding \citep{feng2023language}, false context following \citep{halawi2023overthinking}, and more \citep{zhang2023towards}, .

\subsection{Limitations and Considerations for the Implementation Level}
One risk at the implementation level is that research can be overly influenced by the tools that are most easily available at the time (for example, neuroscientists focused on single cell activity in response to simple stimuli as this is what could be measured in early neurophyisological experiments). Further, it can be difficult to know which implementation-level details are worth focusing on, without a guiding understanding based on knowledge of the algorithm being implemented -- and focusing on some details can be misleading. For instance, in complex systems, simple correlations between parts and behaviors can paint an inaccurate picture of the larger causal relationships in the system \citep{wagner1999causality, gomez2017causal}. Relatedly, studies at the implementation level frequently take a reductionist approach in order to characterize the role of a specific sub-component, but this role may not be constant across different tasks and contexts \citep{Krakauer2017}.  

\section{The Case of Deception, Revisited} \label{deception-part2}

Armed with the arsenal of methods at different levels of analysis, how might we revisit the study of deception?

At the computational level, we would define the necessary criteria that LLM behavior should meet to be classified as deception; develop detailed benchmarks to measure deceptive behaviors in a wide range of conditions (see \citet{chern2024behonest, sharma2024towards} for related work); and conduct comprehensive behavioral assessments of LLM deceptive behavior while systematically varying both input properties (e.g., prompting strategy, usage of specific words) and model properties (e.g., makeup of the training data, finetuning tasks, model architecture). Based on these assessments, we can develop theoretical models (Bayesian or otherwise) that can predict whether a given LLM would exhibit deceptive behavior given a particular input and its current internal state. Analysis at this level alone can yield useful insights for adjusting training and/or inference procedures to control LLM deception.

At the algorithmic/representational level, we would examine the representational spaces across LLM layers and test which of these spaces systematically separate instances of deception from instances of honesty. We could then trace how specific inputs (e.g., deception triggers; \citet{hubinger2024sleeper}) might alter the latent space trajectory of an LLM, such that the generated tokens deviate from the ground truth trajectory to a deception trajectory. The use of high-level representational trajectories can help generalize over many specific word choices to a more general case of deception. A representation-level analysis can also highlight whether a model is using a single algorithm for different deception cases or whether deception behavior is supported by multiple diverse algorithms. 

At the implementational level, we would trace the circuits that might selectively activate during deception behavior and/or causally contribute to that behavior (see \citet{li2024inference} for an example). The circuits might be responsible for tracking different kinds of deception-relevant information, such as the true state of the world, the agent intention, the user's current and predicted beliefs, etc; a systematic understanding of information that should be traced down in the circuit-level can come from the understanding of behavior and algorithms proposed at the previous two levels; conversely, the algorithm can be adjusted based on the insights gleaned from implementation analyses. 

Overall, breaking down an analysis of a cognitive ability into Marr's levels can provide a clear framework for rigorously studying this system.

\section{Discussion}

Ultimately, the goal of interpretability is in many ways aligned with the goal of neuroscience and cognitive science: to understand how the configuration of neurons in a system gives rise to behavior and to decompose the model and its behaviors into individual components such that we can manipulate, fix, or predict system behaviors by manipulating components. Here we discussed a common organizing framework used by scientists who study the brain and mind: Marr's levels. These levels are meant to help simplify the problem of studying complex neural systems by delineating between different types of questions one could ask and the form the answers could take. We hope this framework, along with the specific examples of neuroscience findings at each level, will provide inspiration and fodder to the interpretability community as they tackle similarly complex problems. 

A proper understanding of AI systems has practical implications. Interpretability researchers are keen to understand emergent capabilities such as in-context learning, jailbreaking, and planning \citep{OlahJermyn2023}; alignment researchers are interested in detecting deception in models, and the broader ML community is keen to use interpretability tools to achieve other desiderata of ML systems, such as fairness, privacy, reliability, and trust \citep{doshivelez2017rigorous}, and remove "bad capabilities" such as hallucination and bias \citep{meng2022locating, zou2023representation}. It is of vast importance to understand their mechanism, but to understand machine behaviors we have to in the end form computational theories that explain and predict behaviors \citep{griffiths2023bayes, xie2022explanation, von2023transformers}, and follow these computational descriptions up with algorithmic and implementation level understanding.

\subsection{What is necessary for understanding?}
What counts as understanding of the brain and how success at understanding is measured are debated topics amongst neuroscientists and cognitive scientists \citep{lazebnik2002can, jonas2017could, richards2019deep, lillicrap2019does,lisman2015challenge, thompson2021forms, craver2007explaining, lindsay2023testing}. Some common themes are a focus on prediction (e.g. predicting neural activity or behavior), control (manipulating components of the brain and seeing expected outcomes), or simplicity (a description of the brain's behavior that is condensed yet still productive in some way). 

Insofar as interpretability is part of the applied engineering discipline of AI, its aim is to build systems that work. Therefore, understanding in this field is perhaps best measured as an enhanced ability to fix existing systems or train better ones. Not all demonstrations of a causal relationship between internal model components and behavior will lead to a robust ability to edit or fix a model. This is due to several reasons including:
\begin{enumerate}
    \item \textbf{Self-Repair Phenomenon} Manipulations of internal components may not introduce the expected fixes due to the system's ability to self-repair \citep{mcgrath2023hydra}.
    \item \textbf{Multiple Realizability} This concept suggests that there are multiple mechanisms capable of performing the same function \citep{sober1999multiple, Krakauer2017}. For example, different mechanisms can perform in-context learning \citep{olsson2022context}. Therefore, identifying a causal relationship between parts of a model and behavior for one set of inputs or in one model may not generalize to other situations.
\end{enumerate}

Additionally, while we may identify the internal causal mechanisms, we often lack understanding of why and how these mechanisms develop. The "in-context learning" paper \citep{olsson2022context} argues that the accumulation of induction heads explains the majority of in-context learning capabilities, but it does not address why induction heads accumulate during the training stage. For example, is this due to architecture or hyper-parameters? Without answering this, it would be difficult to introduce effective edits to the systems or build other systems that implement similar algorithms.

We hope the concepts we have introduced in this article may help develop a fuller understanding of neural networks that admits more avenues for edits and fixes. For example, by reverse engineering the \textit{algorithm} that a high-performing model implements, we may be able to devise different \textit{implementations} for it that are better by some metric, e.g., more parameter efficient. Or a better understanding of what \textit{algorithms} achieve specific \textit{computational} goals in a model could make prediction of model behavior simpler and more robust. 

\subsection{Benefits and limitations of a Marr's Levels approach}
The study of the brain has been heavily influenced both explicitly and implicitly by the framework for understanding laid out by Marr \citep{peebles2015thirty,bechtel2015non, hardcastle2015marr, johnson2017marr, baggio2015logic, eliasmith2015marr, mitchell2006mentalizing}. Having Marr's levels as a principle for scaffolding and synthesizing research has helped organize the study of a very complex subject. Attempting to explain mental phenomena without acknowledgement of these separate levels can lead to confusion \citep{verdejo2011levels, rumelhart1985levels}.

Despite its success, issues can still arise when trying to use Marr's levels to understand a neural system. For example, the distinction between levels is not always clear and researchers with different agendas may divide things differently. 

Let's take the example of texture-bias in CNNs \citep{geirhos2018imagenet}. One way to describe this according to Marr's levels is to say: 1.) the computational goal of the system is to identify objects in images, 2.) the algorithm it uses involves looking for certain textures associated with different objects, 3.) and the implementation can be found in convolutional filters that identify textures. According to this view, one could say that the computational level is the same for both CNNs and the brain's visual system, but that the algorithm used is different because the brain relies more on shape \citep{landau1992syntactic, geirhos2018imagenet, hermann2022shape} . Another way to describe this system however is to incorporate the behavioral finding of texture bias into the computational description. For example, we could say the computational goal is to recognize objects according to texture. Then the algorithm may be described in more specific terms regarding how different textures are compared. Described this way, CNNs and the brain differ on the computational level. Many times, it is not clear which way of breaking systems down is more `correct' and indeed, different researchers may be perfectly justified in taking different approaches. 

The above example also highlights a lesson of particular relevance to researchers studying artificial neural networks: what you think you are training your model to do may not actually be what it does. CNNs are not explicitly trained to perform texture recognition, but this may be what they learn after training on data that is meant to represent `object recognition' more broadly. For this reason, we believe careful behavioral  studies are important for defining the computational level description and will lead to more productive explorations of algorithm and implementation. This behavioral documentation has always been important for studying the brain, as there is of course no a priori knowledge of the `objective function' there. 

It is also helpful to recognize that a full understanding may require iterating through levels in order to refine computational goals and identify the part of the physical implementation that is of interest. A good example of this comes from the notion of "core object recognition" in the study of biological vision \citep{dicarlo2012does}. Early psychophysics experiments at the computational level identified many different visual capabilities in humans and animals, including various types of object recognition and non-object skills such as visual tracking. Implementation-level analyses of neural activity showed that the ventral stream of the visual system was more involved in object recognition and the dorsal stream more involved in motion and location tracking \citep{de2011usefulness}. Subsequent and more precise behavioral investigations characterized the visual system's ability to rapidly identify simple and centrally-presented objects despite variations in size and rotation \citep{fabre2011characteristics}. This is now known as 'core object recognition' and its speed suggests it is primarily the results of the feedforward pass of the ventral visual stream \citep{dicarlo2012does}. Algorithmic level investigations have come to describe the feedforward ventral stream as implementing an `untangling' of representations that allow different objects to be represented in different and separable parts of activity space \citep{dicarlo2007untangling}. As a result of a combination of behavioral and neural studies, other abilities of the visual system such as robustness to noise or occlusions is believed to be implemented by recurrent connections \citep{lamme2000distinct}.  In total, this shows how a dissection at the implementation level can narrow what behaviors are of interest at the computational level, and these findings jointly constrain the algorithmic level. 

Finally, it is important to note that Marr's framework has not gone unchallenged within the brain sciences. To start, there has been some debate about which level is the best to start at, with the computational level being the default but arguments more aligned with the algorithmic or implementational level have been made \citep{cooper2015beyond, gyorgy2019brain}. Various researchers have also offered revisions to the framework including changes to existing levels and addition of new levels; some have questioned the general utility of the levels overall \citep{cooper2018relation, Griffiths2015RationalUO, bechtel1994levels, hardcastle2015marr, mcclamrock1991marr, bickle2015marr, pillow2024cross, poggio2012levels}. We are not arguing here that Marr's levels is the only viable framework for studying neural information processing systems, but just that it may prove useful for interpretability researchers as they tackle similar problems as found in the brain sciences.

\subsection{What are some notable lessons from the history of studying the brain?}
In the spirit of helping the interpretability community learn from decades of past research studying the brain, we include here a list of some major themes that have affected neuroscience research (while noting that the list is far from comprehensive):
\begin{itemize}
    \item \textbf{From single neuron to neuron population} \label{popcoding}Neuroscientists once focused on studying individual neurons but multi-neural recording methods now enable the study of ensembles of neurons \citep{yuste2015neuron}. In interpretability, Anthropic's recent paper represents a similar shift from individual neurons to groups of neurons \citep{bricken2023towards}. 
    \item \textbf{Multiple realizability and context-dependency} Precise studies of even small neural circuits have shown that the same function can be implemented by a wide range of different connectivity patterns \citep{prinz2004similar}. What's more, the same circuit can implement different functions depending on other contextual signals such as the presence of neuromodulators \citep{harris1991modulation}.  
    \item \textbf{Tools and guiding vision} While much of neuroscience focused on improving the techniques used for recording from neurons and measuring their connectivity, it also became clear that an improvement in tools is not automatically leading to a more holistic understanding of the brain and behaviors\citep{Krakauer2017}.
    \item \textbf{The omnipresence of information} Recent work, particularly in mice, has shown that many different types of information (behavioral state, visual stimulus, etc.) can be read out from the activity of different neural populations all across the brain \citep{ren2021characterizing}. This challenges a long-standing emphasis on modularity in neuroscience. However, locations where information is present are not always shown to be causally relevant for behavior. This raises questions of whether an experimenter's ability to decode information from a neural population indicates anything about how or if that information is actually used by that population \citep{de2016neuroimaging}.
    \item \textbf{Circuit vs Representation views of cognition} The `Sherringtonian' view of cognition emphasizes the role of specific neurons in circuits that carry out particular computations; however this emphasis has been displaced in much of neuroscience by a more `Hopfieldian' view of cognition \citep{barack2021two}, which focuses on representational spaces to gain a more holistic understanding of the inner workings of brains.  Similarly, the representation engineering technique takes inspiration from the Hopfieldian view of cognition \citep{zou2023representation}. 
The existence of some analogous shifts in interpretability communities indicate that lessons learned from the broad neuroscience community could be of use to expedite interpretability research.   
\end{itemize}
\subsection{Where can the analogs between the study of the brain and interpretability be misleading?}
While there are rich synergies to be mined between the way we approach biological and artificial information processing systems, it is important to couch such efforts in the recognition that biological and artificial systems are fundamentally different. In addition to differences mentioned throughout, we highlight a few more pertinent ones here. 

First, artificial “neurons” are not biological neurons; real neurons – cells in our brains – have rich chemical interchanges and distinct functional units \citep{Hassabis2017NeuroscienceInspiredAI}; in contrast, artificial neurons simply compute a weighted sum of the inputs and apply some nonlinear function over the outputs. There are a myriad of other cell types, chemical processes, and physical constraints that are relevant to the computations observed in brain, but are not directly captured by the standard rate-based neurons used in commonly used neural networks \citep{achterberg2023spatially, perez2021neural, gast2024neural}. The complexity of biological neural systems may make the separation between algorithmic and implementation level clearer, as the biological complexities offers many possible implementations of a given algorithm. In systems using simpler neural units, such as ANNs, this distinction may be weaker.

Second, the degree to which we can access and manipulate biological and artificial systems is different. There are a couple of reasons for this. For one, stochasticity in biological brains differs from that of ANNs. Neuroscience and cognitive science findings suggest that neurons can carry out probabilistic inference, which can lead to neural spike outputs differing even in identical experiments \citep{buesing2011neural}.  Even in scenarios where no probabilistic inference is involved per se, the role that a unique neuron can take from situation to situation can differ strongly \citep{duncan2001adaptive}. In contrast, in artificial systems, we can directly inject more or less stochasticity \citep{sabuncu2020intelligence}. Further, for technical and ethical reasons, we often face acute limitations on how far we can directly observe or intervene on biological networks to probe their function. While some artificial systems may too have barriers to direct probing (e.g., if behind an API), we see the prospects for direct editability as an important distinction between biological and artificial information processing systems when designing interpretability studies. We point to \citep{olah2021interpretability} for a useful informal write-up on these topics.

Third, questions around the architectural parity between biological and artificial neural networks remain (and are actively debated). The extent to which the brain implements backpropagation or a similar algorithm \citep{lillicrap2020backpropagation} is debated, even though biologically-plausible mechanisms of backpropagation have been found \citep{greedy2022single}. Additionally, there are notable differences between their learning efficiency and generalization (biological systems learn very few examples but generalize far better \citep{lake2015human}), energy efficiency (brains are orders of magnitude more efficient than deep learning models) \citep{laughlin2003communication, wan2024efficient} and neurogenesis (the ability to grow new neurons throughout life) \citep{aimone2011resolving}. Some of these differences may make localizing functions to specific parts of the system more difficult in ANNs than in brains. For example, if biological learning rules act more locally, they may lead to more modular structure and function; this may make mapping algorithm to implementation easier.

\section{Conclusion}
We started this review by assessing how the research fields focused on the brain (i.e., neuroscience, cognitive science, and psychology) can help to advance AI interpretability efforts. In this work, we focus on the framework of Marr's levels of analysis - to understand neural systems and detail methods from the implementation level, algorithmic/representational level, and computational level. At each level, we provide a sampling of existing methods that are helping brain researchers rigorously refine our understanding of neural systems. We observe that, thus far, much of the work in the interpretability community might can be seen as focusing on the implementation level. Marr's levels of analysis provide one helpful framework to organize interpretability research and evaluate and refine our understanding. Each level offers a non-redundant contribution to our understanding, whereas the connections between levels introduce constraints to levels of analysis that otherwise stay elusive to us. A more integrated understanding that explicitly studies phenomena on each level and connects across them might offer a more fruitful path forward. In turn, we see immense promise in scientific insights and new analysis tools developed from the AI interpretability research community feeding back to those studying biological brains. The study of complex neural systems -- whether biological or artificial -- is just that: complex. We need all the tools we can, and different perspectives, thinking about these problems

\section*{Acknowledgments}
We would like to thanks 

\bibliography{references}

\begin{thebibliography}{361}
\providecommand{\natexlab}[1]{#1}
\providecommand{\url}[1]{\texttt{#1}}
\expandafter\ifx\csname urlstyle\endcsname\relax
  \providecommand{\doi}[1]{doi: #1}\else
  \providecommand{\doi}{doi: \begingroup \urlstyle{rm}\Url}\fi

\bibitem[Abbott(1994)]{abbott1994decoding}
L.~Abbott.
\newblock Decoding neuronal firing and modelling neural networks.
\newblock \emph{Quarterly reviews of biophysics}, 27\penalty0 (3):\penalty0 291--331, 1994.

\bibitem[Abe et~al.(2007)Abe, Suzuki, Mori, Itoh, and Fujii]{abe2007deceiving}
N.~Abe, M.~Suzuki, E.~Mori, M.~Itoh, and T.~Fujii.
\newblock Deceiving others: distinct neural responses of the prefrontal cortex and amygdala in simple fabrication and deception with social interactions.
\newblock \emph{Journal of Cognitive Neuroscience}, 19\penalty0 (2):\penalty0 287--295, 2007.

\bibitem[Achard and Bullmore(2007)]{achard2007efficiency}
S.~Achard and E.~Bullmore.
\newblock Efficiency and cost of economical brain functional networks.
\newblock \emph{PLoS Computational Biology}, 3\penalty0 (2):\penalty0 e17, 2007.

\bibitem[Achiam et~al.(2023)Achiam, Adler, Agarwal, Ahmad, Akkaya, Aleman, Almeida, Altenschmidt, Altman, Anadkat, et~al.]{achiam2023gpt}
J.~Achiam, S.~Adler, S.~Agarwal, L.~Ahmad, I.~Akkaya, F.~L. Aleman, D.~Almeida, J.~Altenschmidt, S.~Altman, S.~Anadkat, et~al.
\newblock Gpt-4 technical report.
\newblock \emph{arXiv preprint arXiv:2303.08774}, 2023.

\bibitem[Achterberg et~al.(2023)Achterberg, Akarca, Strouse, Duncan, and Astle]{achterberg2023spatially}
J.~Achterberg, D.~Akarca, D.~J. Strouse, J.~Duncan, and D.~E. Astle.
\newblock Spatially embedded recurrent neural networks reveal widespread links between structural and functional neuroscience findings.
\newblock \emph{Nature Machine Intelligence}, 2023.

\bibitem[Acker et~al.(2016)Acker, Pino, Boyden, and Desimone]{acker2016fef}
L.~Acker, E.~N. Pino, E.~S. Boyden, and R.~Desimone.
\newblock Fef inactivation with improved optogenetic methods.
\newblock \emph{Proceedings of the National Academy of Sciences}, 113\penalty0 (46):\penalty0 E7297--E7306, 2016.

\bibitem[Adams et~al.(2012)Adams, Watson, Pearson, and Platt]{adams2012neuroethology}
G.~K. Adams, K.~K. Watson, J.~Pearson, and M.~L. Platt.
\newblock Neuroethology of decision-making.
\newblock \emph{Current opinion in neurobiology}, 22\penalty0 (6):\penalty0 982--989, 2012.

\bibitem[Aimone et~al.(2011)Aimone, Deng, and Gage]{aimone2011resolving}
J.~B. Aimone, W.~Deng, and F.~H. Gage.
\newblock Resolving new memories: a critical look at the dentate gyrus, adult neurogenesis, and pattern separation.
\newblock \emph{Neuron}, 70\penalty0 (4):\penalty0 589--596, 2011.

\bibitem[Aitchison and Lengyel(2016)]{aitchison2016hamiltonian}
L.~Aitchison and M.~Lengyel.
\newblock The hamiltonian brain: Efficient probabilistic inference with excitatory-inhibitory neural circuit dynamics.
\newblock \emph{PLoS computational biology}, 12\penalty0 (12):\penalty0 e1005186, 2016.

\bibitem[Al-Shawaf(2024)]{al2024levels}
L.~Al-Shawaf.
\newblock Levels of analysis and explanatory progress in psychology: Integrating frameworks from biology and cognitive science for a more comprehensive science of the mind.
\newblock \emph{Psychological review}, 2024.

\bibitem[Alain and Bengio(2016)]{alain2016understanding}
G.~Alain and Y.~Bengio.
\newblock Understanding intermediate layers using linear classifier probes.
\newblock \emph{arXiv preprint arXiv:1610.01644}, 2016.

\bibitem[Altman and Krzywinski(2018)]{altman2018curse}
N.~Altman and M.~Krzywinski.
\newblock The curse (s) of dimensionality.
\newblock \emph{Nat Methods}, 15\penalty0 (6):\penalty0 399--400, 2018.

\bibitem[Amer and Maul(2019)]{amer2019review}
M.~Amer and T.~Maul.
\newblock A review of modularization techniques in artificial neural networks.
\newblock \emph{Artificial Intelligence Review}, 52:\penalty0 527--561, 2019.

\bibitem[Anand and Dhikav(2012)]{anand2012hippocampus}
K.~S. Anand and V.~Dhikav.
\newblock Hippocampus in health and disease: An overview.
\newblock \emph{Annals of Indian Academy of Neurology}, 15\penalty0 (4):\penalty0 239--246, 2012.

\bibitem[Anderson(2015)]{anderson2015can}
B.~L. Anderson.
\newblock Can computational goals inform theories of vision?
\newblock \emph{Topics in Cognitive Science}, 7\penalty0 (2):\penalty0 274--286, 2015.

\bibitem[Anderson(1972)]{anderson1972more}
P.~W. Anderson.
\newblock More is different: Broken symmetry and the nature of the hierarchical structure of science.
\newblock \emph{Science}, 177\penalty0 (4047):\penalty0 393--396, 1972.

\bibitem[Antonello et~al.(2023)Antonello, Vaidya, and Huth]{antonello2023scaling}
R.~Antonello, A.~Vaidya, and A.~Huth.
\newblock Scaling laws for language encoding models in fmri.
\newblock In A.~Oh, T.~Naumann, A.~Globerson, K.~Saenko, M.~Hardt, and S.~Levine, editors, \emph{Advances in Neural Information Processing Systems}, volume~36, pages 21895--21907. Curran Associates, Inc., 2023.
\newblock URL \url{https://proceedings.neurips.cc/paper_files/paper/2023/file/4533e4a352440a32558c1c227602c323-Paper-Conference.pdf}.

\bibitem[Anwar et~al.(2024)Anwar, Saparov, Rando, Paleka, Turpin, Hase, Lubana, Jenner, Casper, Sourbut, et~al.]{anwar2024foundational}
U.~Anwar, A.~Saparov, J.~Rando, D.~Paleka, M.~Turpin, P.~Hase, E.~S. Lubana, E.~Jenner, S.~Casper, O.~Sourbut, et~al.
\newblock Foundational challenges in assuring alignment and safety of large language models.
\newblock \emph{arXiv preprint arXiv:2404.09932}, 2024.

\bibitem[Aoi and Pillow(2018)]{aoi2018model}
M.~Aoi and J.~W. Pillow.
\newblock Model-based targeted dimensionality reduction for neuronal population data.
\newblock \emph{Advances in neural information processing systems}, 31, 2018.

\bibitem[Aoi et~al.(2020)Aoi, Mante, and Pillow]{aoi2020prefrontal}
M.~C. Aoi, V.~Mante, and J.~W. Pillow.
\newblock Prefrontal cortex exhibits multidimensional dynamic encoding during decision-making.
\newblock \emph{Nature neuroscience}, 23\penalty0 (11):\penalty0 1410--1420, 2020.

\bibitem[Ashida and Carr(2011)]{AshidaCarr2011}
G.~Ashida and C.~E. Carr.
\newblock Sound localization: Jeffress and beyond.
\newblock \emph{Current Opinion in Neurobiology}, 21\penalty0 (5):\penalty0 745--751, 2011.
\newblock \doi{10.1016/j.conb.2011.05.008}.
\newblock URL \url{https://pubmed.ncbi.nlm.nih.gov/21646012/}.

\bibitem[Averbeck et~al.(2006)Averbeck, Latham, and Pouget]{averbeck2006neural}
B.~B. Averbeck, P.~E. Latham, and A.~Pouget.
\newblock Neural correlations, population coding and computation.
\newblock \emph{Nature reviews neuroscience}, 7\penalty0 (5):\penalty0 358--366, 2006.

\bibitem[Baggio et~al.(2015)Baggio, van Lambalgen, and Hagoort]{baggio2015logic}
G.~Baggio, M.~van Lambalgen, and P.~Hagoort.
\newblock Logic as marr's computational level: Four case studies.
\newblock \emph{Topics in Cognitive Science}, 7\penalty0 (2):\penalty0 287--298, 2015.

\bibitem[Baker et~al.(2011)Baker, Saxe, and Tenenbaum]{baker2011bayesian}
C.~Baker, R.~Saxe, and J.~Tenenbaum.
\newblock Bayesian theory of mind: Modeling joint belief-desire attribution.
\newblock In \emph{Proceedings of the annual meeting of the cognitive science society}, volume~33, 2011.

\bibitem[Baker et~al.(2009)Baker, Saxe, and Tenenbaum]{baker2009action}
C.~L. Baker, R.~Saxe, and J.~B. Tenenbaum.
\newblock Action understanding as inverse planning.
\newblock \emph{Cognition}, 113\penalty0 (3):\penalty0 329--349, 2009.

\bibitem[Baker et~al.(2017)Baker, Jara-Ettinger, Saxe, and Tenenbaum]{baker2017rational}
C.~L. Baker, J.~Jara-Ettinger, R.~Saxe, and J.~B. Tenenbaum.
\newblock Rational quantitative attribution of beliefs, desires and percepts in human mentalizing.
\newblock \emph{Nature Human Behaviour}, 1\penalty0 (4):\penalty0 0064, 2017.

\bibitem[Balestriero et~al.(2023)Balestriero, Cosentino, and Shekkizhar]{balestriero2023characterizing}
R.~Balestriero, R.~Cosentino, and S.~Shekkizhar.
\newblock Characterizing large language model geometry solves toxicity detection and generation.
\newblock \emph{arXiv preprint arXiv:2312.01648}, 2023.

\bibitem[Banich and Compton(2018)]{banich2018cognitive}
M.~T. Banich and R.~J. Compton.
\newblock \emph{Cognitive neuroscience}.
\newblock Cambridge University Press, 2018.

\bibitem[Barack and Krakauer(2021)]{barack2021two}
D.~L. Barack and J.~W. Krakauer.
\newblock Two views on the cognitive brain.
\newblock \emph{Nature Reviews Neuroscience}, 22\penalty0 (6):\penalty0 359--371, 2021.

\bibitem[Barlow et~al.(1961)]{barlow1961possible}
H.~B. Barlow et~al.
\newblock Possible principles underlying the transformation of sensory messages.
\newblock \emph{Sensory communication}, 1\penalty0 (01):\penalty0 217--233, 1961.

\bibitem[Bass and Chagnaud(2012)]{bass2012shared}
A.~H. Bass and B.~P. Chagnaud.
\newblock Shared developmental and evolutionary origins for neural basis of vocal--acoustic and pectoral--gestural signaling.
\newblock \emph{Proceedings of the National Academy of Sciences}, 109\penalty0 (supplement\_1):\penalty0 10677--10684, 2012.

\bibitem[Battaglia et~al.(2013)Battaglia, Hamrick, and Tenenbaum]{battaglia2013simulation}
P.~W. Battaglia, J.~B. Hamrick, and J.~B. Tenenbaum.
\newblock Simulation as an engine of physical scene understanding.
\newblock \emph{Proceedings of the National Academy of Sciences}, 110\penalty0 (45):\penalty0 18327--18332, 2013.

\bibitem[Bau et~al.(2020{\natexlab{a}})Bau, Zhu, Strobelt, Lapedriza, Zhou, and Torralba]{bau2020understanding}
D.~Bau, J.-Y. Zhu, H.~Strobelt, A.~Lapedriza, B.~Zhou, and A.~Torralba.
\newblock Understanding the role of individual units in a deep neural network.
\newblock \emph{Proceedings of the National Academy of Sciences}, 117\penalty0 (48):\penalty0 30071--30078, 2020{\natexlab{a}}.

\bibitem[Bau et~al.(2020{\natexlab{b}})Bau, Zhu, Strobelt, Lapedriza, Zhou, and Torralba]{bau2020units}
D.~Bau, J.-Y. Zhu, H.~Strobelt, A.~Lapedriza, B.~Zhou, and A.~Torralba.
\newblock Understanding the role of individual units in a deep neural network.
\newblock \emph{Proceedings of the National Academy of Sciences}, 2020{\natexlab{b}}.
\newblock ISSN 0027-8424.
\newblock \doi{10.1073/pnas.1907375117}.
\newblock URL \url{https://www.pnas.org/content/early/2020/08/31/1907375117}.

\bibitem[Bechtel(1994)]{bechtel1994levels}
W.~Bechtel.
\newblock Levels of description and explanation in cognitive science.
\newblock \emph{Minds and Machines}, 4:\penalty0 1--25, 1994.

\bibitem[Bechtel(2009)]{bechtel2009looking}
W.~Bechtel.
\newblock Looking down, around, and up: Mechanistic explanation in psychology.
\newblock \emph{Philosophical Psychology}, 22\penalty0 (5):\penalty0 543--564, 2009.

\bibitem[Bechtel and Richardson(2010)]{bechtel2010discovering}
W.~Bechtel and R.~C. Richardson.
\newblock \emph{Discovering complexity: Decomposition and localization as strategies in scientific research}.
\newblock MIT press, 2010.

\bibitem[Bechtel and Shagrir(2015)]{bechtel2015non}
W.~Bechtel and O.~Shagrir.
\newblock The non-redundant contributions of marr's three levels of analysis for explaining information-processing mechanisms.
\newblock \emph{Topics in Cognitive Science}, 7\penalty0 (2):\penalty0 312--322, 2015.

\bibitem[Belinkov(2022{\natexlab{a}})]{belinkov-2022-probing}
Y.~Belinkov.
\newblock Probing classifiers: Promises, shortcomings, and advances.
\newblock \emph{Computational Linguistics}, 48\penalty0 (1):\penalty0 207--219, Mar. 2022{\natexlab{a}}.
\newblock \doi{10.1162/coli_a_00422}.
\newblock URL \url{https://aclanthology.org/2022.cl-1.7}.

\bibitem[Belinkov(2022{\natexlab{b}})]{belinkov2022probing}
Y.~Belinkov.
\newblock Probing classifiers: Promises, shortcomings, and advances.
\newblock \emph{Computational Linguistics}, 48\penalty0 (1):\penalty0 207--219, 2022{\natexlab{b}}.

\bibitem[Benichov et~al.(2016)Benichov, Benezra, Vallentin, Globerson, Long, and Tchernichovski]{benichov2016forebrain}
J.~I. Benichov, S.~E. Benezra, D.~Vallentin, E.~Globerson, M.~A. Long, and O.~Tchernichovski.
\newblock The forebrain song system mediates predictive call timing in female and male zebra finches.
\newblock \emph{Current Biology}, 26\penalty0 (3):\penalty0 309--318, 2016.

\bibitem[Bergmann and Hartwigsen(2021)]{bergmann2021inferring}
T.~O. Bergmann and G.~Hartwigsen.
\newblock Inferring causality from noninvasive brain stimulation in cognitive neuroscience.
\newblock \emph{Journal of cognitive neuroscience}, 33\penalty0 (2):\penalty0 195--225, 2021.

\bibitem[Bernardi et~al.(2020)Bernardi, Benna, Rigotti, Munuera, Fusi, and Salzman]{bernardi2020geometry}
S.~Bernardi, M.~K. Benna, M.~Rigotti, J.~Munuera, S.~Fusi, and C.~D. Salzman.
\newblock The geometry of abstraction in the hippocampus and prefrontal cortex.
\newblock \emph{Cell}, 183\penalty0 (4):\penalty0 954--967, 2020.

\bibitem[Bickle(2015)]{bickle2015marr}
J.~Bickle.
\newblock Marr and reductionism.
\newblock \emph{Topics in cognitive science}, 7\penalty0 (2):\penalty0 299--311, 2015.

\bibitem[Bills et~al.(2023)Bills, Cammarata, Mossing, Tillman, Gao, Goh, Sutskever, Leike, Wu, and Saunders]{bills2023language}
S.~Bills, N.~Cammarata, D.~Mossing, H.~Tillman, L.~Gao, G.~Goh, I.~Sutskever, J.~Leike, J.~Wu, and W.~Saunders.
\newblock Language models can explain neurons in language models.
\newblock \url{https://openaipublic.blob.core.windows.net/neuron-explainer/paper/index.html}, 2023.

\bibitem[Blokpoel(2018)]{blokpoel2018sculpting}
M.~Blokpoel.
\newblock Sculpting computational-level models.
\newblock \emph{Topics in cognitive science}, 10\penalty0 (3):\penalty0 641--648, 2018.

\bibitem[Botvinick et~al.(2019)Botvinick, Ritter, Wang, Kurth-Nelson, Blundell, and Hassabis]{botvinick2019reinforcement}
M.~Botvinick, S.~Ritter, J.~X. Wang, Z.~Kurth-Nelson, C.~Blundell, and D.~Hassabis.
\newblock Reinforcement learning, fast and slow.
\newblock \emph{Trends in cognitive sciences}, 23\penalty0 (5):\penalty0 408--422, 2019.

\bibitem[Bowers(2009)]{bowers2009biological}
J.~S. Bowers.
\newblock On the biological plausibility of grandmother cells: implications for neural network theories in psychology and neuroscience.
\newblock \emph{Psychological review}, 116\penalty0 (1):\penalty0 220, 2009.

\bibitem[Brendel et~al.(2011)Brendel, Romo, and Machens]{brendel2011demixed}
W.~Brendel, R.~Romo, and C.~K. Machens.
\newblock Demixed principal component analysis.
\newblock \emph{Advances in neural information processing systems}, 24, 2011.

\bibitem[Bricken et~al.(2023)Bricken, Templeton, Batson, Chen, Jermyn, Conerly, Turner, Anil, Denison, Askell, et~al.]{bricken2023towards}
T.~Bricken, A.~Templeton, J.~Batson, B.~Chen, A.~Jermyn, T.~Conerly, N.~Turner, C.~Anil, C.~Denison, A.~Askell, et~al.
\newblock Towards monosemanticity: Decomposing language models with dictionary learning.
\newblock \emph{Transformer Circuits Thread}, 2, 2023.

\bibitem[Buckner and Krienen(2013)]{buckner2013evolution}
R.~L. Buckner and F.~M. Krienen.
\newblock The evolution of distributed association networks in the human brain.
\newblock \emph{Trends in cognitive sciences}, 17\penalty0 (12):\penalty0 648--665, 2013.

\bibitem[Buesing et~al.(2011)Buesing, Bill, Nessler, and Maass]{buesing2011neural}
L.~Buesing, J.~Bill, B.~Nessler, and W.~Maass.
\newblock Neural dynamics as sampling: a model for stochastic computation in recurrent networks of spiking neurons.
\newblock \emph{PLoS computational biology}, 7\penalty0 (11):\penalty0 e1002211, 2011.

\bibitem[Bullmore and Sporns(2012)]{bullmore2012economy}
E.~Bullmore and O.~Sporns.
\newblock The economy of brain network organization.
\newblock \emph{Nature Reviews Neuroscience}, 13\penalty0 (5):\penalty0 336--349, 2012.

\bibitem[Bush and Mosteller(1955)]{bush1955stochastic}
R.~R. Bush and F.~Mosteller.
\newblock \emph{Stochastic Models for Learning}.
\newblock John Wiley \& Sons, New York, 1955.

\bibitem[Butts and Goldman(2006)]{butts2006tuning}
D.~A. Butts and M.~S. Goldman.
\newblock Tuning curves, neuronal variability, and sensory coding.
\newblock \emph{PLoS biology}, 4\penalty0 (4):\penalty0 e92, 2006.

\bibitem[Cammarata et~al.(2020)Cammarata, Goh, Carter, Schubert, Petrov, and Olah]{cammarata2020curve}
N.~Cammarata, G.~Goh, S.~Carter, L.~Schubert, M.~Petrov, and C.~Olah.
\newblock Curve detectors.
\newblock \emph{Distill}, 5\penalty0 (6):\penalty0 e00024--003, 2020.

\bibitem[Carlini et~al.(2022)Carlini, Ippolito, Jagielski, Lee, Tramer, and Zhang]{carlini2022quantifying}
N.~Carlini, D.~Ippolito, M.~Jagielski, K.~Lee, F.~Tramer, and C.~Zhang.
\newblock Quantifying memorization across neural language models.
\newblock \emph{arXiv preprint arXiv:2202.07646}, 2022.

\bibitem[Carrasco et~al.(2009)Carrasco, Eckstein, Verghese, Boynton, and Treue]{carrasco2009visual}
M.~Carrasco, M.~Eckstein, P.~Verghese, G.~Boynton, and S.~Treue.
\newblock Visual attention: Neurophysiology, psychophysics and cognitive neuroscience, 2009.

\bibitem[Caucheteux et~al.(2021)Caucheteux, Gramfort, and King]{caucheteux2021disentangling}
C.~Caucheteux, A.~Gramfort, and J.-R. King.
\newblock Disentangling syntax and semantics in the brain with deep networks.
\newblock In M.~Meila and T.~Zhang, editors, \emph{Proceedings of the 38th International Conference on Machine Learning}, volume 139 of \emph{Proceedings of Machine Learning Research}, pages 1336--1348. PMLR, 18--24 Jul 2021.
\newblock URL \url{https://proceedings.mlr.press/v139/caucheteux21a.html}.

\bibitem[Chang et~al.(2013)Chang, Brent, Adams, Klein, Pearson, Watson, and Platt]{chang2013neuroethology}
S.~W. Chang, L.~J. Brent, G.~K. Adams, J.~T. Klein, J.~M. Pearson, K.~K. Watson, and M.~L. Platt.
\newblock Neuroethology of primate social behavior.
\newblock \emph{Proceedings of the National Academy of Sciences}, 110\penalty0 (supplement\_2):\penalty0 10387--10394, 2013.

\bibitem[Chater(1996)]{chater1996reconciling}
N.~Chater.
\newblock Reconciling simplicity and likelihood principles in perceptual organization.
\newblock \emph{Psychological Review}, 103\penalty0 (3):\penalty0 566, 1996.

\bibitem[Chater and Manning(2006)]{chater2006probabilisticlanguage}
N.~Chater and C.~D. Manning.
\newblock Probabilistic models of language processing and acquisition.
\newblock \emph{Trends in cognitive sciences}, 10\penalty0 (7):\penalty0 335--344, 2006.

\bibitem[Chater et~al.(2006)Chater, Tenenbaum, and Yuille]{chater2006probabilistic}
N.~Chater, J.~B. Tenenbaum, and A.~Yuille.
\newblock Probabilistic models of cognition: Conceptual foundations.
\newblock \emph{Trends in cognitive sciences}, 10\penalty0 (7):\penalty0 287--291, 2006.

\bibitem[Chater et~al.(2010)Chater, Oaksford, Hahn, and Heit]{l2008bayesian}
N.~Chater, M.~Oaksford, U.~Hahn, and E.~Heit.
\newblock Bayesian models of cognition.
\newblock \emph{Wiley Interdisciplinary Reviews: Cognitive Science}, 1\penalty0 (6):\penalty0 811--823, 2010.

\bibitem[Chern et~al.(2024)Chern, Hu, Yang, Chern, Guo, Jin, Wang, and Liu]{chern2024behonest}
S.~Chern, Z.~Hu, Y.~Yang, E.~Chern, Y.~Guo, J.~Jin, B.~Wang, and P.~Liu.
\newblock Behonest: Benchmarking honesty of large language models.
\newblock \emph{arXiv preprint arXiv:2406.13261}, 2024.

\bibitem[Chua et~al.(2018)Chua, Calandra, McAllister, and Levine]{chua2018deep}
K.~Chua, R.~Calandra, R.~McAllister, and S.~Levine.
\newblock Deep reinforcement learning in a handful of trials using probabilistic dynamics models, 2018.

\bibitem[Chung and Abbott(2021)]{chung2021neural}
S.~Chung and L.~Abbott.
\newblock Neural population geometry: An approach for understanding biological and artificial neural networks.
\newblock \emph{Current opinion in neurobiology}, 70:\penalty0 137--144, 2021.

\bibitem[Chung et~al.(2018)Chung, Lee, and Sompolinsky]{chung2018classification}
S.~Chung, D.~D. Lee, and H.~Sompolinsky.
\newblock Classification and geometry of general perceptual manifolds.
\newblock \emph{Physical Review X}, 8\penalty0 (3):\penalty0 031003, 2018.

\bibitem[Churchland et~al.(2012)Churchland, Cunningham, Kaufman, Foster, Nuyujukian, Ryu, and Shenoy]{churchland2012neural}
M.~M. Churchland, J.~P. Cunningham, M.~T. Kaufman, J.~D. Foster, P.~Nuyujukian, S.~I. Ryu, and K.~V. Shenoy.
\newblock Neural population dynamics during reaching.
\newblock \emph{Nature}, 487\penalty0 (7405):\penalty0 51--56, 2012.

\bibitem[Cisek(2019)]{cisek2019resynthesizing}
P.~Cisek.
\newblock Resynthesizing behavior through phylogenetic refinement.
\newblock \emph{Attention, Perception, \& Psychophysics}, 81:\penalty0 2265--2287, 2019.

\bibitem[Clark and Yuille(2013)]{clark2013data}
J.~J. Clark and A.~L. Yuille.
\newblock \emph{Data fusion for sensory information processing systems}, volume 105.
\newblock Springer Science \& Business Media, 2013.

\bibitem[Clark et~al.(2019)Clark, Khandelwal, Levy, and Manning]{clark2019does}
K.~Clark, U.~Khandelwal, O.~Levy, and C.~D. Manning.
\newblock What does bert look at? an analysis of bert's attention.
\newblock \emph{arXiv preprint arXiv:1906.04341}, 2019.

\bibitem[Clune et~al.(2013)Clune, Mouret, and Lipson]{clune2013evolutionary}
J.~Clune, J.-B. Mouret, and H.~Lipson.
\newblock The evolutionary origins of modularity.
\newblock \emph{Proc. R. Soc. B}, 280\penalty0 (20122863), 2013.
\newblock \doi{10.1098/rspb.2012.2863}.

\bibitem[Coen et~al.(2023)Coen, Sit, Wells, Carandini, and Harris]{coen2023mouse}
P.~Coen, T.~P. Sit, M.~J. Wells, M.~Carandini, and K.~D. Harris.
\newblock Mouse frontal cortex mediates additive multisensory decisions.
\newblock \emph{Neuron}, 111\penalty0 (15):\penalty0 2432--2447, 2023.

\bibitem[Cohen et~al.(2020)Cohen, Chung, Lee, and Sompolinsky]{cohen2020separability}
U.~Cohen, S.~Chung, D.~D. Lee, and H.~Sompolinsky.
\newblock Separability and geometry of object manifolds in deep neural networks.
\newblock \emph{Nature communications}, 11\penalty0 (1):\penalty0 746, 2020.

\bibitem[Contini et~al.(2017)Contini, Wardle, and Carlson]{contini2017decoding}
E.~W. Contini, S.~G. Wardle, and T.~A. Carlson.
\newblock Decoding the time-course of object recognition in the human brain: From visual features to categorical decisions.
\newblock \emph{Neuropsychologia}, 105:\penalty0 165--176, 2017.

\bibitem[Cooper and Peebles(2015)]{cooper2015beyond}
R.~P. Cooper and D.~Peebles.
\newblock Beyond single-level accounts: The role of cognitive architectures in cognitive scientific explanation.
\newblock \emph{Topics in cognitive science}, 7\penalty0 (2):\penalty0 243--258, 2015.

\bibitem[Cooper and Peebles(2018)]{cooper2018relation}
R.~P. Cooper and D.~Peebles.
\newblock On the relation between marr's levels: A response to blokpoel (2017).
\newblock \emph{Topics in Cognitive Science}, 10\penalty0 (3):\penalty0 649--653, 2018.

\bibitem[Coutanche et~al.(2011)Coutanche, Thompson-Schill, and Schultz]{coutanche2011multi}
M.~N. Coutanche, S.~L. Thompson-Schill, and R.~T. Schultz.
\newblock Multi-voxel pattern analysis of fmri data predicts clinical symptom severity.
\newblock \emph{Neuroimage}, 57\penalty0 (1):\penalty0 113--123, 2011.

\bibitem[Craver(2007)]{craver2007explaining}
C.~F. Craver.
\newblock \emph{Explaining the brain: Mechanisms and the mosaic unity of neuroscience}.
\newblock Clarendon Press, 2007.

\bibitem[Cunningham et~al.(2023)Cunningham, Ewart, Riggs, Huben, and Sharkey]{cunningham2023sparse}
H.~Cunningham, A.~Ewart, L.~Riggs, R.~Huben, and L.~Sharkey.
\newblock Sparse autoencoders find highly interpretable features in language models.
\newblock \emph{arXiv preprint arXiv:2309.08600}, 2023.

\bibitem[Cunningham and Yu(2014)]{cunningham2014dimensionality}
J.~P. Cunningham and B.~M. Yu.
\newblock Dimensionality reduction for large-scale neural recordings.
\newblock \emph{Nature neuroscience}, 17\penalty0 (11):\penalty0 1500--1509, 2014.

\bibitem[Dai et~al.(2021)Dai, Dong, Hao, Sui, Chang, and Wei]{dai2021knowledge}
D.~Dai, L.~Dong, Y.~Hao, Z.~Sui, B.~Chang, and F.~Wei.
\newblock Knowledge neurons in pretrained transformers.
\newblock \emph{arXiv preprint arXiv:2104.08696}, 2021.

\bibitem[Datta et~al.(2019)Datta, Anderson, Branson, Perona, and Leifer]{datta2019computational}
S.~R. Datta, D.~J. Anderson, K.~Branson, P.~Perona, and A.~Leifer.
\newblock Computational neuroethology: a call to action.
\newblock \emph{Neuron}, 104\penalty0 (1):\penalty0 11--24, 2019.

\bibitem[Davies et~al.(2023)Davies, Langosco, and Krueger]{davies2023unifying}
X.~Davies, L.~Langosco, and D.~Krueger.
\newblock Unifying grokking and double descent, 2023.

\bibitem[Dayan and Abbott(2005)]{dayan2005theoretical}
P.~Dayan and L.~F. Abbott.
\newblock \emph{Theoretical neuroscience: computational and mathematical modeling of neural systems}.
\newblock MIT press, 2005.

\bibitem[de~Haan and Cowey(2011)]{de2011usefulness}
E.~H. de~Haan and A.~Cowey.
\newblock On the usefulness of ‘what’and ‘where’pathways in vision.
\newblock \emph{Trends in cognitive sciences}, 15\penalty0 (10):\penalty0 460--466, 2011.

\bibitem[de~Wit et~al.(2016)de~Wit, Alexander, Ekroll, and Wagemans]{de2016neuroimaging}
L.~de~Wit, D.~Alexander, V.~Ekroll, and J.~Wagemans.
\newblock Is neuroimaging measuring information in the brain?
\newblock \emph{Psychonomic bulletin \& review}, 23:\penalty0 1415--1428, 2016.

\bibitem[Dewsbury(1978)]{dewsbury1978comparative}
D.~A. Dewsbury.
\newblock \emph{Comparative animal behavior.}
\newblock McGraw-Hill, 1978.

\bibitem[DiCarlo and Cox(2007)]{dicarlo2007untangling}
J.~J. DiCarlo and D.~D. Cox.
\newblock Untangling invariant object recognition.
\newblock \emph{Trends in cognitive sciences}, 11\penalty0 (8):\penalty0 333--341, 2007.

\bibitem[DiCarlo et~al.(2012)DiCarlo, Zoccolan, and Rust]{dicarlo2012does}
J.~J. DiCarlo, D.~Zoccolan, and N.~C. Rust.
\newblock How does the brain solve visual object recognition?
\newblock \emph{Neuron}, 73\penalty0 (3):\penalty0 415--434, 2012.

\bibitem[Dirani and Pylkk{\"a}nen(2023)]{dirani2023time}
J.~Dirani and L.~Pylkk{\"a}nen.
\newblock The time course of cross-modal representations of conceptual categories.
\newblock \emph{Neuroimage}, 277:\penalty0 120254, 2023.

\bibitem[Donnelly and Roegiest(2019)]{donnelly2019interpretability}
J.~Donnelly and A.~Roegiest.
\newblock On interpretability and feature representations: an analysis of the sentiment neuron.
\newblock In \emph{Advances in Information Retrieval: 41st European Conference on IR Research, ECIR 2019, Cologne, Germany, April 14--18, 2019, Proceedings, Part I 41}, pages 795--802. Springer, 2019.

\bibitem[Doshi-Velez and Kim(2017)]{doshivelez2017rigorous}
F.~Doshi-Velez and B.~Kim.
\newblock Towards a rigorous science of interpretable machine learning, 2017.

\bibitem[Dronkers et~al.(2007)Dronkers, Plaisant, Iba-Zizen, and Cabanis]{dronkers2007paul}
N.~F. Dronkers, O.~Plaisant, M.~T. Iba-Zizen, and E.~A. Cabanis.
\newblock Paul broca's historic cases: high resolution mr imaging of the brains of leborgne and lelong.
\newblock \emph{Brain}, 130\penalty0 (5):\penalty0 1432--1441, 2007.

\bibitem[Dudchenko and Wallace(2018)]{dudchenko2018neuroethology}
P.~A. Dudchenko and D.~Wallace.
\newblock Neuroethology of spatial cognition.
\newblock \emph{Current Biology}, 28\penalty0 (17):\penalty0 R988--R992, 2018.

\bibitem[Duncan(2001)]{duncan2001adaptive}
J.~Duncan.
\newblock An adaptive coding model of neural function in prefrontal cortex.
\newblock \emph{Nature reviews neuroscience}, 2\penalty0 (11):\penalty0 820--829, 2001.

\bibitem[Duncan(2010)]{duncan2010multiple}
J.~Duncan.
\newblock The multiple-demand (md) system of the primate brain: mental programs for intelligent behaviour.
\newblock \emph{Trends in cognitive sciences}, 14\penalty0 (4):\penalty0 172--179, 2010.

\bibitem[Dupré~la Tour et~al.(2024)Dupré~la Tour, Visconti~di Oleggio~Castello, and Gallant]{dupre2024voxelwise}
T.~Dupré~la Tour, M.~Visconti~di Oleggio~Castello, and J.~L. Gallant.
\newblock The voxelwise modeling framework: a tutorial introduction to fitting encoding models to f{MRI} data, 2024.

\bibitem[Dutta et~al.(2024)Dutta, Singh, Chakrabarti, and Chakraborty]{dutta2024think}
S.~Dutta, J.~Singh, S.~Chakrabarti, and T.~Chakraborty.
\newblock How to think step-by-step: A mechanistic understanding of chain-of-thought reasoning.
\newblock \emph{arXiv preprint arXiv:2402.18312}, 2024.

\bibitem[Echeveste et~al.(2020)Echeveste, Aitchison, Hennequin, and Lengyel]{echeveste2020cortical}
R.~Echeveste, L.~Aitchison, G.~Hennequin, and M.~Lengyel.
\newblock Cortical-like dynamics in recurrent circuits optimized for sampling-based probabilistic inference.
\newblock \emph{Nature neuroscience}, 23\penalty0 (9):\penalty0 1138--1149, 2020.

\bibitem[Elhage et~al.(2022{\natexlab{a}})Elhage, Hume, Olsson, Nanda, Henighan, Johnston, ElShowk, Joseph, DasSarma, Mann, Hernandez, Askell, Ndousse, Drain, Chen, Bai, Ganguli, Lovitt, Hatfield-Dodds, Kernion, Conerly, Kravec, Fort, Kadavath, Jacobson, Tran-Johnson, Kaplan, Clark, Brown, McCandlish, Amodei, and Olah]{elhage2022solu}
N.~Elhage, T.~Hume, C.~Olsson, N.~Nanda, T.~Henighan, S.~Johnston, S.~ElShowk, N.~Joseph, N.~DasSarma, B.~Mann, D.~Hernandez, A.~Askell, K.~Ndousse, D.~Drain, A.~Chen, Y.~Bai, D.~Ganguli, L.~Lovitt, Z.~Hatfield-Dodds, J.~Kernion, T.~Conerly, S.~Kravec, S.~Fort, S.~Kadavath, J.~Jacobson, E.~Tran-Johnson, J.~Kaplan, J.~Clark, T.~Brown, S.~McCandlish, D.~Amodei, and C.~Olah.
\newblock Softmax linear units.
\newblock \emph{Transformer Circuits Thread}, 2022{\natexlab{a}}.
\newblock https://transformer-circuits.pub/2022/solu/index.html.

\bibitem[Elhage et~al.(2022{\natexlab{b}})Elhage, Hume, Olsson, Schiefer, Henighan, Kravec, Hatfield-Dodds, Lasenby, Drain, Chen, et~al.]{elhage2022toy}
N.~Elhage, T.~Hume, C.~Olsson, N.~Schiefer, T.~Henighan, S.~Kravec, Z.~Hatfield-Dodds, R.~Lasenby, D.~Drain, C.~Chen, et~al.
\newblock Toy models of superposition.
\newblock \emph{arXiv preprint arXiv:2209.10652}, 2022{\natexlab{b}}.

\bibitem[Eliasmith and Kolbeck(2015)]{eliasmith2015marr}
C.~Eliasmith and C.~Kolbeck.
\newblock Marr's attacks: On reductionism and vagueness.
\newblock \emph{Topics in cognitive science}, 7\penalty0 (2):\penalty0 323--335, 2015.

\bibitem[Ernst and Banks(2002)]{ernst2002humans}
M.~O. Ernst and M.~S. Banks.
\newblock Humans integrate visual and haptic information in a statistically optimal fashion.
\newblock \emph{Nature}, 415\penalty0 (6870):\penalty0 429--433, 2002.

\bibitem[Ewert and Ewert(1980)]{ewert1980neuroethology}
J.-P. Ewert and J.-P. Ewert.
\newblock \emph{What is Neuroethology?}
\newblock Springer, 1980.

\bibitem[Fabre-Thorpe(2011)]{fabre2011characteristics}
M.~Fabre-Thorpe.
\newblock The characteristics and limits of rapid visual categorization.
\newblock \emph{Frontiers in psychology}, 2:\penalty0 243, 2011.

\bibitem[(FAIR)† et~al.(2022)(FAIR)†, Bakhtin, Brown, Dinan, Farina, Flaherty, Fried, Goff, Gray, Hu, et~al.]{meta2022human}
M.~F. A. R. D.~T. (FAIR)†, A.~Bakhtin, N.~Brown, E.~Dinan, G.~Farina, C.~Flaherty, D.~Fried, A.~Goff, J.~Gray, H.~Hu, et~al.
\newblock Human-level play in the game of diplomacy by combining language models with strategic reasoning.
\newblock \emph{Science}, 378\penalty0 (6624):\penalty0 1067--1074, 2022.

\bibitem[Fedorenko and Varley(2016)]{fedorenko2016language}
E.~Fedorenko and R.~Varley.
\newblock Language and thought are not the same thing: evidence from neuroimaging and neurological patients.
\newblock \emph{Annals of the New York Academy of Sciences}, 1369\penalty0 (1):\penalty0 132--153, 2016.

\bibitem[Fedorenko et~al.(2011)Fedorenko, Behr, and Kanwisher]{fedorenko2011functional}
E.~Fedorenko, M.~K. Behr, and N.~Kanwisher.
\newblock Functional specificity for high-level linguistic processing in the human brain.
\newblock \emph{Proceedings of the National Academy of Sciences}, 108\penalty0 (39):\penalty0 16428--16433, 2011.

\bibitem[Feng and Steinhardt(2023)]{feng2023language}
J.~Feng and J.~Steinhardt.
\newblock How do language models bind entities in context?
\newblock \emph{arXiv preprint arXiv:2310.17191}, 2023.

\bibitem[Fodor(1983)]{fodor1983modularity}
J.~A. Fodor.
\newblock \emph{The modularity of mind: an essay on faculty psychology}.
\newblock MIT Press, Cambridge, Mass, 1983.

\bibitem[Fusi et~al.(2016)Fusi, Miller, and Rigotti]{fusi2016neurons}
S.~Fusi, E.~K. Miller, and M.~Rigotti.
\newblock Why neurons mix: high dimensionality for higher cognition.
\newblock \emph{Current opinion in neurobiology}, 37:\penalty0 66--74, 2016.

\bibitem[Gallicchio and Scardapane(2020)]{gallicchio2020deep}
C.~Gallicchio and S.~Scardapane.
\newblock Deep randomized neural networks.
\newblock In \emph{Recent Trends in Learning From Data: Tutorials from the INNS Big Data and Deep Learning Conference (INNSBDDL2019)}, pages 43--68. Springer, 2020.

\bibitem[Ganis and Keenan(2009)]{ganis2009cognitive}
G.~Ganis and J.~P. Keenan.
\newblock The cognitive neuroscience of deception.
\newblock \emph{Social Neuroscience}, 4\penalty0 (6):\penalty0 465--472, 2009.

\bibitem[Gast et~al.(2024)Gast, Solla, and Kennedy]{gast2024neural}
R.~Gast, S.~A. Solla, and A.~Kennedy.
\newblock Neural heterogeneity controls computations in spiking neural networks.
\newblock \emph{Proceedings of the National Academy of Sciences}, 121\penalty0 (3):\penalty0 e2311885121, 2024.

\bibitem[Geiger et~al.(2021)Geiger, Lu, Icard, and Potts]{geiger2021causal}
A.~Geiger, H.~Lu, T.~Icard, and C.~Potts.
\newblock Causal abstractions of neural networks.
\newblock \emph{Advances in Neural Information Processing Systems}, 34:\penalty0 9574--9586, 2021.

\bibitem[Geiger et~al.(2023)Geiger, Wu, Potts, Icard, and Goodman]{geiger2023finding}
A.~Geiger, Z.~Wu, C.~Potts, T.~Icard, and N.~D. Goodman.
\newblock Finding alignments between interpretable causal variables and distributed neural representations.
\newblock \emph{arXiv preprint arXiv:2303.02536}, 2023.

\bibitem[Geirhos et~al.(2018)Geirhos, Rubisch, Michaelis, Bethge, Wichmann, and Brendel]{geirhos2018imagenet}
R.~Geirhos, P.~Rubisch, C.~Michaelis, M.~Bethge, F.~A. Wichmann, and W.~Brendel.
\newblock Imagenet-trained cnns are biased towards texture; increasing shape bias improves accuracy and robustness.
\newblock \emph{arXiv preprint arXiv:1811.12231}, 2018.

\bibitem[Gershman et~al.(2015)Gershman, Horvitz, and Tenenbaum]{gershman2015computational}
S.~J. Gershman, E.~J. Horvitz, and J.~B. Tenenbaum.
\newblock Computational rationality: A converging paradigm for intelligence in brains, minds, and machines.
\newblock \emph{Science}, 349\penalty0 (6245):\penalty0 273--278, 2015.

\bibitem[Gescheider(2013)]{gescheider2013psychophysics}
G.~A. Gescheider.
\newblock \emph{Psychophysics: the fundamentals}.
\newblock Psychology Press, 2013.

\bibitem[Gigerenzer(1991)]{gigerenzer1991tools}
G.~Gigerenzer.
\newblock From tools to theories: A heuristic of discovery in cognitive psychology.
\newblock \emph{Psychological review}, 98\penalty0 (2):\penalty0 254, 1991.

\bibitem[Goh et~al.(2021)Goh, Cammarata, Voss, Carter, Petrov, Schubert, Radford, and Olah]{goh2021multimodal}
G.~Goh, N.~Cammarata, C.~Voss, S.~Carter, M.~Petrov, L.~Schubert, A.~Radford, and C.~Olah.
\newblock Multimodal neurons in artificial neural networks.
\newblock \emph{Distill}, 6\penalty0 (3):\penalty0 e30, 2021.

\bibitem[Gomez-Marin(2017)]{gomez2017causal}
A.~Gomez-Marin.
\newblock Causal circuit explanations of behavior: Are necessity and sufficiency necessary and sufficient?
\newblock \emph{Decoding Neural Circuit Structure and Function: Cellular Dissection Using Genetic Model Organisms}, pages 283--306, 2017.

\bibitem[Goodman and Frank(2016)]{goodman2016pragmatic}
N.~D. Goodman and M.~C. Frank.
\newblock Pragmatic language interpretation as probabilistic inference.
\newblock \emph{Trends in cognitive sciences}, 20\penalty0 (11):\penalty0 818--829, 2016.

\bibitem[Gopnik et~al.(2004)Gopnik, Glymour, Sobel, Schulz, Kushnir, and Danks]{gopnik2004theory}
A.~Gopnik, C.~Glymour, D.~M. Sobel, L.~E. Schulz, T.~Kushnir, and D.~Danks.
\newblock A theory of causal learning in children: causal maps and bayes nets.
\newblock \emph{Psychological review}, 111\penalty0 (1):\penalty0 3, 2004.

\bibitem[Greedy et~al.(2022)Greedy, Zhu, Pemberton, Mellor, and Ponte~Costa]{greedy2022single}
W.~Greedy, H.~W. Zhu, J.~Pemberton, J.~Mellor, and R.~Ponte~Costa.
\newblock Single-phase deep learning in cortico-cortical networks.
\newblock \emph{Advances in Neural Information Processing Systems}, 35:\penalty0 24213--24225, 2022.

\bibitem[Griffiths and Tenenbaum(2006)]{griffiths2006optimal}
T.~L. Griffiths and J.~B. Tenenbaum.
\newblock Optimal predictions in everyday cognition.
\newblock \emph{Psychological science}, 17\penalty0 (9):\penalty0 767--773, 2006.

\bibitem[Griffiths et~al.(2010)Griffiths, Chater, Kemp, Perfors, and Tenenbaum]{griffiths2010probabilistic}
T.~L. Griffiths, N.~Chater, C.~Kemp, A.~Perfors, and J.~B. Tenenbaum.
\newblock Probabilistic models of cognition: Exploring representations and inductive biases.
\newblock \emph{Trends in cognitive sciences}, 14\penalty0 (8):\penalty0 357--364, 2010.

\bibitem[Griffiths et~al.(2012)Griffiths, Vul, and Sanborn]{griffiths2012bridging}
T.~L. Griffiths, E.~Vul, and A.~N. Sanborn.
\newblock Bridging levels of analysis for probabilistic models of cognition.
\newblock \emph{Current Directions in Psychological Science}, 21\penalty0 (4):\penalty0 263--268, 2012.

\bibitem[Griffiths et~al.(2015)Griffiths, Lieder, and Goodman]{Griffiths2015RationalUO}
T.~L. Griffiths, F.~Lieder, and N.~D. Goodman.
\newblock Rational use of cognitive resources: Levels of analysis between the computational and the algorithmic.
\newblock \emph{Topics in cognitive science}, 7 2:\penalty0 217--29, 2015.
\newblock URL \url{https://api.semanticscholar.org/CorpusID:2412970}.

\bibitem[Griffiths et~al.(2023)Griffiths, Zhu, Grant, and McCoy]{griffiths2023bayes}
T.~L. Griffiths, J.-Q. Zhu, E.~Grant, and R.~T. McCoy.
\newblock Bayes in the age of intelligent machines, 2023.

\bibitem[Grootswagers et~al.(2017)Grootswagers, Wardle, and Carlson]{grootswagers2017decoding}
T.~Grootswagers, S.~G. Wardle, and T.~A. Carlson.
\newblock Decoding dynamic brain patterns from evoked responses: A tutorial on multivariate pattern analysis applied to time series neuroimaging data.
\newblock \emph{Journal of cognitive neuroscience}, 29\penalty0 (4):\penalty0 677--697, 2017.

\bibitem[Grossberg(2000)]{grossberg2000complementary}
S.~Grossberg.
\newblock The complementary brain: Unifying brain dynamics and modularity.
\newblock \emph{Trends in cognitive sciences}, 4\penalty0 (6):\penalty0 233--246, 2000.

\bibitem[Grothe(2003)]{Grothe2003}
B.~Grothe.
\newblock New roles for synaptic inhibition in sound localization.
\newblock \emph{Nature Reviews Neuroscience}, 4:\penalty0 540--550, 2003.
\newblock \doi{10.1038/nrn1136}.

\bibitem[Gurnee and Tegmark(2023)]{gurnee2023language}
W.~Gurnee and M.~Tegmark.
\newblock Language models represent space and time.
\newblock \emph{arXiv preprint arXiv:2310.02207}, 2023.

\bibitem[Gurnee et~al.(2023)Gurnee, Nanda, Pauly, Harvey, Troitskii, and Bertsimas]{gurnee2023finding}
W.~Gurnee, N.~Nanda, M.~Pauly, K.~Harvey, D.~Troitskii, and D.~Bertsimas.
\newblock Finding neurons in a haystack: Case studies with sparse probing.
\newblock \emph{arXiv preprint arXiv:2305.01610}, 2023.

\bibitem[Gy{\"o}rgy~Buzs{\'a}ki(2019)]{gyorgy2019brain}
M.~Gy{\"o}rgy~Buzs{\'a}ki.
\newblock \emph{The brain from inside out}.
\newblock Oxford University Press, 2019.

\bibitem[Hafting et~al.(2005)Hafting, Fyhn, Molden, Moser, and Moser]{hafting2005microstructure}
T.~Hafting, M.~Fyhn, S.~Molden, M.-B. Moser, and E.~I. Moser.
\newblock Microstructure of a spatial map in the entorhinal cortex.
\newblock \emph{Nature}, 436\penalty0 (7052):\penalty0 801--806, 2005.
\newblock \doi{10.1038/nature03721}.

\bibitem[Hagendorff(2024)]{hagendorff2024deception}
T.~Hagendorff.
\newblock Deception abilities emerged in large language models.
\newblock \emph{Proceedings of the National Academy of Sciences}, 121\penalty0 (24):\penalty0 e2317967121, 2024.

\bibitem[Halawi et~al.(2023)Halawi, Denain, and Steinhardt]{halawi2023overthinking}
D.~Halawi, J.-S. Denain, and J.~Steinhardt.
\newblock Overthinking the truth: Understanding how language models process false demonstrations.
\newblock \emph{arXiv preprint arXiv:2307.09476}, 2023.

\bibitem[Hallett(2007)]{hallett2007transcranial}
M.~Hallett.
\newblock Transcranial magnetic stimulation: a primer.
\newblock \emph{Neuron}, 55\penalty0 (2):\penalty0 187--199, 2007.

\bibitem[Hamrick and Mohamed(2020)]{hamrick2020levels}
J.~Hamrick and S.~Mohamed.
\newblock Levels of analysis for machine learning.
\newblock \emph{arXiv preprint arXiv:2004.05107}, 2020.

\bibitem[Hardcastle and Hardcastle(2015)]{hardcastle2015marr}
V.~G. Hardcastle and K.~Hardcastle.
\newblock Marr's levels revisited: understanding how brains break.
\newblock \emph{Topics in Cognitive Science}, 7\penalty0 (2):\penalty0 259--273, 2015.

\bibitem[Harris and Mrsic-Flogel(2013)]{harris2013cortical}
K.~D. Harris and T.~D. Mrsic-Flogel.
\newblock Cortical connectivity and sensory coding.
\newblock \emph{Nature}, 503\penalty0 (7474):\penalty0 51--58, 2013.

\bibitem[Harris-Warrick and Marder(1991)]{harris1991modulation}
R.~M. Harris-Warrick and E.~Marder.
\newblock Modulation of neural networks for behavior.
\newblock \emph{Annual review of neuroscience}, 14\penalty0 (1):\penalty0 39--57, 1991.

\bibitem[Hassabis et~al.(2017)Hassabis, Kumaran, Summerfield, and Botvinick]{Hassabis2017NeuroscienceInspiredAI}
D.~Hassabis, D.~Kumaran, C.~Summerfield, and M.~M. Botvinick.
\newblock Neuroscience-inspired artificial intelligence.
\newblock \emph{Neuron}, 95:\penalty0 245--258, 2017.
\newblock URL \url{https://api.semanticscholar.org/CorpusID:4511529}.

\bibitem[Hermann et~al.(2020)Hermann, Chen, and Kornblith]{hermann2020origins}
K.~Hermann, T.~Chen, and S.~Kornblith.
\newblock The origins and prevalence of texture bias in convolutional neural networks.
\newblock \emph{Advances in Neural Information Processing Systems}, 33:\penalty0 19000--19015, 2020.

\bibitem[Hermann and Firestone(2022)]{hermann2022shape}
K.~L. Hermann and C.~Firestone.
\newblock Shape bias at a glance: Comparing human and machine vision on equal terms.
\newblock \emph{Journal of Vision}, 22\penalty0 (14):\penalty0 3255--3255, 2022.

\bibitem[Hestness et~al.(2017)Hestness, Narang, Ardalani, Diamos, Jun, Kianinejad, Patwary, Yang, and Zhou]{hestness2017deep}
J.~Hestness, S.~Narang, N.~Ardalani, G.~Diamos, H.~Jun, H.~Kianinejad, M.~M.~A. Patwary, Y.~Yang, and Y.~Zhou.
\newblock Deep learning scaling is predictable, empirically.
\newblock \emph{arXiv preprint arXiv:1712.00409}, 2017.

\bibitem[Hewitt and Liang(2019)]{hewitt-liang-2019-designing}
J.~Hewitt and P.~Liang.
\newblock Designing and interpreting probes with control tasks.
\newblock In \emph{Proceedings of the 2019 Conference on Empirical Methods in Natural Language Processing and the 9th International Joint Conference on Natural Language Processing (EMNLP-IJCNLP)}, pages 2733--2743, Hong Kong, China, Nov. 2019. Association for Computational Linguistics.
\newblock \doi{10.18653/v1/D19-1275}.
\newblock URL \url{https://www.aclweb.org/anthology/D19-1275}.

\bibitem[Hewitt and Manning(2019)]{hewitt2019structural}
J.~Hewitt and C.~D. Manning.
\newblock A structural probe for finding syntax in word representations.
\newblock In \emph{Proceedings of the 2019 Conference of the North American Chapter of the Association for Computational Linguistics: Human Language Technologies, Volume 1 (Long and Short Papers)}, pages 4129--4138, 2019.

\bibitem[Ho et~al.(2022)Ho, Saxe, and Cushman]{ho2022planning}
M.~K. Ho, R.~Saxe, and F.~Cushman.
\newblock Planning with theory of mind.
\newblock \emph{Trends in Cognitive Sciences}, 26\penalty0 (11):\penalty0 959--971, 2022.

\bibitem[Holdgraf et~al.(2017)Holdgraf, Rieger, Micheli, Martin, Knight, and Theunissen]{holdgraf2017encoding}
C.~R. Holdgraf, J.~W. Rieger, C.~Micheli, S.~Martin, R.~T. Knight, and F.~E. Theunissen.
\newblock Encoding and decoding models in cognitive electrophysiology.
\newblock \emph{Frontiers in systems neuroscience}, 11:\penalty0 61, 2017.

\bibitem[Hollerman and Schultz(1998)]{hollerman1998dopamine}
J.~R. Hollerman and W.~Schultz.
\newblock Dopamine neurons report an error in the temporal prediction of reward during learning.
\newblock \emph{Nature neuroscience}, 1\penalty0 (4):\penalty0 304--309, 1998.

\bibitem[Hoogland et~al.(2024)Hoogland, Wang, Farrugia-Roberts, Carroll, Wei, and Murfet]{hoogland2024developmental}
J.~Hoogland, G.~Wang, M.~Farrugia-Roberts, L.~Carroll, S.~Wei, and D.~Murfet.
\newblock The developmental landscape of in-context learning, 2024.

\bibitem[Houlihan et~al.(2023)Houlihan, Kleiman-Weiner, Hewitt, Tenenbaum, and Saxe]{houlihan2023emotion}
S.~D. Houlihan, M.~Kleiman-Weiner, L.~B. Hewitt, J.~B. Tenenbaum, and R.~Saxe.
\newblock Emotion prediction as computation over a generative theory of mind.
\newblock \emph{Philosophical Transactions of the Royal Society A}, 381\penalty0 (2251):\penalty0 20220047, 2023.

\bibitem[Huang et~al.(2011)Huang, Haith, Mazzoni, and Krakauer]{huang2011rethinking}
V.~S. Huang, A.~Haith, P.~Mazzoni, and J.~W. Krakauer.
\newblock Rethinking motor learning and savings in adaptation paradigms: model-free memory for successful actions combines with internal models.
\newblock \emph{Neuron}, 70\penalty0 (4):\penalty0 787--801, 2011.

\bibitem[Hubel and Wiesel(1962)]{hubel1962receptive}
D.~H. Hubel and T.~N. Wiesel.
\newblock Receptive fields, binocular interaction and functional architecture in the cat's visual cortex.
\newblock \emph{The Journal of physiology}, 160\penalty0 (1):\penalty0 106, 1962.

\bibitem[Hubinger et~al.(2024)Hubinger, Denison, Mu, Lambert, Tong, MacDiarmid, Lanham, Ziegler, Maxwell, Cheng, Jermyn, Askell, Radhakrishnan, Anil, Duvenaud, Ganguli, Barez, Clark, Ndousse, Sachan, Sellitto, Sharma, DasSarma, Grosse, Kravec, Bai, Witten, Favaro, Brauner, Karnofsky, Christiano, Bowman, Graham, Kaplan, Mindermann, Greenblatt, Shlegeris, Schiefer, and Perez]{hubinger2024sleeper}
E.~Hubinger, C.~Denison, J.~Mu, M.~Lambert, M.~Tong, M.~MacDiarmid, T.~Lanham, D.~M. Ziegler, T.~Maxwell, N.~Cheng, A.~Jermyn, A.~Askell, A.~Radhakrishnan, C.~Anil, D.~Duvenaud, D.~Ganguli, F.~Barez, J.~Clark, K.~Ndousse, K.~Sachan, M.~Sellitto, M.~Sharma, N.~DasSarma, R.~Grosse, S.~Kravec, Y.~Bai, Z.~Witten, M.~Favaro, J.~Brauner, H.~Karnofsky, P.~Christiano, S.~R. Bowman, L.~Graham, J.~Kaplan, S.~Mindermann, R.~Greenblatt, B.~Shlegeris, N.~Schiefer, and E.~Perez.
\newblock Sleeper agents: Training deceptive llms that persist through safety training, 2024.

\bibitem[Huff et~al.(2018)Huff, Mahabadi, and Tadi]{huff2018neuroanatomy}
T.~Huff, N.~Mahabadi, and P.~Tadi.
\newblock \emph{Neuroanatomy, visual cortex}.
\newblock StatPearls Publishing, 2018.

\bibitem[Hulse et~al.(2018)Hulse, Fowler, and Honig]{hulse2018cognitive}
S.~H. Hulse, H.~Fowler, and W.~K. Honig.
\newblock \emph{Cognitive processes in animal behavior}.
\newblock Routledge, 2018.

\bibitem[Huth et~al.(2016)Huth, De~Heer, Griffiths, Theunissen, and Gallant]{huth2016natural}
A.~G. Huth, W.~A. De~Heer, T.~L. Griffiths, F.~E. Theunissen, and J.~L. Gallant.
\newblock Natural speech reveals the semantic maps that tile human cerebral cortex.
\newblock \emph{Nature}, 532\penalty0 (7600):\penalty0 453--458, 2016.

\bibitem[Ito et~al.(2015)Ito, Zhang, Witter, Moser, and Moser]{ito2015prefrontal}
H.~T. Ito, S.-J. Zhang, M.~P. Witter, E.~I. Moser, and M.-B. Moser.
\newblock A prefrontal--thalamo--hippocampal circuit for goal-directed spatial navigation.
\newblock \emph{Nature}, 522\penalty0 (7554):\penalty0 50--55, 2015.

\bibitem[Ivanova et~al.(2021)Ivanova, Hewitt, and Zaslavsky]{ivanova2021probing}
A.~A. Ivanova, J.~Hewitt, and N.~Zaslavsky.
\newblock Probing artificial neural networks: Insights from neuroscience.
\newblock In \emph{{ICLR} 2021 Workshop ``How Can Findings About The Brain Improve AI Systems{?}''}, 2021.

\bibitem[Ivanova et~al.(2022)Ivanova, Schrimpf, Anzellotti, Zaslavsky, Fedorenko, and Isik]{ivanova2022beyond}
A.~A. Ivanova, M.~Schrimpf, S.~Anzellotti, N.~Zaslavsky, E.~Fedorenko, and L.~Isik.
\newblock Beyond linear regression: mapping models in cognitive neuroscience should align with research goals.
\newblock \emph{Neurons, Behavior, Data Analysis, and Theory}, 8 2022.

\bibitem[Izawa and Shadmehr(2011)]{izawa2011learning}
J.~Izawa and R.~Shadmehr.
\newblock Learning from sensory and reward prediction errors during motor adaptation.
\newblock \emph{PLoS computational biology}, 7\penalty0 (3):\penalty0 e1002012, 2011.

\bibitem[Jeffress(1948)]{Jeffress1948}
L.~A. Jeffress.
\newblock A place theory of sound localization.
\newblock \emph{Journal of Comparative and Physiological Psychology}, 41\penalty0 (1):\penalty0 35--39, 1948.
\newblock \doi{10.1037/h0061495}.

\bibitem[Ji et~al.(2023)Ji, Lee, Frieske, Yu, Su, Xu, Ishii, Bang, Madotto, and Fung]{Ji_2023}
Z.~Ji, N.~Lee, R.~Frieske, T.~Yu, D.~Su, Y.~Xu, E.~Ishii, Y.~J. Bang, A.~Madotto, and P.~Fung.
\newblock Survey of hallucination in natural language generation.
\newblock \emph{ACM Computing Surveys}, 55\penalty0 (12):\penalty0 1–38, Mar. 2023.
\newblock ISSN 1557-7341.
\newblock \doi{10.1145/3571730}.
\newblock URL \url{http://dx.doi.org/10.1145/3571730}.

\bibitem[Johnson(2017)]{johnson2017marr}
M.~Johnson.
\newblock Marr’s levels and the minimalist program.
\newblock \emph{Psychonomic bulletin \& review}, 24:\penalty0 171--174, 2017.

\bibitem[Johnston and Fusi(2023)]{johnston2023abstract}
W.~J. Johnston and S.~Fusi.
\newblock Abstract representations emerge naturally in neural networks trained to perform multiple tasks.
\newblock \emph{Nature Communications}, 14\penalty0 (1):\penalty0 1040, 2023.

\bibitem[Johnston et~al.(2020)Johnston, Palmer, and Freedman]{johnston2020nonlinear}
W.~J. Johnston, S.~E. Palmer, and D.~J. Freedman.
\newblock Nonlinear mixed selectivity supports reliable neural computation.
\newblock \emph{PLoS computational biology}, 16\penalty0 (2):\penalty0 e1007544, 2020.

\bibitem[Jonas and Kording(2017)]{jonas2017could}
E.~Jonas and K.~P. Kording.
\newblock Could a neuroscientist understand a microprocessor?
\newblock \emph{PLoS computational biology}, 13\penalty0 (1):\penalty0 e1005268, 2017.

\bibitem[Jordan et~al.(2019)Jordan, Petrovici, Breitwieser, Schemmel, Meier, Diesmann, and Tetzlaff]{jordan2019deterministic}
J.~Jordan, M.~A. Petrovici, O.~Breitwieser, J.~Schemmel, K.~Meier, M.~Diesmann, and T.~Tetzlaff.
\newblock Deterministic networks for probabilistic computing.
\newblock \emph{Scientific reports}, 9\penalty0 (1):\penalty0 18303, 2019.

\bibitem[Josselyn and Tonegawa(2020)]{josselyn2020memory}
S.~A. Josselyn and S.~Tonegawa.
\newblock Memory engrams: Recalling the past and imagining the future.
\newblock \emph{Science}, 367\penalty0 (6473):\penalty0 eaaw4325, 2020.

\bibitem[Josselyn et~al.(2015)Josselyn, K{\"o}hler, and Frankland]{josselyn2015finding}
S.~A. Josselyn, S.~K{\"o}hler, and P.~W. Frankland.
\newblock Finding the engram.
\newblock \emph{Nature Reviews Neuroscience}, 16\penalty0 (9):\penalty0 521--534, 2015.

\bibitem[Kaelbling et~al.(1998)Kaelbling, Littman, and Cassandra]{kaelbling1998planning}
L.~Kaelbling, M.~Littman, and A.~Cassandra.
\newblock Planning and acting in partially observable domains in: Artificial intelligence.
\newblock \emph{Artificial intelligence}, 1998.

\bibitem[Kanwisher et~al.(1997)Kanwisher, McDermott, and Chun]{kanwisher1997fusiform}
N.~Kanwisher, J.~McDermott, and M.~M. Chun.
\newblock The fusiform face area: a module in human extrastriate cortex specialized for face perception.
\newblock \emph{Journal of neuroscience}, 17\penalty0 (11):\penalty0 4302--4311, 1997.

\bibitem[Kaplan et~al.(2020)Kaplan, McCandlish, Henighan, Brown, Chess, Child, Gray, Radford, Wu, and Amodei]{kaplan2020scaling}
J.~Kaplan, S.~McCandlish, T.~Henighan, T.~B. Brown, B.~Chess, R.~Child, S.~Gray, A.~Radford, J.~Wu, and D.~Amodei.
\newblock Scaling laws for neural language models.
\newblock \emph{arXiv preprint arXiv:2001.08361}, 2020.

\bibitem[Karim et~al.(2010)Karim, Schneider, Lotze, Veit, Sauseng, Braun, and Birbaumer]{karim2010truth}
A.~A. Karim, M.~Schneider, M.~Lotze, R.~Veit, P.~Sauseng, C.~Braun, and N.~Birbaumer.
\newblock The truth about lying: inhibition of the anterior prefrontal cortex improves deceptive behavior.
\newblock \emph{Cerebral cortex}, 20\penalty0 (1):\penalty0 205--213, 2010.

\bibitem[Karmiloff-Smith(1994)]{karmiloff1994beyond}
B.~A. Karmiloff-Smith.
\newblock Beyond modularity: A developmental perspective on cognitive science.
\newblock \emph{European journal of disorders of communication}, 29\penalty0 (1):\penalty0 95--105, 1994.

\bibitem[Kashtan and Alon(2005)]{kashtan2005spontaneous}
N.~Kashtan and U.~Alon.
\newblock Spontaneous evolution of modularity and network motifs.
\newblock \emph{Proceedings of the National Academy of Sciences}, 102\penalty0 (39):\penalty0 13773--13778, 2005.
\newblock \doi{10.1073/pnas.0503610102}.

\bibitem[Kauf et~al.(2024)Kauf, Tuckute, Levy, Andreas, and Fedorenko]{kauf2024lexical}
C.~Kauf, G.~Tuckute, R.~Levy, J.~Andreas, and E.~Fedorenko.
\newblock Lexical-semantic content, not syntactic structure, is the main contributor to ann-brain similarity of fmri responses in the language network.
\newblock \emph{Neurobiology of Language}, 5\penalty0 (1):\penalty0 7--42, 2024.

\bibitem[Kay et~al.(2008)Kay, Naselaris, Prenger, and Gallant]{kay2008identifying}
K.~N. Kay, T.~Naselaris, R.~J. Prenger, and J.~L. Gallant.
\newblock Identifying natural images from human brain activity.
\newblock \emph{Nature}, 452\penalty0 (7185):\penalty0 352--355, 2008.

\bibitem[Kersten and Yuille(2003)]{kersten2003bayesian}
D.~Kersten and A.~Yuille.
\newblock Bayesian models of object perception.
\newblock \emph{Current opinion in neurobiology}, 13\penalty0 (2):\penalty0 150--158, 2003.

\bibitem[Khaligh-Razavi et~al.(2017)Khaligh-Razavi, Henriksson, Kay, and Kriegeskorte]{khaligh2017fixed}
S.-M. Khaligh-Razavi, L.~Henriksson, K.~Kay, and N.~Kriegeskorte.
\newblock Fixed versus mixed rsa: Explaining visual representations by fixed and mixed feature sets from shallow and deep computational models.
\newblock \emph{Journal of Mathematical Psychology}, 76:\penalty0 184--197, 2017.

\bibitem[Kim et~al.(2017)Kim, Jeffery, and Maguire]{kim2017multivoxel}
M.~Kim, K.~J. Jeffery, and E.~A. Maguire.
\newblock Multivoxel pattern analysis reveals 3d place information in the human hippocampus.
\newblock \emph{Journal of Neuroscience}, 37\penalty0 (16):\penalty0 4270--4279, 2017.

\bibitem[Kobak et~al.(2016)Kobak, Brendel, Constantinidis, Feierstein, Kepecs, Mainen, Qi, Romo, Uchida, and Machens]{kobak2016demixed}
D.~Kobak, W.~Brendel, C.~Constantinidis, C.~E. Feierstein, A.~Kepecs, Z.~F. Mainen, X.-L. Qi, R.~Romo, N.~Uchida, and C.~K. Machens.
\newblock Demixed principal component analysis of neural population data.
\newblock \emph{elife}, 5:\penalty0 e10989, 2016.

\bibitem[K{\"o}rding and Wolpert(2004)]{kording2004bayesian}
K.~P. K{\"o}rding and D.~M. Wolpert.
\newblock Bayesian integration in sensorimotor learning.
\newblock \emph{Nature}, 427\penalty0 (6971):\penalty0 244--247, 2004.

\bibitem[Kornblith et~al.(2019)Kornblith, Norouzi, Lee, and Hinton]{kornblith2019similarity}
S.~Kornblith, M.~Norouzi, H.~Lee, and G.~Hinton.
\newblock Similarity of neural network representations revisited.
\newblock In \emph{International conference on machine learning}, pages 3519--3529. PMLR, 2019.

\bibitem[Krakauer et~al.(2017)Krakauer, Ghazanfar, Gomez-Marin, MacIver, and Poeppel]{Krakauer2017}
J.~W. Krakauer, A.~A. Ghazanfar, A.~Gomez-Marin, M.~A. MacIver, and D.~Poeppel.
\newblock Neuroscience needs behavior: Correcting a reductionist bias.
\newblock \emph{Neuron}, 93\penalty0 (3):\penalty0 480--490, 2017.
\newblock \doi{10.1016/j.neuron.2016.12.041}.

\bibitem[Kriegeskorte and Douglas(2019)]{kriegeskorte2019interpreting}
N.~Kriegeskorte and P.~K. Douglas.
\newblock Interpreting encoding and decoding models.
\newblock \emph{Current opinion in neurobiology}, 55:\penalty0 167--179, 2019.

\bibitem[Kriegeskorte and Kievit(2013)]{kriegeskorte2013representational}
N.~Kriegeskorte and R.~A. Kievit.
\newblock Representational geometry: integrating cognition, computation, and the brain.
\newblock \emph{Trends in cognitive sciences}, 17\penalty0 (8):\penalty0 401--412, 2013.

\bibitem[Kriegeskorte and Wei(2021)]{kriegeskorte2021neural}
N.~Kriegeskorte and X.-X. Wei.
\newblock Neural tuning and representational geometry.
\newblock \emph{Nature Reviews Neuroscience}, 22\penalty0 (11):\penalty0 703--718, 2021.

\bibitem[Lagnado et~al.(2013)Lagnado, Gerstenberg, and Zultan]{lagnado2013causal}
D.~A. Lagnado, T.~Gerstenberg, and R.~Zultan.
\newblock Causal responsibility and counterfactuals.
\newblock \emph{Cognitive science}, 37\penalty0 (6):\penalty0 1036--1073, 2013.

\bibitem[Lake et~al.(2015)Lake, Salakhutdinov, and Tenenbaum]{lake2015human}
B.~M. Lake, R.~Salakhutdinov, and J.~B. Tenenbaum.
\newblock Human-level concept learning through probabilistic program induction.
\newblock \emph{Science}, 350\penalty0 (6266):\penalty0 1332--1338, 2015.

\bibitem[Lake et~al.(2017)Lake, Ullman, Tenenbaum, and Gershman]{lake2017building}
B.~M. Lake, T.~D. Ullman, J.~B. Tenenbaum, and S.~J. Gershman.
\newblock Building machines that learn and think like people.
\newblock \emph{Behavioral and brain sciences}, 40:\penalty0 e253, 2017.

\bibitem[Lakretz et~al.(2019)Lakretz, Kruszewski, Desbordes, Hupkes, Dehaene, and Baroni]{lakretz2019emergence}
Y.~Lakretz, G.~Kruszewski, T.~Desbordes, D.~Hupkes, S.~Dehaene, and M.~Baroni.
\newblock The emergence of number and syntax units in lstm language models.
\newblock \emph{arXiv preprint arXiv:1903.07435}, 2019.

\bibitem[Lamme and Roelfsema(2000)]{lamme2000distinct}
V.~A. Lamme and P.~R. Roelfsema.
\newblock The distinct modes of vision offered by feedforward and recurrent processing.
\newblock \emph{Trends in neurosciences}, 23\penalty0 (11):\penalty0 571--579, 2000.

\bibitem[Landau et~al.(1992)Landau, Smith, and Jones]{landau1992syntactic}
B.~Landau, L.~B. Smith, and S.~Jones.
\newblock Syntactic context and the shape bias in children's and adults' lexical learning.
\newblock \emph{Journal of Memory and Language}, 31\penalty0 (6):\penalty0 807--825, 1992.

\bibitem[Laughlin and Sejnowski(2003)]{laughlin2003communication}
S.~B. Laughlin and T.~J. Sejnowski.
\newblock Communication in neuronal networks.
\newblock \emph{Science}, 301\penalty0 (5641):\penalty0 1870--1874, 2003.

\bibitem[Lazebnik(2002)]{lazebnik2002can}
Y.~Lazebnik.
\newblock Can a biologist fix a radio?—or, what i learned while studying apoptosis.
\newblock \emph{Cancer cell}, 2\penalty0 (3):\penalty0 179--182, 2002.

\bibitem[Lee et~al.(2024)Lee, Bai, Pres, Wattenberg, Kummerfeld, and Mihalcea]{lee2024mechanistic}
A.~Lee, X.~Bai, I.~Pres, M.~Wattenberg, J.~K. Kummerfeld, and R.~Mihalcea.
\newblock A mechanistic understanding of alignment algorithms: A case study on dpo and toxicity.
\newblock \emph{arXiv preprint arXiv:2401.01967}, 2024.

\bibitem[Leininger and Kelley(2015)]{leininger2015evolution}
E.~C. Leininger and D.~B. Kelley.
\newblock Evolution of courtship songs in xenopus: vocal pattern generation and sound production.
\newblock \emph{Cytogenetic and genome research}, 145\penalty0 (3-4):\penalty0 302--314, 2015.

\bibitem[Li et~al.(2024)Li, Patel, Vi{\'e}gas, Pfister, and Wattenberg]{li2024inference}
K.~Li, O.~Patel, F.~Vi{\'e}gas, H.~Pfister, and M.~Wattenberg.
\newblock Inference-time intervention: Eliciting truthful answers from a language model.
\newblock \emph{Advances in Neural Information Processing Systems}, 36, 2024.

\bibitem[Lieberum et~al.(2023)Lieberum, Rahtz, Kram{\'a}r, Irving, Shah, and Mikulik]{lieberum2023does}
T.~Lieberum, M.~Rahtz, J.~Kram{\'a}r, G.~Irving, R.~Shah, and V.~Mikulik.
\newblock Does circuit analysis interpretability scale? evidence from multiple choice capabilities in chinchilla.
\newblock \emph{arXiv preprint arXiv:2307.09458}, 2023.

\bibitem[Lillicrap and Kording(2019)]{lillicrap2019does}
T.~P. Lillicrap and K.~P. Kording.
\newblock What does it mean to understand a neural network?
\newblock \emph{arXiv preprint arXiv:1907.06374}, 2019.

\bibitem[Lillicrap et~al.(2020)Lillicrap, Santoro, Marris, Akerman, and Hinton]{lillicrap2020backpropagation}
T.~P. Lillicrap, A.~Santoro, L.~Marris, C.~J. Akerman, and G.~Hinton.
\newblock Backpropagation and the brain.
\newblock \emph{Nature Reviews Neuroscience}, 21\penalty0 (6):\penalty0 335--346, 2020.

\bibitem[Lindsay and Bau(2023)]{lindsay2023testing}
G.~W. Lindsay and D.~Bau.
\newblock Testing methods of neural systems understanding.
\newblock \emph{Cognitive Systems Research}, 82:\penalty0 101156, 2023.

\bibitem[Lindsay et~al.(2022)Lindsay, Mrsic-Flogel, and Sahani]{lindsay2022bio}
G.~W. Lindsay, T.~D. Mrsic-Flogel, and M.~Sahani.
\newblock Bio-inspired neural networks implement different recurrent visual processing strategies than task-trained ones do.
\newblock \emph{bioRxiv}, pages 2022--03, 2022.

\bibitem[Lisman(2015)]{lisman2015challenge}
J.~Lisman.
\newblock The challenge of understanding the brain: where we stand in 2015.
\newblock \emph{Neuron}, 86\penalty0 (4):\penalty0 864--882, 2015.

\bibitem[Liu et~al.(2022{\natexlab{a}})Liu, Kitouni, Nolte, Michaud, Tegmark, and Williams]{liu2022towards}
Z.~Liu, O.~Kitouni, N.~S. Nolte, E.~Michaud, M.~Tegmark, and M.~Williams.
\newblock Towards understanding grokking: An effective theory of representation learning.
\newblock \emph{Advances in Neural Information Processing Systems}, 35:\penalty0 34651--34663, 2022{\natexlab{a}}.

\bibitem[Liu et~al.(2022{\natexlab{b}})Liu, Michaud, and Tegmark]{liu2022omnigrok}
Z.~Liu, E.~J. Michaud, and M.~Tegmark.
\newblock Omnigrok: Grokking beyond algorithmic data.
\newblock In \emph{The Eleventh International Conference on Learning Representations}, 2022{\natexlab{b}}.

\bibitem[Liu et~al.(2023)Liu, Khona, Fiete, and Tegmark]{liu2023growing}
Z.~Liu, M.~Khona, I.~R. Fiete, and M.~Tegmark.
\newblock Growing brains: Co-emergence of anatomical and functional modularity in recurrent neural networks.
\newblock \emph{arXiv preprint arXiv:2310.07711v1}, 2023.

\bibitem[Lourie et~al.(2023)Lourie, Cho, and He]{lourie2023show}
N.~Lourie, K.~Cho, and H.~He.
\newblock Show your work with confidence: Confidence bands for tuning curves.
\newblock \emph{arXiv preprint arXiv:2311.09480}, 2023.

\bibitem[Love(2015)]{love2015algorithmic}
B.~C. Love.
\newblock The algorithmic level is the bridge between computation and brain.
\newblock \emph{Topics in cognitive science}, 7\penalty0 (2):\penalty0 230--242, 2015.

\bibitem[Luber et~al.(2009)Luber, Fisher, Appelbaum, Ploesser, and Lisanby]{luber2009non}
B.~Luber, C.~Fisher, P.~S. Appelbaum, M.~Ploesser, and S.~H. Lisanby.
\newblock Non-invasive brain stimulation in the detection of deception: Scientific challenges and ethical consequences.
\newblock \emph{Behavioral sciences \& the law}, 27\penalty0 (2):\penalty0 191--208, 2009.

\bibitem[Mahmoudi et~al.(2012)Mahmoudi, Takerkart, Regragui, Boussaoud, and Brovelli]{mahmoudi2012multivoxel}
A.~Mahmoudi, S.~Takerkart, F.~Regragui, D.~Boussaoud, and A.~Brovelli.
\newblock Multivoxel pattern analysis for fmri data: a review.
\newblock \emph{Computational and mathematical methods in medicine}, 2012\penalty0 (1):\penalty0 961257, 2012.

\bibitem[Mahon and Cantlon(2011)]{mahon2011specialization}
B.~Z. Mahon and J.~F. Cantlon.
\newblock The specialization of function: Cognitive and neural perspectives.
\newblock \emph{Cognitive neuropsychology}, 28\penalty0 (3-4):\penalty0 147--155, 2011.

\bibitem[Mante et~al.(2013)Mante, Sussillo, Shenoy, and Newsome]{mante2013context}
V.~Mante, D.~Sussillo, K.~V. Shenoy, and W.~T. Newsome.
\newblock Context-dependent computation by recurrent dynamics in prefrontal cortex.
\newblock \emph{nature}, 503\penalty0 (7474):\penalty0 78--84, 2013.

\bibitem[Margalit et~al.(2024)Margalit, Lee, Finzi, DiCarlo, Grill-Spector, and Yamins]{margalit2024unifying}
E.~Margalit, H.~Lee, D.~Finzi, J.~J. DiCarlo, K.~Grill-Spector, and D.~L. Yamins.
\newblock A unifying framework for functional organization in early and higher ventral visual cortex.
\newblock \emph{Neuron}, 2024.

\bibitem[Marr(2010)]{marr2010vision}
D.~Marr.
\newblock \emph{Vision: A computational investigation into the human representation and processing of visual information}.
\newblock MIT press, 2010.

\bibitem[McClamrock(1991)]{mcclamrock1991marr}
R.~McClamrock.
\newblock Marr's three levels: A re-evaluation.
\newblock \emph{Minds and Machines}, 1:\penalty0 185--196, 1991.

\bibitem[McCoy et~al.(2023)McCoy, Yao, Friedman, Hardy, and Griffiths]{mccoy2023embers}
R.~T. McCoy, S.~Yao, D.~Friedman, M.~Hardy, and T.~L. Griffiths.
\newblock Embers of autoregression: Understanding large language models through the problem they are trained to solve.
\newblock \emph{arXiv preprint arXiv:2309.13638}, 2023.

\bibitem[McDougall et~al.(2023)McDougall, Conmy, Rushing, McGrath, and Nanda]{mcdougall2023copy}
C.~McDougall, A.~Conmy, C.~Rushing, T.~McGrath, and N.~Nanda.
\newblock Copy suppression: Comprehensively understanding an attention head.
\newblock \emph{arXiv preprint arXiv:2310.04625}, 2023.

\bibitem[McGrath et~al.(2023)McGrath, Rahtz, Kramar, Mikulik, and Legg]{mcgrath2023hydra}
T.~McGrath, M.~Rahtz, J.~Kramar, V.~Mikulik, and S.~Legg.
\newblock The hydra effect: Emergent self-repair in language model computations, 2023.

\bibitem[Meng et~al.(2022)Meng, Bau, Andonian, and Belinkov]{meng2022locating}
K.~Meng, D.~Bau, A.~Andonian, and Y.~Belinkov.
\newblock Locating and editing factual associations in gpt.
\newblock \emph{Advances in Neural Information Processing Systems}, 35:\penalty0 17359--17372, 2022.

\bibitem[Meunier et~al.(2010)Meunier, Lambiotte, and Bullmore]{MeunierLambiotteBullmore2010}
D.~Meunier, R.~Lambiotte, and E.~T. Bullmore.
\newblock Modular and hierarchically modular organization of brain networks.
\newblock \emph{Frontiers in Neuroscience}, 4:\penalty0 200, 2010.
\newblock \doi{10.3389/fnins.2010.00200}.

\bibitem[Michaels et~al.(2016)Michaels, Dann, and Scherberger]{michaels2016neural}
J.~A. Michaels, B.~Dann, and H.~Scherberger.
\newblock Neural population dynamics during reaching are better explained by a dynamical system than representational tuning.
\newblock \emph{PLoS computational biology}, 12\penalty0 (11):\penalty0 e1005175, 2016.

\bibitem[Mikolov et~al.(2013)Mikolov, Sutskever, Chen, Corrado, and Dean]{mikolov2013distributed}
T.~Mikolov, I.~Sutskever, K.~Chen, G.~S. Corrado, and J.~Dean.
\newblock Distributed representations of words and phrases and their compositionality.
\newblock \emph{Advances in neural information processing systems}, 26, 2013.

\bibitem[Miller et~al.(2024)Miller, Brincat, and Roy]{miller2024cognition}
E.~K. Miller, S.~L. Brincat, and J.~E. Roy.
\newblock Cognition is an emergent property.
\newblock \emph{Current Opinion in Behavioral Sciences}, 57:\penalty0 101388, 2024.

\bibitem[Mitchell(2006)]{mitchell2006mentalizing}
J.~P. Mitchell.
\newblock Mentalizing and marr: an information processing approach to the study of social cognition.
\newblock \emph{Brain research}, 1079\penalty0 (1):\penalty0 66--75, 2006.

\bibitem[Mitchell et~al.(2008)Mitchell, Shinkareva, Carlson, Chang, Malave, Mason, and Just]{mitchell2008predicting}
T.~M. Mitchell, S.~V. Shinkareva, A.~Carlson, K.-M. Chang, V.~L. Malave, R.~A. Mason, and M.~A. Just.
\newblock Predicting human brain activity associated with the meanings of nouns.
\newblock \emph{science}, 320\penalty0 (5880):\penalty0 1191--1195, 2008.

\bibitem[Nanda et~al.(2023)Nanda, Chan, Lieberum, Smith, and Steinhardt]{nanda2023progress}
N.~Nanda, L.~Chan, T.~Lieberum, J.~Smith, and J.~Steinhardt.
\newblock Progress measures for grokking via mechanistic interpretability.
\newblock \emph{arXiv preprint arXiv:2301.05217}, 2023.

\bibitem[Narayanan et~al.(2022)Narayanan, Deshmukh, Dogan, and Balasubramanian]{narayanan2022challenges}
V.~Narayanan, A.~A. Deshmukh, U.~Dogan, and V.~N. Balasubramanian.
\newblock On challenges in unsupervised domain generalization.
\newblock In \emph{NeurIPS 2021 Workshop on Pre-registration in Machine Learning}, pages 42--58. PMLR, 2022.

\bibitem[Naselaris et~al.(2011)Naselaris, Kay, Nishimoto, and Gallant]{naselaris2011encoding}
T.~Naselaris, K.~N. Kay, S.~Nishimoto, and J.~L. Gallant.
\newblock Encoding and decoding in fmri.
\newblock \emph{Neuroimage}, 56\penalty0 (2):\penalty0 400--410, 2011.

\bibitem[Naselaris et~al.(2015)Naselaris, Olman, Stansbury, Ugurbil, and Gallant]{naselaris2015voxel}
T.~Naselaris, C.~A. Olman, D.~E. Stansbury, K.~Ugurbil, and J.~L. Gallant.
\newblock A voxel-wise encoding model for early visual areas decodes mental images of remembered scenes.
\newblock \emph{Neuroimage}, 105:\penalty0 215--228, 2015.

\bibitem[Niv and Langdon(2016)]{niv2016reinforcement}
Y.~Niv and A.~Langdon.
\newblock Reinforcement learning with marr.
\newblock \emph{Current opinion in behavioral sciences}, 11:\penalty0 67--73, 2016.

\bibitem[Nogueira et~al.(2023)Nogueira, Rodgers, Bruno, and Fusi]{nogueira2023geometry}
R.~Nogueira, C.~C. Rodgers, R.~M. Bruno, and S.~Fusi.
\newblock The geometry of cortical representations of touch in rodents.
\newblock \emph{Nature Neuroscience}, 26\penalty0 (2):\penalty0 239--250, 2023.

\bibitem[Norman et~al.(2006)Norman, Polyn, Detre, and Haxby]{norman2006beyond}
K.~A. Norman, S.~M. Polyn, G.~J. Detre, and J.~V. Haxby.
\newblock Beyond mind-reading: multi-voxel pattern analysis of fmri data.
\newblock \emph{Trends in cognitive sciences}, 10\penalty0 (9):\penalty0 424--430, 2006.

\bibitem[Norman-Haignere et~al.(2015)Norman-Haignere, Kanwisher, and McDermott]{norman2015distinct}
S.~Norman-Haignere, N.~G. Kanwisher, and J.~H. McDermott.
\newblock Distinct cortical pathways for music and speech revealed by hypothesis-free voxel decomposition.
\newblock \emph{neuron}, 88\penalty0 (6):\penalty0 1281--1296, 2015.

\bibitem[O'Doherty et~al.(2003)O'Doherty, Dayan, Friston, Critchley, and Dolan]{o2003temporal}
J.~P. O'Doherty, P.~Dayan, K.~Friston, H.~Critchley, and R.~J. Dolan.
\newblock Temporal difference models and reward-related learning in the human brain.
\newblock \emph{Neuron}, 38\penalty0 (2):\penalty0 329--337, 2003.

\bibitem[O'Keefe(1978)]{okeefe1978hippocampus}
J.~O'Keefe.
\newblock \emph{The Hippocampus as a Cognitive Map}.
\newblock Clarendon Press, 1978.
\newblock ISBN 978-0198572060.

\bibitem[Olah(2021)]{olah2021interpretability}
C.~Olah.
\newblock Interpretability vs neuroscience [rough note].
\newblock \url{https://colah.github.io/notes/interp-v-neuro/}, 2021.
\newblock Accessed: [Insert today's date here].

\bibitem[Olah and Jermyn(2023)]{OlahJermyn2023}
C.~Olah and A.~Jermyn.
\newblock July 2023 update: Safety features.
\newblock \url{https://transformer-circuits.pub/2023/july-update/index.html#safety-features}, 2023.

\bibitem[Olah et~al.(2020)Olah, Cammarata, Schubert, Goh, Petrov, and Carter]{olah2020zoom}
C.~Olah, N.~Cammarata, L.~Schubert, G.~Goh, M.~Petrov, and S.~Carter.
\newblock Zoom in: An introduction to circuits.
\newblock \emph{Distill}, 2020.
\newblock \doi{10.23915/distill.00024.001}.
\newblock https://distill.pub/2020/circuits/zoom-in.

\bibitem[Olsson et~al.(2022)Olsson, Elhage, Nanda, Joseph, DasSarma, Henighan, Mann, Askell, Bai, Chen, et~al.]{olsson2022context}
C.~Olsson, N.~Elhage, N.~Nanda, N.~Joseph, N.~DasSarma, T.~Henighan, B.~Mann, A.~Askell, Y.~Bai, A.~Chen, et~al.
\newblock In-context learning and induction heads.
\newblock \emph{arXiv preprint arXiv:2209.11895}, 2022.

\bibitem[Park et~al.(2024)Park, Goldstein, O’Gara, Chen, and Hendrycks]{park2024ai}
P.~S. Park, S.~Goldstein, A.~O’Gara, M.~Chen, and D.~Hendrycks.
\newblock Ai deception: A survey of examples, risks, and potential solutions.
\newblock \emph{Patterns}, 5\penalty0 (5), 2024.

\bibitem[Pearl(1988)]{pearl1988probabilistic}
J.~Pearl.
\newblock \emph{Probabilistic reasoning in intelligent systems: networks of plausible inference}.
\newblock Morgan kaufmann, 1988.

\bibitem[Peebles and Cooper(2015)]{peebles2015thirty}
D.~Peebles and R.~P. Cooper.
\newblock Thirty years after marr's vision: levels of analysis in cognitive science, 2015.

\bibitem[Perez-Nieves et~al.(2021)Perez-Nieves, Leung, Dragotti, and Goodman]{perez2021neural}
N.~Perez-Nieves, V.~C. Leung, P.~L. Dragotti, and D.~F. Goodman.
\newblock Neural heterogeneity promotes robust learning.
\newblock \emph{Nature communications}, 12\penalty0 (1):\penalty0 5791, 2021.

\bibitem[Pillow(2024)]{pillow2024cross}
J.~W. Pillow.
\newblock Cross talk opposing view: Marr's three levels of analysis are not useful as a framework for neuroscience.
\newblock \emph{The Journal of Physiology}, 602\penalty0 (9):\penalty0 1915--1917, 2024.

\bibitem[Pimentel et~al.(2020)Pimentel, Valvoda, Maudslay, Zmigrod, Williams, and Cotterell]{pimentel2020information}
T.~Pimentel, J.~Valvoda, R.~H. Maudslay, R.~Zmigrod, A.~Williams, and R.~Cotterell.
\newblock Information-theoretic probing for linguistic structure.
\newblock \emph{arXiv preprint arXiv:2004.03061}, 2020.

\bibitem[Poeppel and Adolfi(2020)]{poeppel2020against}
D.~Poeppel and F.~Adolfi.
\newblock Against the epistemological primacy of the hardware: The brain from inside out, turned upside down.
\newblock \emph{Eneuro}, 7\penalty0 (4), 2020.

\bibitem[Poggio(2012)]{poggio2012levels}
T.~Poggio.
\newblock The levels of understanding framework, revised.
\newblock \emph{Perception}, 41\penalty0 (9):\penalty0 1017--1023, 2012.

\bibitem[Pouget et~al.(2000)Pouget, Dayan, and Zemel]{pouget2000information}
A.~Pouget, P.~Dayan, and R.~Zemel.
\newblock Information processing with population codes.
\newblock \emph{Nature Reviews Neuroscience}, 1\penalty0 (2):\penalty0 125--132, 2000.

\bibitem[Power et~al.(2022)Power, Burda, Edwards, Babuschkin, and Misra]{power2022grokking}
A.~Power, Y.~Burda, H.~Edwards, I.~Babuschkin, and V.~Misra.
\newblock Grokking: Generalization beyond overfitting on small algorithmic datasets.
\newblock \emph{arXiv preprint arXiv:2201.02177}, 2022.

\bibitem[Prins et~al.(2016)]{prins2016psychophysics}
N.~Prins et~al.
\newblock \emph{Psychophysics: a practical introduction}.
\newblock Academic Press, 2016.

\bibitem[Prinz et~al.(2004)Prinz, Bucher, and Marder]{prinz2004similar}
A.~A. Prinz, D.~Bucher, and E.~Marder.
\newblock Similar network activity from disparate circuit parameters.
\newblock \emph{Nature neuroscience}, 7\penalty0 (12):\penalty0 1345--1352, 2004.

\bibitem[Quiroga and Panzeri(2009)]{quiroga2009extracting}
R.~Q. Quiroga and S.~Panzeri.
\newblock Extracting information from neuronal populations: information theory and decoding approaches.
\newblock \emph{Nature Reviews Neuroscience}, 10\penalty0 (3):\penalty0 173--185, 2009.

\bibitem[Quiroga et~al.(2005)Quiroga, Reddy, Kreiman, Koch, and Fried]{quiroga2005invariant}
R.~Q. Quiroga, L.~Reddy, G.~Kreiman, C.~Koch, and I.~Fried.
\newblock Invariant visual representation by single neurons in the human brain.
\newblock \emph{Nature}, 435\penalty0 (7045):\penalty0 1102--1107, 2005.

\bibitem[Radford et~al.(2017)Radford, Jozefowicz, and Sutskever]{radford2017learning}
A.~Radford, R.~Jozefowicz, and I.~Sutskever.
\newblock Learning to generate reviews and discovering sentiment.
\newblock \emph{arXiv preprint arXiv:1704.01444}, 2017.

\bibitem[Rahwan et~al.(2019)Rahwan, Cebrian, Obradovich, Bongard, Bonnefon, Breazeal, Crandall, Christakis, Couzin, Jackson, et~al.]{rahwan2019machine}
I.~Rahwan, M.~Cebrian, N.~Obradovich, J.~Bongard, J.-F. Bonnefon, C.~Breazeal, J.~W. Crandall, N.~A. Christakis, I.~D. Couzin, M.~O. Jackson, et~al.
\newblock Machine behaviour.
\newblock \emph{Nature}, 568\penalty0 (7753):\penalty0 477--486, 2019.

\bibitem[Rajan et~al.(2016)Rajan, Harvey, and Tank]{rajan2016recurrent}
K.~Rajan, C.~D. Harvey, and D.~W. Tank.
\newblock Recurrent network models of sequence generation and memory.
\newblock \emph{Neuron}, 90\penalty0 (1):\penalty0 128--142, 2016.

\bibitem[R{\"a}uker et~al.(2022)R{\"a}uker, Ho, Casper, and Hadfield-Menell]{rauker2022toward}
T.~R{\"a}uker, A.~Ho, S.~Casper, and D.~Hadfield-Menell.
\newblock Toward transparent ai: A survey on interpreting the inner structures of deep neural networks.
\newblock \emph{arXiv preprint arXiv:2207.13243}, 2022.

\bibitem[Rehman and Al~Khalili(2023)]{rehman2019neuroanatomy}
A.~Rehman and Y.~Al~Khalili.
\newblock \emph{Neuroanatomy, occipital lobe}.
\newblock StatPearls Publishing, 2023.

\bibitem[Ren and Komiyama(2021)]{ren2021characterizing}
C.~Ren and T.~Komiyama.
\newblock Characterizing cortex-wide dynamics with wide-field calcium imaging.
\newblock \emph{Journal of Neuroscience}, 41\penalty0 (19):\penalty0 4160--4168, 2021.

\bibitem[Rescorla(1972)]{rescorla1972theory}
R.~A. Rescorla.
\newblock A theory of pavlovian conditioning: Variations in the effectiveness of reinforcement and non-reinforcement.
\newblock \emph{Classical conditioning, Current research and theory}, 2:\penalty0 64--69, 1972.

\bibitem[Reynolds(1987)]{Reynolds1987FlocksHA}
C.~W. Reynolds.
\newblock Flocks, herds, and schools: a distributed behavioral model.
\newblock \emph{Seminal graphics: pioneering efforts that shaped the field}, 1987.
\newblock URL \url{https://api.semanticscholar.org/CorpusID:546350}.

\bibitem[Richards et~al.(2019)Richards, Lillicrap, Beaudoin, Bengio, Bogacz, Christensen, Clopath, Costa, de~Berker, Ganguli, et~al.]{richards2019deep}
B.~A. Richards, T.~P. Lillicrap, P.~Beaudoin, Y.~Bengio, R.~Bogacz, A.~Christensen, C.~Clopath, R.~P. Costa, A.~de~Berker, S.~Ganguli, et~al.
\newblock A deep learning framework for neuroscience.
\newblock \emph{Nature neuroscience}, 22\penalty0 (11):\penalty0 1761--1770, 2019.

\bibitem[Riddoch and Humphreys(1983)]{riddoch1983effect}
M.~J. Riddoch and G.~W. Humphreys.
\newblock The effect of cueing on unilateral neglect.
\newblock \emph{Neuropsychologia}, 21\penalty0 (6):\penalty0 589--599, 1983.

\bibitem[Rigotti et~al.(2010)Rigotti, Rubin, Wang, and Fusi]{rigotti2010internal}
M.~Rigotti, D.~B.~D. Rubin, X.-J. Wang, and S.~Fusi.
\newblock Internal representation of task rules by recurrent dynamics: the importance of the diversity of neural responses.
\newblock \emph{Frontiers in computational neuroscience}, 4:\penalty0 24, 2010.

\bibitem[Ritchie et~al.(2019)Ritchie, Kaplan, and Klein]{ritchie2019decoding}
J.~B. Ritchie, D.~M. Kaplan, and C.~Klein.
\newblock Decoding the brain: Neural representation and the limits of multivariate pattern analysis in cognitive neuroscience.
\newblock \emph{The British journal for the philosophy of science}, 70\penalty0 (2):\penalty0 581--607, 2019.

\bibitem[Robertson et~al.(2003)Robertson, Theoret, and Pascual-Leone]{robertson2003studies}
E.~Robertson, H.~Theoret, and A.~Pascual-Leone.
\newblock Studies in cognition: the problems solved and created by transcranial magnetic stimulation.
\newblock \emph{Journal of cognitive neuroscience}, 15\penalty0 (7):\penalty0 948--960, 2003.

\bibitem[Rumelhart and McClelland(1985)]{rumelhart1985levels}
D.~E. Rumelhart and J.~L. McClelland.
\newblock Levels indeed! a response to broadbent.
\newblock \emph{Journal of Experimental Psychology: General}, 1985.

\bibitem[Russo et~al.(2020)Russo, Khajeh, Bittner, Perkins, Cunningham, Abbott, and Churchland]{russo2020neural}
A.~A. Russo, R.~Khajeh, S.~R. Bittner, S.~M. Perkins, J.~P. Cunningham, L.~F. Abbott, and M.~M. Churchland.
\newblock Neural trajectories in the supplementary motor area and motor cortex exhibit distinct geometries, compatible with different classes of computation.
\newblock \emph{Neuron}, 107\penalty0 (4):\penalty0 745--758, 2020.

\bibitem[Sabuncu(2020)]{sabuncu2020intelligence}
M.~R. Sabuncu.
\newblock Intelligence plays dice: Stochasticity is essential for machine learning, 2020.

\bibitem[Salles et~al.(2020)Salles, Evers, and Farisco]{salles2020anthropomorphism}
A.~Salles, K.~Evers, and M.~Farisco.
\newblock Anthropomorphism in ai.
\newblock \emph{AJOB neuroscience}, 11\penalty0 (2):\penalty0 88--95, 2020.

\bibitem[Samborska et~al.(2022)Samborska, Butler, Walton, Behrens, and Akam]{samborska2022complementary}
V.~Samborska, J.~L. Butler, M.~E. Walton, T.~E. Behrens, and T.~Akam.
\newblock Complementary task representations in hippocampus and prefrontal cortex for generalizing the structure of problems.
\newblock \emph{Nature Neuroscience}, 25\penalty0 (10):\penalty0 1314--1326, 2022.

\bibitem[Sanes and Evarts(1984)]{sanes1984motor}
J.~Sanes and E.~Evarts.
\newblock Motor psychophysics.
\newblock \emph{Human Neurobiology}, 2\penalty0 (4):\penalty0 217--225, 1984.

\bibitem[Saxe and Kanwisher(2003)]{saxe2003people}
R.~Saxe and N.~Kanwisher.
\newblock People thinking about thinking people: The role of the temporo-parietal junction in "theory of mind".
\newblock \emph{NeuroImage}, 19\penalty0 (4):\penalty0 1835--1842, 2003.
\newblock \doi{10.1016/s1053-8119(03)00230-1}.

\bibitem[Schrimpf et~al.(2018)Schrimpf, Kubilius, Hong, Majaj, Rajalingham, Issa, Kar, Bashivan, Prescott-Roy, Geiger, et~al.]{schrimpf2018brain}
M.~Schrimpf, J.~Kubilius, H.~Hong, N.~J. Majaj, R.~Rajalingham, E.~B. Issa, K.~Kar, P.~Bashivan, J.~Prescott-Roy, F.~Geiger, et~al.
\newblock Brain-score: Which artificial neural network for object recognition is most brain-like?
\newblock \emph{BioRxiv}, page 407007, 2018.

\bibitem[Schubert et~al.(2021)Schubert, Voss, Cammarata, Goh, and Olah]{schubert2021high}
L.~Schubert, C.~Voss, N.~Cammarata, G.~Goh, and C.~Olah.
\newblock High-low frequency detectors.
\newblock \emph{Distill}, 6\penalty0 (1):\penalty0 e00024--005, 2021.

\bibitem[Scotti et~al.(2024)Scotti, Banerjee, Goode, Shabalin, Nguyen, Dempster, Verlinde, Yundler, Weisberg, Norman, et~al.]{Scotti2023Reconstructing}
P.~Scotti, A.~Banerjee, J.~Goode, S.~Shabalin, A.~Nguyen, A.~Dempster, N.~Verlinde, E.~Yundler, D.~Weisberg, K.~Norman, et~al.
\newblock Reconstructing the mind's eye: fmri-to-image with contrastive learning and diffusion priors.
\newblock \emph{Advances in Neural Information Processing Systems}, 36, 2024.

\bibitem[Seung et~al.(1992)Seung, Sompolinsky, and Tishby]{seung1992statistical}
H.~S. Seung, H.~Sompolinsky, and N.~Tishby.
\newblock Statistical mechanics of learning from examples.
\newblock \emph{Physical review A}, 45\penalty0 (8):\penalty0 6056, 1992.

\bibitem[Shallice(1988)]{shallice1988neuropsychology}
T.~Shallice.
\newblock \emph{From neuropsychology to mental structure}.
\newblock Cambridge University Press, 1988.

\bibitem[Sharma et~al.(2024)Sharma, Tong, Korbak, Duvenaud, Askell, Bowman, DURMUS, Hatfield-Dodds, Johnston, Kravec, Maxwell, McCandlish, Ndousse, Rausch, Schiefer, Yan, Zhang, and Perez]{sharma2024towards}
M.~Sharma, M.~Tong, T.~Korbak, D.~Duvenaud, A.~Askell, S.~R. Bowman, E.~DURMUS, Z.~Hatfield-Dodds, S.~R. Johnston, S.~M. Kravec, T.~Maxwell, S.~McCandlish, K.~Ndousse, O.~Rausch, N.~Schiefer, D.~Yan, M.~Zhang, and E.~Perez.
\newblock Towards understanding sycophancy in language models.
\newblock In \emph{The Twelfth International Conference on Learning Representations}, 2024.
\newblock URL \url{https://openreview.net/forum?id=tvhaxkMKAn}.

\bibitem[Shevlane et~al.(2023)Shevlane, Farquhar, Garfinkel, Phuong, Whittlestone, Leung, Kokotajlo, Marchal, Anderljung, Kolt, et~al.]{shevlane2023model}
T.~Shevlane, S.~Farquhar, B.~Garfinkel, M.~Phuong, J.~Whittlestone, J.~Leung, D.~Kokotajlo, N.~Marchal, M.~Anderljung, N.~Kolt, et~al.
\newblock Model evaluation for extreme risks.
\newblock \emph{arXiv preprint arXiv:2305.15324}, 2023.

\bibitem[Shi et~al.(2019)Shi, Gupta, Boukhvalova, Singer, and Butts]{shi2019functional}
Q.~Shi, P.~Gupta, A.~K. Boukhvalova, J.~H. Singer, and D.~A. Butts.
\newblock Functional characterization of retinal ganglion cells using tailored nonlinear modeling.
\newblock \emph{Scientific reports}, 9\penalty0 (1):\penalty0 8713, 2019.

\bibitem[Shmuelof et~al.(2012)Shmuelof, Huang, Haith, Delnicki, Mazzoni, and Krakauer]{shmuelof2012overcoming}
L.~Shmuelof, V.~S. Huang, A.~M. Haith, R.~J. Delnicki, P.~Mazzoni, and J.~W. Krakauer.
\newblock Overcoming motor “forgetting” through reinforcement of learned actions.
\newblock \emph{Journal of Neuroscience}, 32\penalty0 (42):\penalty0 14617--14621a, 2012.

\bibitem[Simon(1980)]{simon1980cognitive}
H.~A. Simon.
\newblock Cognitive science: The newest science of the artificial.
\newblock \emph{Cognitive science}, 4\penalty0 (1):\penalty0 33--46, 1980.

\bibitem[Simon(2018)]{simon2018near}
H.~A. Simon.
\newblock Near decomposability and complexity: How a mind resides in a brain.
\newblock In \emph{The mind, the brain and complex adaptive systems}, pages 25--44. Routledge, 2018.

\bibitem[Sober(1999)]{sober1999multiple}
E.~Sober.
\newblock The multiple realizability argument against reductionism.
\newblock \emph{Philosophy of science}, 66\penalty0 (4):\penalty0 542--564, 1999.

\bibitem[Sohn et~al.(2019)Sohn, Narain, Meirhaeghe, and Jazayeri]{sohn2019bayesian}
H.~Sohn, D.~Narain, N.~Meirhaeghe, and M.~Jazayeri.
\newblock Bayesian computation through cortical latent dynamics.
\newblock \emph{Neuron}, 103\penalty0 (5):\penalty0 934--947, 2019.

\bibitem[Sporns and Betzel(2016)]{sporns2016modular}
O.~Sporns and R.~F. Betzel.
\newblock Modular brain networks.
\newblock \emph{Annual Review of Psychology}, 67:\penalty0 613--640, 2016.
\newblock \doi{10.1146/annurev-psych-122414-033634}.

\bibitem[Stankiewicz et~al.(2006)Stankiewicz, Legge, Mansfield, and Schlicht]{stankiewicz2006lost}
B.~J. Stankiewicz, G.~E. Legge, J.~S. Mansfield, and E.~J. Schlicht.
\newblock Lost in virtual space: studies in human and ideal spatial navigation.
\newblock \emph{Journal of Experimental Psychology: Human Perception and Performance}, 32\penalty0 (3):\penalty0 688, 2006.

\bibitem[Stevens and Newman(1936)]{stevens1936localization}
S.~S. Stevens and E.~B. Newman.
\newblock The localization of actual sources of sound.
\newblock \emph{The American journal of psychology}, 48\penalty0 (2):\penalty0 297--306, 1936.

\bibitem[Stringer et~al.(2019)Stringer, Pachitariu, Steinmetz, Carandini, and Harris]{stringer2019high}
C.~Stringer, M.~Pachitariu, N.~Steinmetz, M.~Carandini, and K.~D. Harris.
\newblock High-dimensional geometry of population responses in visual cortex.
\newblock \emph{Nature}, 571\penalty0 (7765):\penalty0 361--365, 2019.

\bibitem[Sucholutsky et~al.(2023)Sucholutsky, Muttenthaler, Weller, Peng, Bobu, Kim, Love, Grant, Achterberg, Tenenbaum, et~al.]{sucholutsky2023getting}
I.~Sucholutsky, L.~Muttenthaler, A.~Weller, A.~Peng, A.~Bobu, B.~Kim, B.~C. Love, E.~Grant, J.~Achterberg, J.~B. Tenenbaum, et~al.
\newblock Getting aligned on representational alignment.
\newblock \emph{arXiv preprint arXiv:2310.13018}, 2023.

\bibitem[Sur et~al.(1988)Sur, Garraghty, and Roe]{sur1988experimentally}
M.~Sur, P.~E. Garraghty, and A.~W. Roe.
\newblock Experimentally induced visual projections into auditory thalamus and cortex.
\newblock \emph{Science}, 242\penalty0 (4884):\penalty0 1437--1441, 1988.

\bibitem[Sutton(1988)]{sutton1988learning}
R.~S. Sutton.
\newblock Learning to predict by the methods of temporal differences.
\newblock \emph{Machine learning}, 3:\penalty0 9--44, 1988.

\bibitem[Szegedy et~al.(2013)Szegedy, Zaremba, Sutskever, Bruna, Erhan, Goodfellow, and Fergus]{szegedy2013intriguing}
C.~Szegedy, W.~Zaremba, I.~Sutskever, J.~Bruna, D.~Erhan, I.~Goodfellow, and R.~Fergus.
\newblock Intriguing properties of neural networks.
\newblock \emph{arXiv preprint arXiv:1312.6199}, 2013.

\bibitem[Takahashi et~al.(2012)Takahashi, Narayanan, and Ghazanfar]{takahashi2012computational}
D.~Y. Takahashi, D.~Narayanan, and A.~A. Ghazanfar.
\newblock A computational model for vocal exchange dynamics and their development in marmoset monkeys.
\newblock In \emph{2012 IEEE International Conference on Development and Learning and Epigenetic Robotics (ICDL)}, pages 1--2. IEEE, 2012.

\bibitem[Tang et~al.(2023)Tang, LeBel, Jain, and Huth]{tang2023semantic}
J.~Tang, A.~LeBel, S.~Jain, and A.~G. Huth.
\newblock Semantic reconstruction of continuous language from non-invasive brain recordings.
\newblock \emph{Nature Neuroscience}, 26\penalty0 (5):\penalty0 858--866, 2023.

\bibitem[Taylor et~al.(2014)Taylor, Krakauer, and Ivry]{taylor2014explicit}
J.~A. Taylor, J.~W. Krakauer, and R.~B. Ivry.
\newblock Explicit and implicit contributions to learning in a sensorimotor adaptation task.
\newblock \emph{Journal of Neuroscience}, 34\penalty0 (8):\penalty0 3023--3032, 2014.

\bibitem[Templeton(2024)]{templeton2024scaling}
A.~Templeton.
\newblock \emph{Scaling monosemanticity: Extracting interpretable features from claude 3 sonnet}.
\newblock Anthropic, 2024.

\bibitem[Tenenbaum(1998)]{tenenbaum1998bayesian}
J.~Tenenbaum.
\newblock Bayesian modeling of human concept learning.
\newblock \emph{Advances in neural information processing systems}, 11, 1998.

\bibitem[Tenenbaum and Griffiths(2001)]{tenenbaum2001generalization}
J.~B. Tenenbaum and T.~L. Griffiths.
\newblock Generalization, similarity, and bayesian inference.
\newblock \emph{Behavioral and brain sciences}, 24\penalty0 (4):\penalty0 629--640, 2001.

\bibitem[Tenenbaum et~al.(2011)Tenenbaum, Kemp, Griffiths, and Goodman]{tenenbaum2011grow}
J.~B. Tenenbaum, C.~Kemp, T.~L. Griffiths, and N.~D. Goodman.
\newblock How to grow a mind: Statistics, structure, and abstraction.
\newblock \emph{science}, 331\penalty0 (6022):\penalty0 1279--1285, 2011.

\bibitem[Tenney et~al.(2019{\natexlab{a}})Tenney, Das, and Pavlick]{tenney-etal-2019-bert}
I.~Tenney, D.~Das, and E.~Pavlick.
\newblock {BERT} rediscovers the classical {NLP} pipeline.
\newblock In A.~Korhonen, D.~Traum, and L.~M{\`a}rquez, editors, \emph{Proceedings of the 57th Annual Meeting of the Association for Computational Linguistics}, pages 4593--4601, Florence, Italy, July 2019{\natexlab{a}}. Association for Computational Linguistics.
\newblock \doi{10.18653/v1/P19-1452}.
\newblock URL \url{https://aclanthology.org/P19-1452}.

\bibitem[Tenney et~al.(2019{\natexlab{b}})Tenney, Das, and Pavlick]{tenney2019bert}
I.~Tenney, D.~Das, and E.~Pavlick.
\newblock Bert rediscovers the classical nlp pipeline.
\newblock \emph{arXiv preprint arXiv:1905.05950}, 2019{\natexlab{b}}.

\bibitem[Therrien and Bastian(2015)]{therrien2015cerebellar}
A.~S. Therrien and A.~J. Bastian.
\newblock Cerebellar damage impairs internal predictions for sensory and motor function.
\newblock \emph{Current opinion in neurobiology}, 33:\penalty0 127--133, 2015.

\bibitem[Thompson(2021)]{thompson2021forms}
J.~A. Thompson.
\newblock Forms of explanation and understanding for neuroscience and artificial intelligence.
\newblock \emph{Journal of Neurophysiology}, 2021.

\bibitem[Timme and Lapish(2018)]{timme2018tutorial}
N.~M. Timme and C.~Lapish.
\newblock A tutorial for information theory in neuroscience.
\newblock \emph{e{N}euro}, 5\penalty0 (3), 2018.

\bibitem[Tinbergen(1963{\natexlab{a}})]{Tinbergen1963}
N.~Tinbergen.
\newblock On aims and methods of ethology.
\newblock \emph{Zeitschrift für Tierpsychologie}, 20:\penalty0 410--433, 1963{\natexlab{a}}.
\newblock \doi{10.1111/j.1439-0310.1963.tb01161.x}.

\bibitem[Tinbergen(1963{\natexlab{b}})]{tinbergen1963aims}
N.~Tinbergen.
\newblock On aims and methods of ethology.
\newblock \emph{Zeitschrift f{\"u}r tierpsychologie}, 20\penalty0 (4):\penalty0 410--433, 1963{\natexlab{b}}.

\bibitem[Tu et~al.(2005)Tu, Chen, Yuille, and Zhu]{tu2005image}
Z.~Tu, X.~Chen, A.~L. Yuille, and S.-C. Zhu.
\newblock Image parsing: Unifying segmentation, detection, and recognition.
\newblock \emph{International Journal of computer vision}, 63:\penalty0 113--140, 2005.

\bibitem[Tucker et~al.(2022)Tucker, Eisape, Qian, Levy, and Shah]{tucker-etal-2022-syntax}
M.~Tucker, T.~Eisape, P.~Qian, R.~Levy, and J.~Shah.
\newblock When does syntax mediate neural language model performance? evidence from dropout probes.
\newblock In M.~Carpuat, M.-C. de~Marneffe, and I.~V. Meza~Ruiz, editors, \emph{Proceedings of the 2022 Conference of the North American Chapter of the Association for Computational Linguistics: Human Language Technologies}, pages 5393--5408, Seattle, United States, July 2022. Association for Computational Linguistics.
\newblock \doi{10.18653/v1/2022.naacl-main.394}.
\newblock URL \url{https://aclanthology.org/2022.naacl-main.394}.

\bibitem[Tuli et~al.(2021)Tuli, Dasgupta, Grant, and Griffiths]{tuli2021convolutional}
S.~Tuli, I.~Dasgupta, E.~Grant, and T.~L. Griffiths.
\newblock Are convolutional neural networks or transformers more like human vision?
\newblock \emph{arXiv preprint arXiv:2105.07197}, 2021.

\bibitem[Tye et~al.(2024)Tye, Miller, Taschbach, Benna, Rigotti, and Fusi]{tye2024mixed}
K.~M. Tye, E.~K. Miller, F.~H. Taschbach, M.~K. Benna, M.~Rigotti, and S.~Fusi.
\newblock Mixed selectivity: Cellular computations for complexity.
\newblock \emph{Neuron}, 2024.

\bibitem[Ullman and Tenenbaum(2020)]{ullman2020bayesian}
T.~D. Ullman and J.~B. Tenenbaum.
\newblock Bayesian models of conceptual development: Learning as building models of the world.
\newblock \emph{Annual Review of Developmental Psychology}, 2:\penalty0 533--558, 2020.

\bibitem[Ullman et~al.(2017)Ullman, Spelke, Battaglia, and Tenenbaum]{ullman2017mind}
T.~D. Ullman, E.~Spelke, P.~Battaglia, and J.~B. Tenenbaum.
\newblock Mind games: Game engines as an architecture for intuitive physics.
\newblock \emph{Trends in cognitive sciences}, 21\penalty0 (9):\penalty0 649--665, 2017.

\bibitem[Vaghi et~al.(2017)Vaghi, Luyckx, Sule, Fineberg, Robbins, and De~Martino]{vaghi2017compulsivity}
M.~M. Vaghi, F.~Luyckx, A.~Sule, N.~A. Fineberg, T.~W. Robbins, and B.~De~Martino.
\newblock Compulsivity reveals a novel dissociation between action and confidence.
\newblock \emph{Neuron}, 96\penalty0 (2):\penalty0 348--354, 2017.

\bibitem[van Opheusden et~al.(2023)van Opheusden, Kuperwajs, Galbiati, Bnaya, Li, and Ma]{van2023expertise}
B.~van Opheusden, I.~Kuperwajs, G.~Galbiati, Z.~Bnaya, Y.~Li, and W.~J. Ma.
\newblock Expertise increases planning depth in human gameplay.
\newblock \emph{Nature}, 618\penalty0 (7967):\penalty0 1000--1005, 2023.

\bibitem[Verdejo and Quesada(2011)]{verdejo2011levels}
V.~M. Verdejo and D.~Quesada.
\newblock Levels of explanation vindicated.
\newblock \emph{Review of Philosophy and Psychology}, 2:\penalty0 77--88, 2011.

\bibitem[Vilas et~al.(2024)Vilas, Adolfi, Poeppel, and Roig]{vilas2024position}
M.~G. Vilas, F.~Adolfi, D.~Poeppel, and G.~Roig.
\newblock Position paper: An inner interpretability framework for ai inspired by lessons from cognitive neuroscience.
\newblock \emph{arXiv preprint arXiv:2406.01352}, 2024.

\bibitem[Voita and Titov(2020)]{voita-titov-2020-information}
E.~Voita and I.~Titov.
\newblock Information-theoretic probing with minimum description length.
\newblock In B.~Webber, T.~Cohn, Y.~He, and Y.~Liu, editors, \emph{Proceedings of the 2020 Conference on Empirical Methods in Natural Language Processing (EMNLP)}, pages 183--196, Online, Nov. 2020. Association for Computational Linguistics.
\newblock \doi{10.18653/v1/2020.emnlp-main.14}.
\newblock URL \url{https://aclanthology.org/2020.emnlp-main.14}.

\bibitem[Voita et~al.(2019)Voita, Talbot, Moiseev, Sennrich, and Titov]{voita2019analyzing}
E.~Voita, D.~Talbot, F.~Moiseev, R.~Sennrich, and I.~Titov.
\newblock Analyzing multi-head self-attention: Specialized heads do the heavy lifting, the rest can be pruned.
\newblock \emph{arXiv preprint arXiv:1905.09418}, 2019.

\bibitem[Voita et~al.(2023)Voita, Ferrando, and Nalmpantis]{voita2023neurons}
E.~Voita, J.~Ferrando, and C.~Nalmpantis.
\newblock Neurons in large language models: Dead, n-gram, positional.
\newblock \emph{arXiv preprint arXiv:2309.04827}, 2023.

\bibitem[Von~Oswald et~al.(2023)Von~Oswald, Niklasson, Randazzo, Sacramento, Mordvintsev, Zhmoginov, and Vladymyrov]{von2023transformers}
J.~Von~Oswald, E.~Niklasson, E.~Randazzo, J.~Sacramento, A.~Mordvintsev, A.~Zhmoginov, and M.~Vladymyrov.
\newblock Transformers learn in-context by gradient descent.
\newblock In \emph{International Conference on Machine Learning}, pages 35151--35174. PMLR, 2023.

\bibitem[Voss et~al.(2021)Voss, Goh, Cammarata, Petrov, Schubert, and Olah]{voss2021branch}
C.~Voss, G.~Goh, N.~Cammarata, M.~Petrov, L.~Schubert, and C.~Olah.
\newblock Branch specialization.
\newblock \emph{Distill}, 6\penalty0 (4):\penalty0 e00024--008, 2021.

\bibitem[Wagner(1999)]{wagner1999causality}
A.~Wagner.
\newblock Causality in complex systems.
\newblock \emph{Biology and Philosophy}, 14:\penalty0 83--101, 1999.

\bibitem[Wan et~al.(2024)Wan, Wang, Liu, Alam, Zheng, Liu, Qu, Yan, Zhu, Zhang, Chowdhury, and Zhang]{wan2024efficient}
Z.~Wan, X.~Wang, C.~Liu, S.~Alam, Y.~Zheng, J.~Liu, Z.~Qu, S.~Yan, Y.~Zhu, Q.~Zhang, M.~Chowdhury, and M.~Zhang.
\newblock Efficient large language models: A survey, 2024.

\bibitem[Wang et~al.(2023)Wang, Variengien, Conmy, Shlegeris, and Steinhardt]{wang2023interpretability}
K.~R. Wang, A.~Variengien, A.~Conmy, B.~Shlegeris, and J.~Steinhardt.
\newblock Interpretability in the wild: a circuit for indirect object identification in {GPT}-2 small.
\newblock In \emph{The Eleventh International Conference on Learning Representations}, 2023.
\newblock URL \url{https://openreview.net/forum?id=NpsVSN6o4ul}.

\bibitem[Waskom et~al.(2019)Waskom, Okazawa, and Kiani]{waskom2019designing}
M.~L. Waskom, G.~Okazawa, and R.~Kiani.
\newblock Designing and interpreting psychophysical investigations of cognition.
\newblock \emph{Neuron}, 104\penalty0 (1):\penalty0 100--112, 2019.

\bibitem[Watanabe(2009)]{watanabe2009algebraic}
S.~Watanabe.
\newblock \emph{Algebraic geometry and statistical learning theory}, volume~25.
\newblock Cambridge university press, 2009.

\bibitem[Watson(1913)]{watson1913psychology}
J.~B. Watson.
\newblock Psychology as the behaviorist views it.
\newblock \emph{Psychological review}, 20\penalty0 (2):\penalty0 158, 1913.

\bibitem[Weichwald et~al.(2015)Weichwald, Meyer, {\"O}zdenizci, Sch{\"o}lkopf, Ball, and Grosse-Wentrup]{weichwald2015causal}
S.~Weichwald, T.~Meyer, O.~{\"O}zdenizci, B.~Sch{\"o}lkopf, T.~Ball, and M.~Grosse-Wentrup.
\newblock Causal interpretation rules for encoding and decoding models in neuroimaging.
\newblock \emph{Neuroimage}, 110:\penalty0 48--59, 2015.

\bibitem[Weiss et~al.(2002)Weiss, Simoncelli, and Adelson]{weiss2002motion}
Y.~Weiss, E.~P. Simoncelli, and E.~H. Adelson.
\newblock Motion illusions as optimal percepts.
\newblock \emph{Nature neuroscience}, 5\penalty0 (6):\penalty0 598--604, 2002.

\bibitem[White and Cotterell(2021)]{white2021examining}
J.~C. White and R.~Cotterell.
\newblock Examining the inductive bias of neural language models with artificial languages.
\newblock \emph{arXiv preprint arXiv:2106.01044}, 2021.

\bibitem[Willems(2011)]{willems2011re}
R.~M. Willems.
\newblock Re-appreciating the why of cognition: 35 years after marr and poggio.
\newblock \emph{Frontiers in psychology}, 2:\penalty0 244, 2011.

\bibitem[Williamson and Prybutok(2024)]{williamson2024era}
S.~M. Williamson and V.~Prybutok.
\newblock The era of artificial intelligence deception: Unraveling the complexities of false realities and emerging threats of misinformation.
\newblock \emph{Information}, 15\penalty0 (6):\penalty0 299, 2024.

\bibitem[Wojcik et~al.(2023)Wojcik, Stroud, Wasmuht, Kusunoku, Kadohisa, Myers, Hunt, Duncan, and Stokes]{wojcik2023learning}
M.~J. Wojcik, J.~P. Stroud, D.~F. Wasmuht, M.~Kusunoku, M.~Kadohisa, N.~E. Myers, L.~T. Hunt, J.~Duncan, and M.~G. Stokes.
\newblock Learning shapes neural geometry in the prefrontal cortex.
\newblock \emph{bioRxiv}, pages 2023--04, 2023.

\bibitem[Wolfe(2020)]{wolfe2020visual}
J.~M. Wolfe.
\newblock Visual search: How do we find what we are looking for?
\newblock \emph{Annual review of vision science}, 6:\penalty0 539--562, 2020.

\bibitem[Wong et~al.(2023)Wong, Grand, Lew, Goodman, Mansinghka, Andreas, and Tenenbaum]{wong2023word}
L.~Wong, G.~Grand, A.~K. Lew, N.~D. Goodman, V.~K. Mansinghka, J.~Andreas, and J.~B. Tenenbaum.
\newblock From word models to world models: Translating from natural language to the probabilistic language of thought.
\newblock \emph{arXiv preprint arXiv:2306.12672}, 2023.

\bibitem[Woolgar et~al.(2016)Woolgar, Jackson, and Duncan]{woolgar2016coding}
A.~Woolgar, J.~Jackson, and J.~Duncan.
\newblock Coding of visual, auditory, rule, and response information in the brain: 10 years of multivoxel pattern analysis.
\newblock \emph{Journal of cognitive neuroscience}, 28\penalty0 (10):\penalty0 1433--1454, 2016.

\bibitem[Woolgar et~al.(2018)Woolgar, Duncan, Manes, and Fedorenko]{woolgar2018fluid}
A.~Woolgar, J.~Duncan, F.~Manes, and E.~Fedorenko.
\newblock Fluid intelligence is supported by the multiple-demand system not the language system.
\newblock \emph{Nature Human Behaviour}, 2\penalty0 (3):\penalty0 200--204, 2018.

\bibitem[Xie et~al.(2022)Xie, Raghunathan, Liang, and Ma]{xie2022explanation}
S.~M. Xie, A.~Raghunathan, P.~Liang, and T.~Ma.
\newblock An explanation of in-context learning as implicit bayesian inference, 2022.

\bibitem[Xu and Tenenbaum(2007)]{xu2007word}
F.~Xu and J.~B. Tenenbaum.
\newblock Word learning as bayesian inference.
\newblock \emph{Psychological review}, 114\penalty0 (2):\penalty0 245, 2007.

\bibitem[Xu et~al.(2023)Xu, Wang, Li, Luo, Wang, Liu, and Liu]{xu2023exploring}
Y.~Xu, S.~Wang, P.~Li, F.~Luo, X.~Wang, W.~Liu, and Y.~Liu.
\newblock Exploring large language models for communication games: An empirical study on werewolf.
\newblock \emph{arXiv preprint arXiv:2309.04658}, 2023.

\bibitem[Yamins and DiCarlo(2016)]{yamins2016using}
D.~L. Yamins and J.~J. DiCarlo.
\newblock Using goal-driven deep learning models to understand sensory cortex.
\newblock \emph{Nature neuroscience}, 19\penalty0 (3):\penalty0 356--365, 2016.

\bibitem[Yasuda et~al.(2006)Yasuda, Harvey, Zhong, Sobczyk, Van~Aelst, and Svoboda]{yasuda2006supersensitive}
R.~Yasuda, C.~D. Harvey, H.~Zhong, A.~Sobczyk, L.~Van~Aelst, and K.~Svoboda.
\newblock Supersensitive ras activation in dendrites and spines revealed by two-photon fluorescence lifetime imaging.
\newblock \emph{Nature neuroscience}, 9\penalty0 (2):\penalty0 283--291, 2006.

\bibitem[Yuste(2015)]{yuste2015neuron}
R.~Yuste.
\newblock From the neuron doctrine to neural networks.
\newblock \emph{Nature reviews neuroscience}, 16\penalty0 (8):\penalty0 487--497, 2015.

\bibitem[Zednik and J{\"a}kel(2016)]{zednik2016bayesian}
C.~Zednik and F.~J{\"a}kel.
\newblock Bayesian reverse-engineering considered as a research strategy for cognitive science.
\newblock \emph{Synthese}, 193:\penalty0 3951--3985, 2016.

\bibitem[Zeiler and Fergus(2014)]{zeiler2014visualizing}
M.~D. Zeiler and R.~Fergus.
\newblock Visualizing and understanding convolutional networks.
\newblock In \emph{Computer Vision--ECCV 2014: 13th European Conference, Zurich, Switzerland, September 6-12, 2014, Proceedings, Part I 13}, pages 818--833. Springer, 2014.

\bibitem[Zhang and Nanda(2023)]{zhang2023towards}
F.~Zhang and N.~Nanda.
\newblock Towards best practices of activation patching in language models: Metrics and methods.
\newblock \emph{arXiv preprint arXiv:2309.16042}, 2023.

\bibitem[Zhang et~al.(2023)Zhang, Press, Merrill, Liu, and Smith]{zhang2023language}
M.~Zhang, O.~Press, W.~Merrill, A.~Liu, and N.~A. Smith.
\newblock How language model hallucinations can snowball.
\newblock \emph{arXiv preprint arXiv:2305.13534}, 2023.

\bibitem[Zhao et~al.(2024)Zhao, Yang, Shen, Lakkaraju, and Du]{zhao2024uncoveringlargelanguagemodel}
H.~Zhao, F.~Yang, B.~Shen, H.~Lakkaraju, and M.~Du.
\newblock Towards uncovering how large language model works: An explainability perspective, 2024.
\newblock URL \url{https://arxiv.org/abs/2402.10688}.

\bibitem[Zhou et~al.(2024)Zhou, Alon, Chen, Wang, Agarwal, and Zhou]{zhou2024transformers}
Y.~Zhou, U.~Alon, X.~Chen, X.~Wang, R.~Agarwal, and D.~Zhou.
\newblock Transformers can achieve length generalization but not robustly, 2024.

\bibitem[Zimnik et~al.(2024)Zimnik, Ames, An, Driscoll, Lara, Russo, Susoy, Cunningham, Paninski, Churchland, et~al.]{zimnik2024identifying}
A.~J. Zimnik, K.~C. Ames, X.~An, L.~Driscoll, A.~H. Lara, A.~A. Russo, V.~Susoy, J.~P. Cunningham, L.~Paninski, M.~M. Churchland, et~al.
\newblock Identifying interpretable latent factors with sparse component analysis.
\newblock \emph{bioRxiv}, pages 2024--02, 2024.

\bibitem[Zou et~al.(2023)Zou, Phan, Chen, Campbell, Guo, Ren, Pan, Yin, Mazeika, Dombrowski, et~al.]{zou2023representation}
A.~Zou, L.~Phan, S.~Chen, J.~Campbell, P.~Guo, R.~Ren, A.~Pan, X.~Yin, M.~Mazeika, A.-K. Dombrowski, et~al.
\newblock Representation engineering: A top-down approach to ai transparency.
\newblock \emph{arXiv preprint arXiv:2310.01405}, 2023.

\end{thebibliography}

\end{document}